\documentclass[final,3p,times]{elsarticle}

\usepackage{amssymb}

\usepackage{lineno}

\journal{Energy and AI}

\usepackage[acronym]{glossaries}
\usepackage{graphicx}
\usepackage{caption}
\usepackage{subcaption} 
\usepackage{siunitx}
\usepackage{array}
\usepackage{makecell}
\usepackage{tabularx}
\usepackage{multirow}
\usepackage{url}
\usepackage{rotating}
\usepackage{booktabs}
\usepackage{multirow}
\usepackage{url}

\usepackage{booktabs}
\newacronym{wdns}{WDNs}{Water Distribution Networks}
\newacronym{wdn}{WDN}{Water Distribution Network}
\newacronym{dhns}{DHNs}{District Heating Networks}
\newacronym{dhn}{DHN}{District Heating Network}
\newacronym{dhs}{DHS}{District Heating Systems}

\newacronym{dmas}{DMAs}{District Metered Areas}
\newacronym{dma}{DMA}{District Metered Area}

\newacronym{cwt}{CWT}{Continuous Wavelet Transform}
\newacronym{dl}{DL}{Deep Learning}
\newacronym{ml}{ML}{Machine Learning}
\newacronym{anns}{ANNs}{Artificial Neural Networks}
\newacronym{ann}{ANN}{Artificial Neural Network}
\newacronym{ai}{AI}{Artificial Intelligence}
\newacronym{rnn}{RNNs}{Recurrent Neural Networks}
\newacronym{sari}{SARIMAX}{Seasonal Auto-Regressive Integrated Moving Average with eXogenous factors}
\newacronym{cnn}{CNN}{Convolutional Neural Network}
\newacronym{cnns}{CNNs}{Convolutional Neural Networks}
\newacronym{tft}{TFT}{Temporal Fusion Transformers}

\newacronym{DA}{da}{Day Ahead}
\newacronym{ID}{id}{Inter Day}
\newacronym{rmse}{RMSE}{Root Mean Squared Error}
\newacronym{mse}{MSE}{Mean Squared Error}
\newacronym{mae}{MAE}{Mean Absolute Error}
\newacronym{mape}{MAPE}{Mean Absolute Percentage Error}
\newacronym{smae}{SMAE}{Seasonal Mean Absolute Error}
\newacronym{mlp}{MLP}{Multi-Layer Perceptron}

\newacronym{arima}{ARIMA}{Autoregressive Integrated Moving Average}
\newacronym{sarima}{SARIMA}{Seasonal Auto-Regressive Integrated Moving Average}
\newacronym{sarimax}{SARIMAX}{Seasonal Auto-Regressive Integrated Moving Average with EXogenous regressors}

\newacronym{mlr}{MLR}{Multiple Linear Regression}

\newacronym{rnns}{RNNs}{Recurrent Neural Networks}

\newacronym{lstm}{LSTM}{Long Short Term Memory}
\newacronym{ttm}{TTM}{Tiny Time Mixers}
\newacronym{svr}{SVR}{Support Vector Machines}
\newacronym{dnn}{DNN}{Deep Neural Network}
\newacronym{slp}{SLPs}{Standard Load Profiles}
\newacronym{ffnn}{FFNN}{Feed Forward Neural Networks}
\newacronym{anfis}{ANFIS}{Adaptive Neuro Fuzzy Inference system}
\newacronym{elm}{ELM}{Extreme Learning Machine}
\newacronym{gpr}{GPR}{Gaussian Process Regression}
\newacronym{gam}{GAMs}{Generalized Additive Models}
\newacronym{xai}{XAI}{Explainable AI}
\newacronym{shap}{SHAP}{SHapley Additive exPlanations}
\newacronym{dwt}{DWT}{Discrete Wavelet Transform}
\newacronym{relu}{ReLU}{Rectified Linear Unit}
\newacronym{kan}{KAN}{Kolmogorov-Arnold network}

\makeglossaries

\begin{document}

\begin{frontmatter}



\title{A Deep Learning Framework for Heat Demand Forecasting using Time-Frequency Representations of Decomposed Features}

\author[inst1]{Adithya Ramachandran\corref{cor1}}\ead{adithya.ramachandran@fau.de}

\author[inst1]{Satyaki Chatterjee}
\author[inst2]{Thorkil Flensmark B. Neergaard}
\author[inst3]{Maximilian Oberndoerfer}
\author[inst1]{Andreas Maier}
\author[inst1]{Siming Bayer}

\affiliation[inst1]{organization={Pattern Recognition Lab, Friedrich Alexander Universität},
            addressline={Martenstrasse 3}, 
            city={Erlangen},
            postcode={91054}, 
            state={Bayern},
            country={Germany}}

\affiliation[inst2]{organization={Brønderslev Forsyning A/S},
            addressline={Virksomhedsvej 20}, 
            city={Brønderslev},
            postcode={9700}, 
            country={Denmark}}

\affiliation[inst3]{organization={SBU Analytics and Services, Diehl Metering GmbH},
                    addressline={Donaustrasse 120}, 
                    city={Nuremberg},
                    postcode={90451}, 
                    country={Germany}}

\cortext[cor1]{Corresponding author at: Pattern Recognition Lab, Friedrich Alexander Universität, Erlangen, Germany}



\begin{abstract}

District Heating Systems are essential infrastructure for delivering heat to consumers across a geographic region sustainably, yet efficient management relies on optimizing diverse energy sources, such as wood, gas, electricity, and solar, in response to fluctuating demand. Aligning supply with demand is critical not only for ensuring reliable heat distribution but also for minimizing carbon emissions and extending infrastructure lifespan through lower operating temperatures. 
However, accurate multi-step forecasting to support these goals remains challenging due to complex, non-linear usage patterns and external dependencies.
In this work, we propose a novel deep learning framework for day-ahead heat demand prediction that leverages time-frequency representations of historical data. By applying Continuous Wavelet Transform to decomposed demand and external meteorological factors, our approach enables Convolutional Neural Networks to learn hierarchical temporal features that are often inaccessible to standard time domain models. We systematically evaluate this method against statistical baselines, state-of-the-art Transformers, and emerging foundation models using multi-year data from three distinct Danish districts, a Danish city, and a German city.
The results show a significant advancement, reducing the Mean Absolute Error by $36\%$ to $43\%$ compared to the strongest baselines, achieving forecasting accuracy of up to $95\%$ across annual test datasets. 
Qualitative and statistical analyses further confirm the accuracy and robustness by reliably tracking volatile demand peaks where others fail. 
This work contributes both a high-performance forecasting architecture and critical insights into optimal feature composition, offering a validated solution for modern energy applications.

\end{abstract}

\begin{graphicalabstract}
\includegraphics[width=1.0\linewidth]{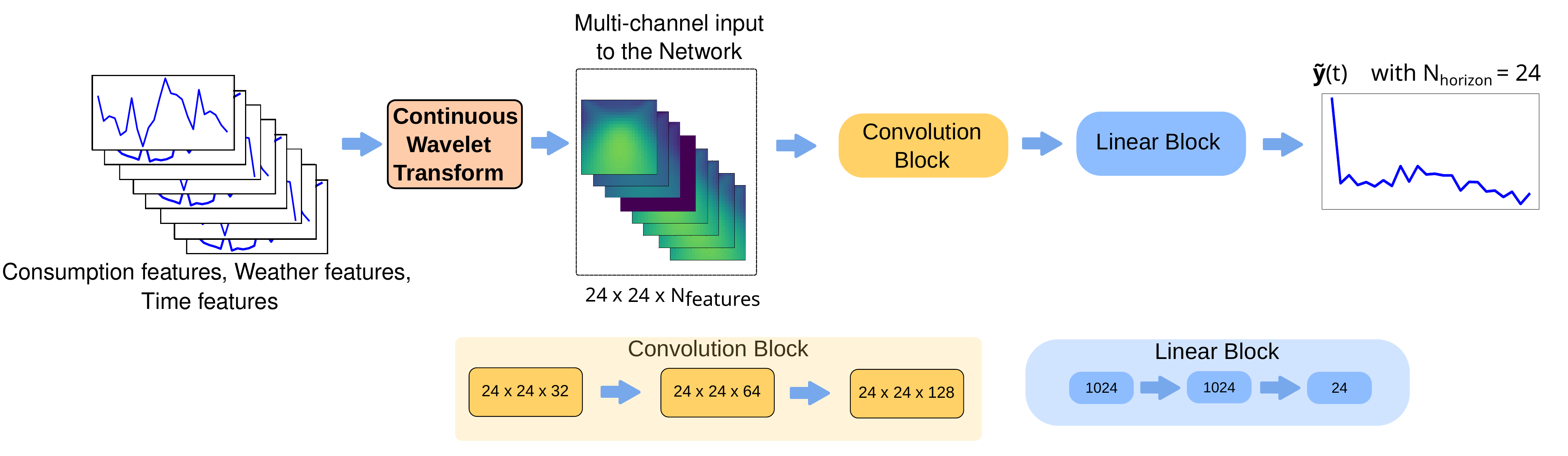}
\end{graphicalabstract}

\begin{highlights}
\item A novel deep learning method using wavelets for heat demand forecasting is proposed.
\item Time-series decomposition of features significantly improves accuracy.
\item Systematic evaluation of the impact of demand, weather, and calendrical features.
\item State-of-the-art benchmarks for 24-step ahead hourly heat demand forecasting.
\item Demonstrates robustness and is validated using multi-year real-world data.
\end{highlights}


\end{frontmatter}


\section{Introduction}
\label{sec:introduction}

Space heating and hot water are essential for wellbeing in cold climates, making \acrfull{dhs} a vital component of sustainable infrastructure that provides scalability, avenues for decarbonization, and diversification of fuel sources \cite{rismanchi_district_2017}. \acrshort{dhs} occupy a critical position in the energy supply chain, distributing thermal energy from centralized generation facilities through underground networks to diverse residential, commercial, and industrial end-users. For efficient monitoring, these networks are typically divided into geographic regions known as \acrfull{dmas} \cite{STECHER2024134016}. With the global push towards climate neutrality \cite{keramidas2021global}, optimizing the balance between supply, which is governed by resource availability, and demand that is shaped by complex consumer behaviours, has become paramount.

In this context, accurate heat demand forecasting is a prerequisite for system optimisation. The advent of smart metering infrastructure \cite{chatterjee2021prediction} has revolutionized the field by enabling high-resolution monitoring of consumption patterns, allowing operators to analyze how extrinsic and intrinsic factors impact demand \cite{schneider2024data}. As \acrshort{dhs} expand across major countries \cite{Miu_Nazare_Diaconu_2022} and increasingly rely on fluctuating renewable energy sources, predictive modelling becomes indispensable. Reliable forecasts allow for better production planning, network balancing, and demand-side management, ultimately minimizing energy losses and peak loads by enabling intelligent control of energy systems to create more cost-effective and sustainable heating solutions.

\subsection{Related Works}
Forecasting approaches aim to predict heat demand with high accuracy and consistency, enabling operators to transition from reactive operations to planned, data-driven energy management. Short-term forecasts allow operators to optimize the operation of heat production units such as boilers, gas engines, and heat pumps, reducing fuel consumption, operational costs, emissions, and the inertial effects of shutdowns and start-ups \cite{pedersen2024annex57}. They also enable proactive adjustment of supply temperatures and flow rates in the distribution network \cite{golla_operational_2020}, which reduces heat losses while ensuring sufficient supply to all consumers. Beyond operational efficiency, reliable forecasts support participation in energy markets and inform fuel procurement strategies, contributing to the stability of the wider energy network \cite{stienecker_impact_2024}\cite{guericke_optimization_2022}.
However, developing effective forecasting models is challenging because they must account for historical consumption patterns, weather variability, psychographic characteristics of consumers, and non-deterministic components that introduce non-linearity and noise \cite{werner_district_2013}. Furthermore, scaling to city-level forecasts requires aggregating data from thousands of heterogeneous sources, which may introduce inconsistencies in measurement protocols \cite{Anand_Nateghi_Alemazkoor_2023}.

Addressing the inherent complexities of heat demand forecasting in \acrshort{dhs} has been the focus of significant research over the past two decades. Forecasting methodologies for \acrshort{dhs} have evolved from classical statistical approaches to sophisticated \acrfull{dl} architectures. Current literature can be broadly categorized into the evolution of time-series architectures and emerging paradigms in representation learning.

\subsubsection{Statistical and Machine Learning}
Early foundational research focused on capturing linear dependencies. Dotzauer \cite{DOTZAUER2002277} established models based on linear least squares, primarily correlating heat demand with outdoor temperature. Recognizing the non-stationary nature of heat demand, classical time series methods found prominence. Grosswindhager et al. \cite{grosswindhager} utilized \acrfull{sarima} within a state-space framework, while Fang \& Lahdelma \cite{fang_evaluation_2016} demonstrated the benefit of incorporating outdoor temperature as an exogenous variable in \acrfull{sarimax} models compared to \acrfull{mlr}.

The advent of \acrfull{ml} offered new avenues for modelling non-linearities. Studies by Idowu et al. \cite{idowu_applied_2016}, Petkovic et al. \cite{petkovic_evaluation_2015}, and Protic et al. \cite{protic_forecasting_2015} investigated a range of shallow algorithms including \acrfull{svr}, Regression Trees, and \acrfull{ffnn}. A comprehensive comparison by Potocnik et al. \cite{potocnik_machine-learning-based_2021} found \acrfull{gpr} and well-tuned neural networks to be particularly effective for multi-step ahead forecasting. However, these methods typically require extensive manual feature engineering and often lack the capacity to model long-range temporal dependencies effectively \cite{potocnik_machine-learning-based_2021}. Furthermore, they are less adaptable to the volatile, non-linear transitions seen in modern energy systems.

\subsubsection{Deep Learning Architectures}
To capture complex temporal dynamics, \acrshort{dl} methodologies gained significant traction. Suryanarayana et al. \cite{suryanarayana_thermal_2018} demonstrated the superiority of deep \acrshort{ffnn}s over linear models in handling higher-dimensional inputs. To address sequential dependencies, Recurrent architectures became the standard. Advanced variants like integrated D-CNN-LSTMs by Yao et al. \cite{yao_integrated_2022} combined the feature extraction capabilities of \acrfull{cnn} with the temporal memory of \acrfull{lstm}. Similarly, Xue et al. \cite{xue_multi-step_2019} explored \acrshort{svr}, \acrshort{dnn}, and XGBoost for multi-step ahead forecasting. Although early \acrshort{dl} models like \acrfull{lstm} and shallow architectures represent a significant advancement over linear methods, establishing supremacy, they often struggle to capture high-frequency volatility and long sequences due to vanishing gradients and a lack of representational power.

These limitations propelled the evolution toward more powerful architectures. In recent years, the broader field of \acrshort{dl} for time series forecasting has undergone significant architectural innovations, largely driven by the success of the Transformer model \cite{vaswani2023attentionneed}. Architectures such as Informer \cite{haoyietal-informer-2021}, Autoformer \cite{wu2021autoformer}, and TimesNet \cite{wu2023timesnet} have demonstrated state-of-the-art performance by effectively capturing long-range dependencies and complex temporal patterns. Their efficacy has been consistently benchmarked on diverse domains, which often serves as a foundational inspiration for \acrshort{dhs} such as the use of Informer \cite{gong_load_2022}, and \acrfull{tft} \cite{frison_evaluating_2024}.
Furthermore, models specific to hourly heat demand forecasting are coming to the forefront, such as the TBMK network, which fuses Bi-Mamba2 and \acrshort{kan}s \cite{TBMK}, and Cakformer, which integrates \acrfull{kan} into a Transformer architecture to capture long-term dependencies \cite{cakformer}.

Despite this cross-domain influence, the systematic adoption and rigorous evaluation of these cutting-edge \acrshort{dl} architectures for 24-step ahead hourly \acrshort{dhs} heat demand forecasting is nascent and reveals critical gaps. For instance, evaluations of models like Informer and \acrshort{tft} were conducted on datasets spanning extremely short periods of $33$ days \cite{gong_load_2022} or from single institutions, raising questions about seasonal robustness and generalisability across heterogeneous consumer profiles \cite{frison_evaluating_2024}. This trend of evaluation on short-term datasets persists across the latest research as well. For instance, the TBMK model was validated on a 68-day winter dataset \cite{TBMK}, and the Cakformer model used datasets spanning only three to four months \cite{cakformer} in its entirety. Such short temporal windows miss crucial shoulder seasons where demand patterns are highly non-linear and fail to capture inter-annual variations. This challenge is compounded by the use of data-hungry Transformer-based models, which contrasts with the general scarcity of extensive, long-duration, public \acrshort{dhs} datasets. Additionally, many studies rely on data from specific industrial sites \cite{CHEN2025126783} or university buildings  \cite{YANG2025127552}, and private sources \cite{ZHANG2024123696}, limiting realistic representation and reproducibility and focusing on homogeneous, rather than diverse, user districts.

The recent use of public datasets, notably from Aalborg, Denmark \cite{aalborg}, signifies a positive step toward more reproducible research. However, its application in prominent recent works highlights a different facet of research. Studies employing this dataset, with active deep learning \cite{huang_active} multi-scale temporal representation \cite{huang_multiscale}, and sparse dynamic graph learning \cite{huang_sparsegraph}, focus on forecasting at a daily resolution and for a homogeneous set of users in isolation. The authors of these works themselves note the need for future validation on datasets with more varied user types and climatic conditions \cite{huang_sparsegraph}. This difference in forecasting objective highlights a persistent gap in demand forecasting for \acrshort{dhs} comprising heterogeneous users that align with real-world \acrshort{dhs} open, which is critical for the intelligent control of \acrshort{dhs} \cite{frison_evaluating_2024}.

In tandem with architectural evolution, advancements in data representation and feature engineering showcased promising avenues for forecasting through time-series decomposition \cite{ts_decomp}, phase space representation \cite{psr}, and time-frequency representations such as wavelets \cite{cwt_forecasting}. Recent work has demonstrated decomposition to be a powerful preprocessing step for industrial heat load forecasting \cite{CHEN2025126783}. Techniques employing \acrfull{cwt} have shown utility in capturing temporal-periodic features that are inaccessible in the standard time domain \cite{ramachandran2022heatdemandforecastingmultiresolutional}, but they have seen limited systematic study for enhancing hourly heat demand predictions \cite{ramachandran2025advancingheatdemandforecasting}.
Concurrently, an entirely new frontier is also emerging with the advent of foundation models \cite{TFM}. These models, pre-trained on vast corpora of temporal data, present a significant opportunity for industrial applications where system-specific historical data might be scarce, an area that remains completely unexplored for heat demand forecasting, underscoring a critical gap requires addressing.

\subsection{Contributions}
\label{sec:contributions}

Our work aims to address the identified gaps in \acrshort{dhs} forecasting through a multi-faceted approach. Our primary contribution is a novel \acrshort{dl} framework that reframes the forecasting problem by transforming one-dimensional time-series into two-dimensional time-frequency representations. The contributions are:

\begin{itemize}
    \item To address the fragmented and often incomparable evaluation landscape, we conduct a comprehensive benchmark of modern \acrshort{ai} architectures. This evaluation, performed on a multi-year, hourly dataset from heterogeneous districts, ranges from canonical \acrshort{lstm}s to state-of-the-art Transformers and mixing models,  including fine-tuned foundation models of TTM \cite{ekambaram2024tinytimemixersttms}  and Chronos-2 \cite{ansari2025chronos2}, thereby establishing a standard for this specific forecasting task. In addition, to demonstrate generalisability, we also showcase the forecasting performance on available public datasets for the \acrshort{dhs} of Flensburg, Germany \cite{stadtwerke_flensburg_gmbh_2019_2562658}, and residential demand from Aalborg, Denmark \cite{aalborg}.
    
    \item To address the ad-hoc nature of feature engineering, we provide a systematic evaluation of the impact of feature representation. This includes a systematic evaluation of encoding schemas for complex, non-standard events like public holidays, and presents a rigorous analysis in addition to reporting established correlations of how the decomposed trend, seasonal, and residual components of both demand and weather series interact, offering a clearer, data-driven understanding of the fundamental drivers of heat demand.
    
    \item To address the limitations of standard one-dimensional models and the underexplored potential of alternative data representations, we introduce a novel time-frequency \acrshort{dl} framework. By transforming specific decomposed time-series into two-dimensional scalograms via a \acrfull{cwt}, our model enables a \acrshort{cnn} to learn hierarchical, temporal-periodic patterns. This approach unequivocally establishes a new performance benchmark, significantly outperforming all baseline models by reducing the mean absolute error by approximately 31\% to 33\% and providing a consistent forecasting accuracy of $95\%$.
\end{itemize}

As \acrshort{dhs} infrastructure becomes more complex and integrated with volatile renewable energy, the demand for forecasting models with such unprecedented accuracy and reliability are an operational necessity. The modelling results, validated on a multi-year dataset, demonstrate a robust solution ready for real-world deployment. By providing more accurate day-ahead forecasts, our work directly contributes to applied values such as optimizing boiler and heat pump schedules to reduce fuel consumption and enabling more effective participation in energy markets, and catering to emerging solutions such as model predictive control of \acrshort{dhs}.
\section{Methodology}

\subsection{Data Overview} \label{sec:data_overview}
\begin{figure}[htbp]
\centering
\includegraphics[width=1.0\linewidth]{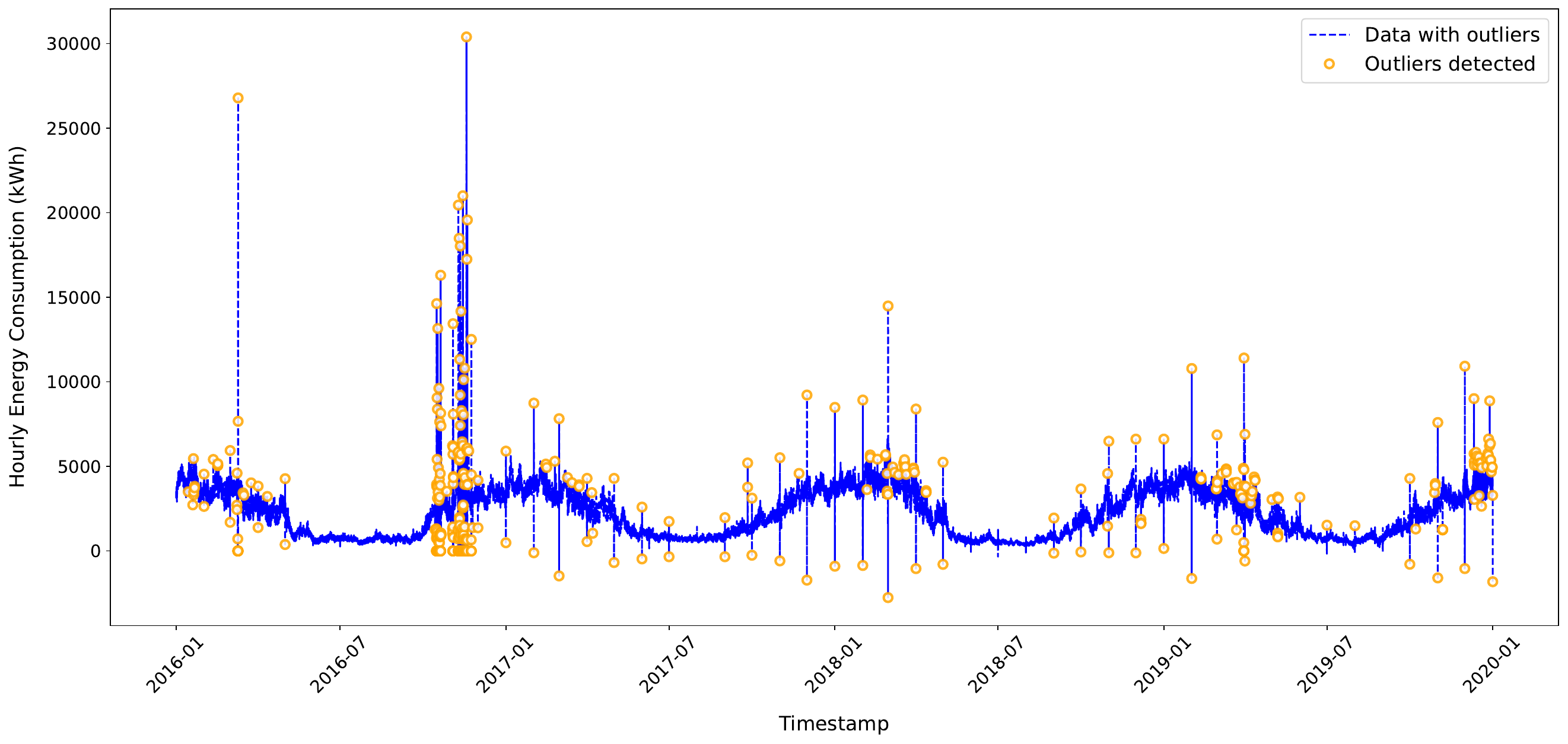}
\caption{Hourly heat consumption data over the period from January 2016 to the end of December 2019 for DMA A. The outliers observed in the data identified through the Savitzky-Golay smoothing are marked using orange circles.}
\label{fig:rawdata_dmaA}
\end{figure}

The heat demand forecasting framework presented herein is primarily developed and evaluated using data from a Danish utility company supplying hot water and space heating via its \acrshort{dhs} to the city of Brønderslev. This dataset comprises retrospective hourly heat consumption from 4910 smart meters across three \acrshort{dmas} of \textit{\acrshort{dma} A, \acrshort{dma} B}, and \textit{\acrshort{dma} C}, spanning June 2015 to November 2020. As raw smart meter readings are cumulative in nature, a first-order difference is applied to derive hourly consumption. These hourly consumption values from individual smart meters are then aggregated for each district to represent its total demand at any given time $t$. The aggregated time series data for \textit{\acrshort{dma} A} with periodic and non-periodic outliers are visualised in Figure \ref{fig:rawdata_dmaA}.

The heat demand data acts as the endogenous variable, while external, independent variables such as those of weather features and calendrical features form the set of exogenous variables that influence heat demand. Instead of forecasted weather data, actual weather data for the retrospective period is obtained and used. Explicit socio-economic or demographic features are not modelled, as these are represented as intrinsic patterns within the recorded heat demand data. Data recorded between June 2015 and 2016, which falls within the digital maturity phase, has been excluded. Over the duration of 5 years, as the number of smart meters capturing the demand of individual customers present in the \acrshort{dma} is not constant, the aggregated demand at time $t$ is averaged by the number of smart meters contributing to the total demand. This addresses the presence of data drifts. As the number of smart meters is known, the demand time series can be rescaled to its original scale after the forecasting process.  Additionally, data from 2020 has been discarded to maintain behavioural consistency and eliminate the influence of COVID-19-related behaviours.

To ensure the framework's robustness and facilitate reproducibility on public benchmarks, the evaluation is extended to include two open-source datasets from distinct geographic locations. First, we utilize the hourly district heating production data from Flensburg, Germany \cite{stadtwerke_flensburg_gmbh_2019_2562658}, spanning from January 2017 to December 2024. Historical data prior to 2017 was excluded due to inconsistencies in recording precision. For this dataset, historical weather variables, specifically ambient temperature and 'feels-like' temperature were retrieved from the ERA5 reanalysis dataset via the Copernicus Climate Data Store \cite{hersbach2020era5}. Corresponding calendrical features were generated based on the public holidays of the state of Schleswig-Holstein.
Second, we employ residential dataset from Aalborg, Denmark \cite{aalborg}, which aggregates hourly consumption readings from 3,021 residential smart meters. Meteorological data for this region, including ambient temperature, minimum temperature, and wind speed, were sourced from the Danish Meteorological Institute (DMI) Open Data API \cite{dmi_api}. The feels-like temperature was explicitly computed using the standard wind chill formula utilized by DMI \cite{windchill}.

\subsection{Data Preprocessing}

\begin{figure}[htbp]
    \centering 

    \begin{subfigure}[b]{0.49\linewidth} 
        \centering

        \includegraphics[width=\linewidth]{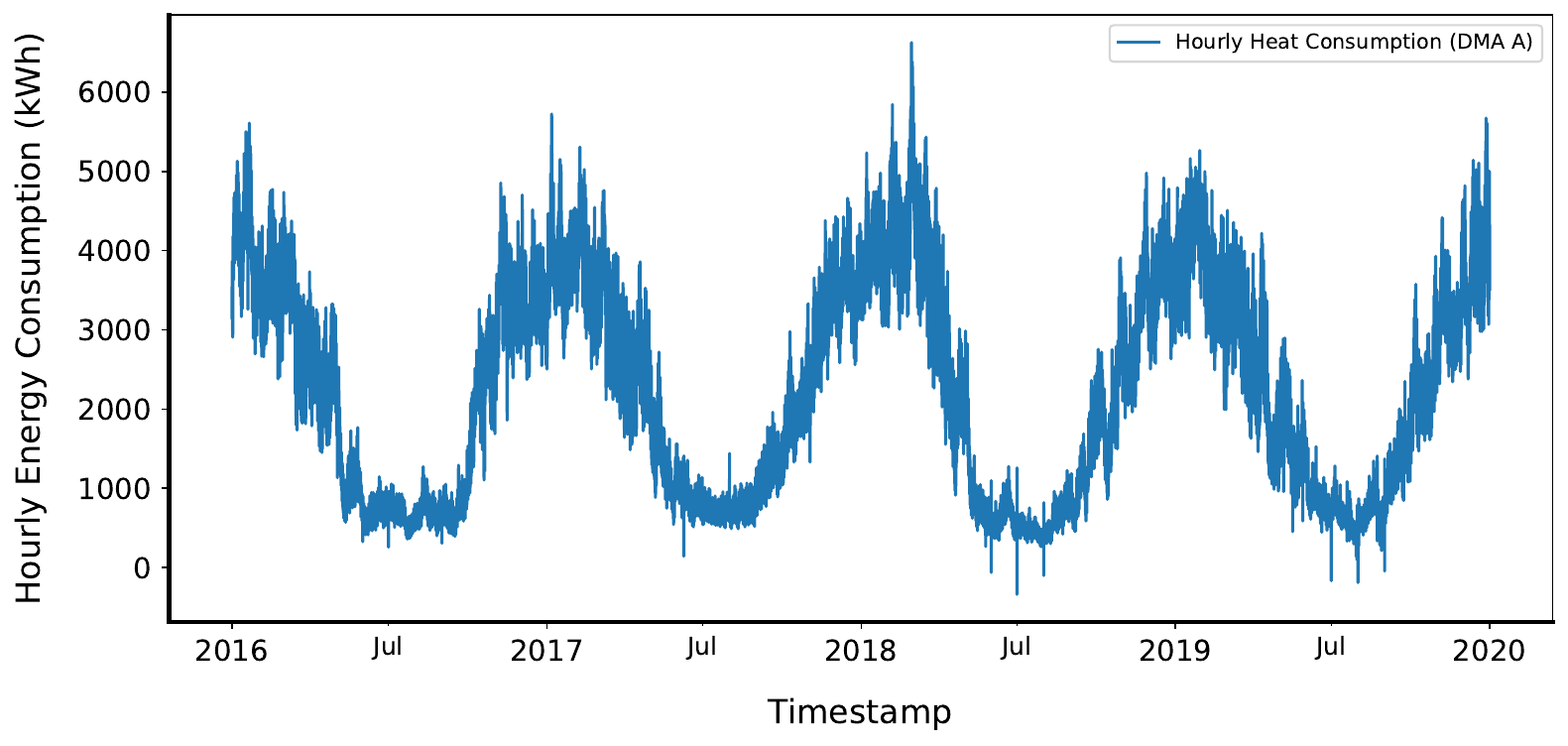}
        \caption{January 2016 - December 2019.}
        \label{fig:correcteddata_dmaA_fullperiod}
        
    \end{subfigure}
    \hfill
    \begin{subfigure}[b]{0.49\linewidth}
        \centering
        
        \includegraphics[width=\linewidth]{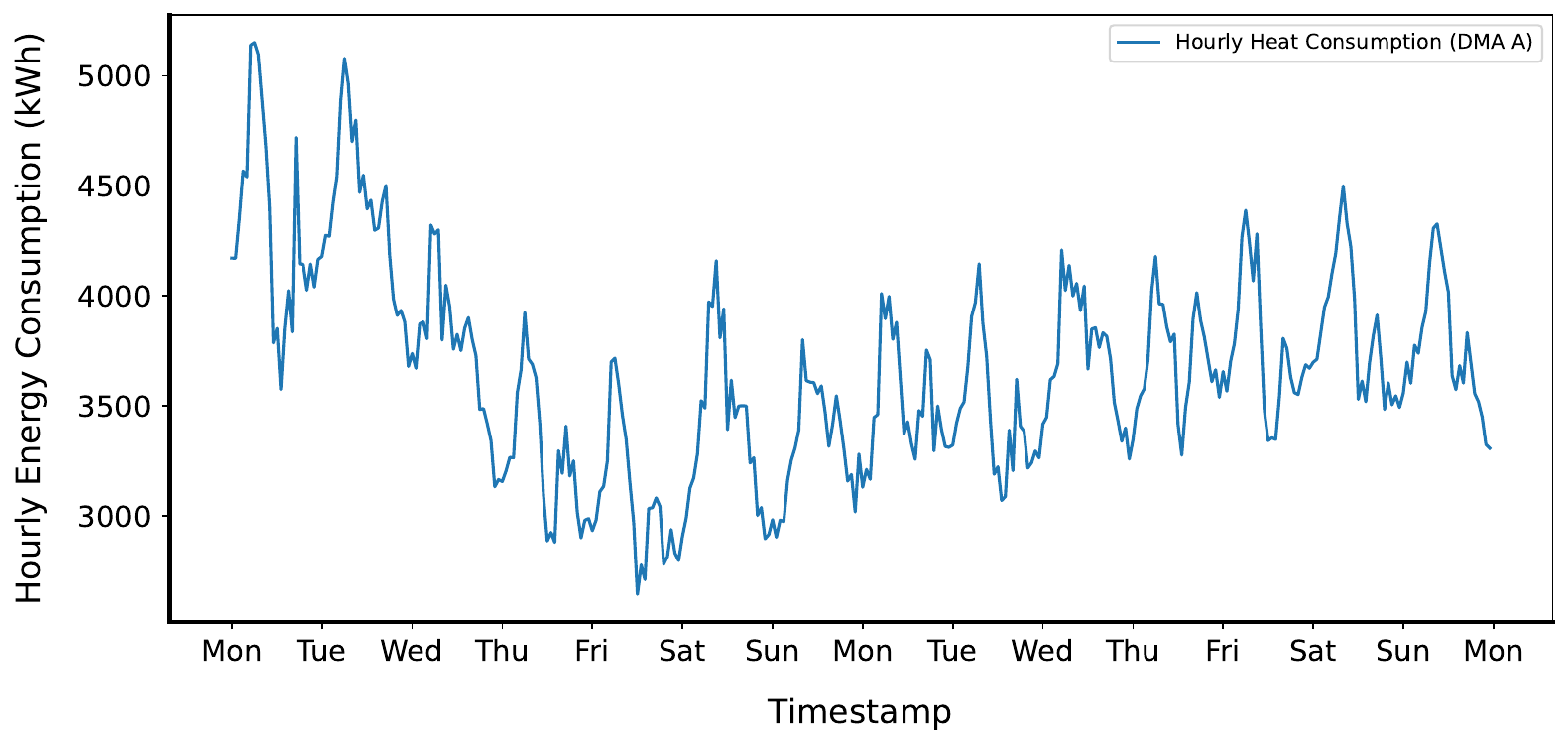}
        \caption{Two weeks in January 2017.}
        \label{fig:correcteddata_dmaA_2weeks_detail}
        
    \end{subfigure}

    \caption{Preprocessed hourly heat consumption data for DMA A. (a) Data from January 2016 to December 2019 after outlier removal. (b) A detailed view of the recorded heat consumption for two weeks in January 2017.}
    \label{fig:dma_A_processed_data_combined}
    
\end{figure}

As seen in Figure \ref{fig:rawdata_dmaA}, the recorded hourly heat consumption data exhibits a behaviour where the magnitude of outliers is seasonal, with larger deviations observed in winter and smaller deviations in summer. Furthermore, the data contains physically implausible negative consumption values. The non-linear, non-stationary nature of this time series, combined with outliers lacking a clear global characteristic, renders traditional outlier handling methods (e.g., Z-score thresholds, median-based approaches) ineffective. An alternative approach targeting the residual component of the time series, rather than its trend and seasonal components, is adopted. The outliers are addressed by applying the Savitzky-Golay smoothing to the studentized residuals, identifying outliers using a Bonferroni correction, and subsequently replacing them with the sum of the time series' trend and seasonal components. Figure \ref{fig:correcteddata_dmaA_fullperiod} illustrates the heat demand data for \textit{\acrshort{dma} A} after outlier processing for the entire data duration, showcasing a yearly periodicity, and Figure \ref{fig:correcteddata_dmaA_2weeks_detail} represents the heat consumption data over two weeks during January 2017, exhibiting non-linearity, non-stationarity, with daily, and weekly periodicity. The preprocessed heat demand data acts as the dependent response variable.

\subsection{Data Analysis}

\begin{figure}[htbp]
\centering
\includegraphics[width=\linewidth]{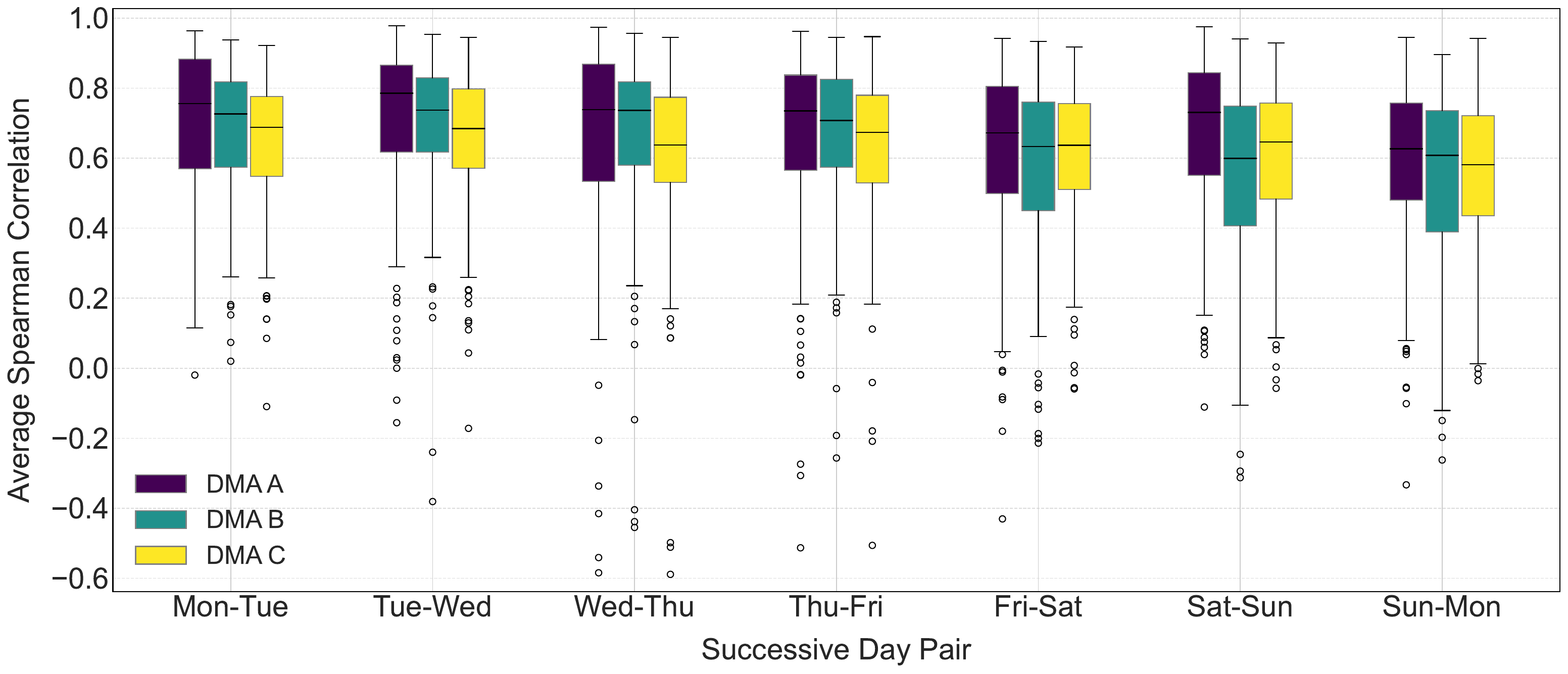}

\caption{The distribution of Spearman correlation coefficients $\rho$ between successive days of the week across different \acrshort{dma}.}
\label{fig:successive_correlation}
\end{figure}

Future heat demand is intrinsically linked to its historical behaviour, which encapsulates the fundamental nature and magnitude of consumption. The historical data also reflects underlying, non-deterministic influences such as demographic and psychographic factors, alongside distinct long-term and short-term periodicities. Illustrative examples of these long-term trends and short-term cyclical patterns are presented in Figure \ref{fig:correcteddata_dmaA_fullperiod} and Figure \ref{fig:correcteddata_dmaA_2weeks_detail}, respectively. To quantitatively assess the strength of these temporal dependencies, we perform an autocorrelation analysis using the Spearman correlation coefficient ($\rho$). This non-parametric measure is chosen for its ability to capture potentially non-linear monotonic relationships or piecewise linearities between the heat demand time series and its lagged versions, without assuming a linear association.
To investigate immediate temporal persistence, the Spearman correlation $\rho$ between heat demand on successive days (i.e., current day versus the next) is computed for all such pairs over the four years for each \acrshort{dma}. As depicted in Figure \ref{fig:successive_correlation}, this analysis reveals a consistently strong positive correlation, with mean values for different \acrshort{dma}s typically ranging between $0.8$ and $0.9$. This signifies a substantial influence of the current day's consumption on the subsequent day's demand. Furthermore, correlations between consecutive weekdays are observed to be marginally higher than those spanning weekday-to-weekend transitions, a difference likely attributable to variations in social and consumer behaviour patterns within the \acrshort{dmas}. It should also be noted that while strong positive correlations between successive day pairs are true for most pairs, there exist instances with weak positive correlations and also negative correlations. \acrshort{dma} A, \acrshort{dma} B, and \acrshort{dma} C have $25, 32, \text{ and } 17$ such negative correlation pairs from 2016, until 2020.

\begin{figure}[htbp]
    \centering 

    \begin{subfigure}[b]{0.41\linewidth} 
        \centering

\includegraphics[width=\linewidth]{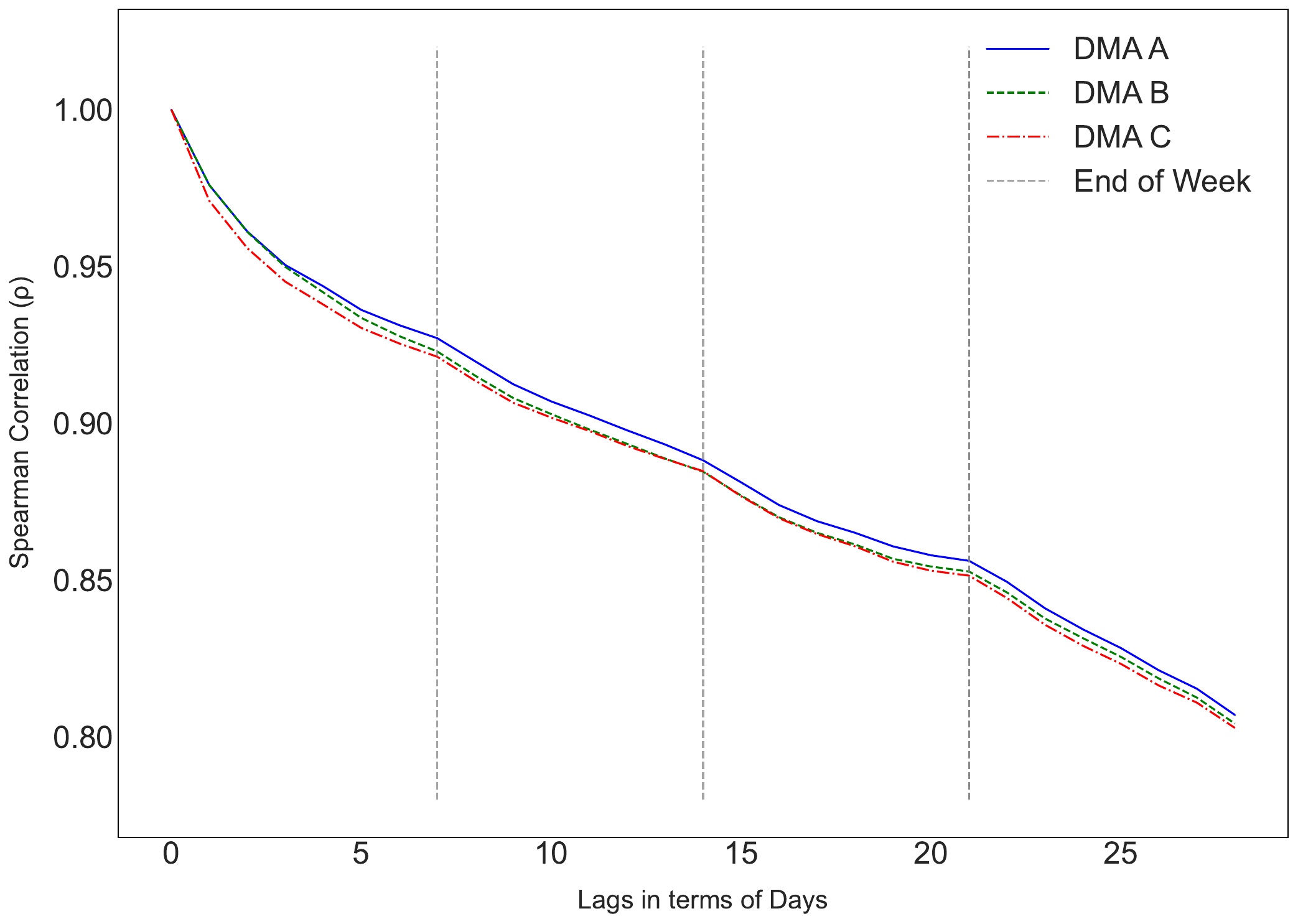}

\caption{4 week correlational analysis}
\label{fig:4week_correlation}
        
    \end{subfigure}
    \hfill
    \begin{subfigure}[b]{0.53\linewidth}
        \centering
        
\includegraphics[width=\linewidth]{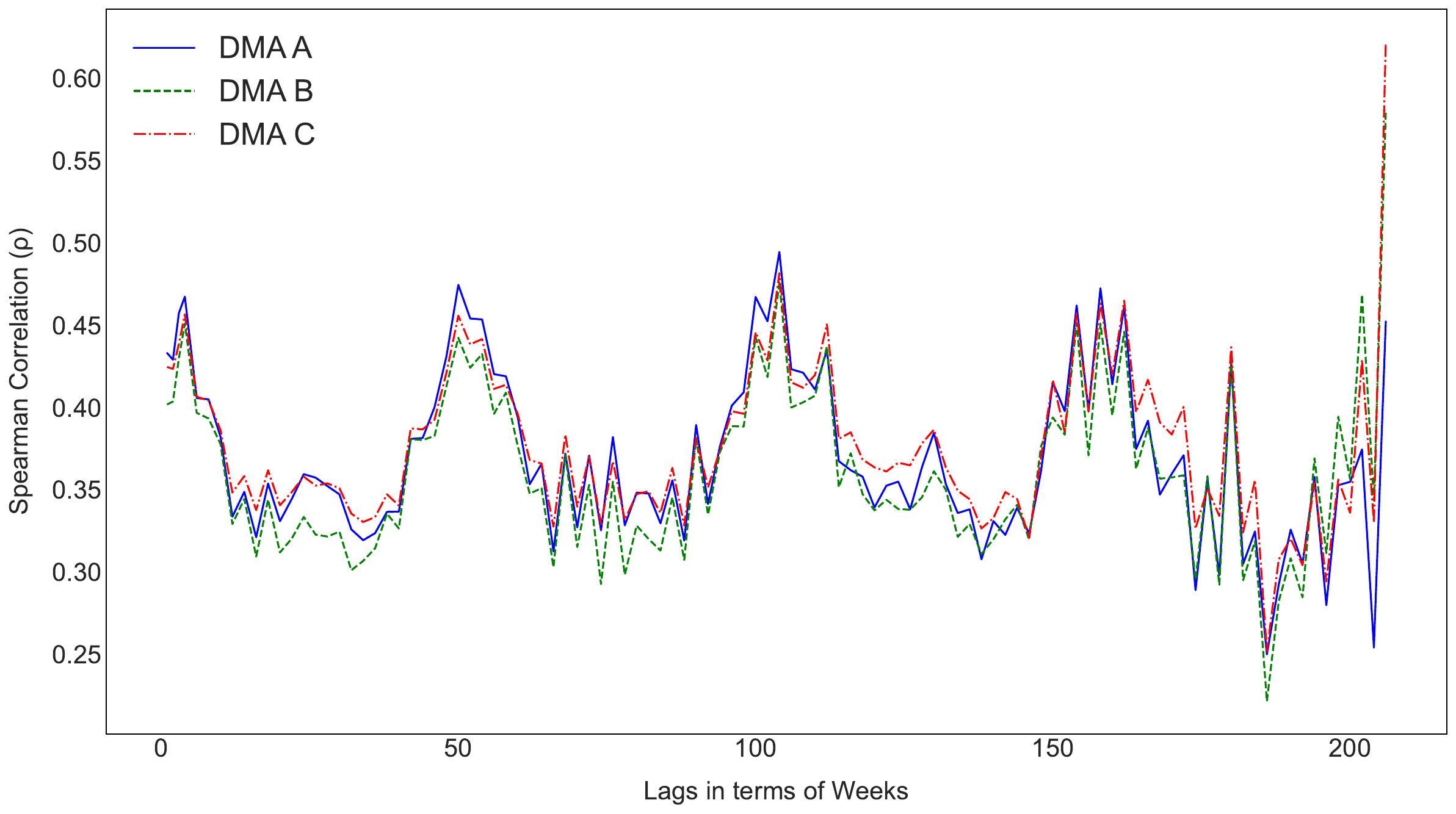}

\caption{204 week correlational analysis}
\label{fig:204week_correlations}
        
    \end{subfigure}

    \caption{Short-term and long-term correlational analysis of heat demand data. (a) Spearman correlation $\rho$ between heat demand data and the lagged version of the heat demand data. Hourly lags equivalent to $1, 2,...,28$ days are considered to evaluate short-term relationships. (b) Spearman correlation $\rho$ of weekly demand across 204 weeks comprised in the four-year heat demand dataset across different \acrshort{dmas}.}
    \label{fig:correlational analysis}
    
\end{figure}

Extending this analysis, we examine the relationship between current heat demand and its historical values at varying lags, specifically from $1$ to $28$ days (equivalent to $24, 48, ..., 672$ hours). Figure \ref{fig:4week_correlation} illustrates that while a strong correlation is evident, it diminishes as the lag increases. Slight periodic increases in correlation occur when the lag is a multiple of 7 days (168 hours), indicating weekly seasonality. This underscores the auto-regressive nature of the heat demand time series. Such insights are crucial for effective feature selection, particularly for \acrshort{ml} and \acrshort{dl} approaches of modelling the demand. Prioritizing features from the recent past (the previous day's demand) and those capturing weekly cycles (demand from the same day of the previous week), while strategically limiting the inclusion of identical features from historical data to avoid redundancy, increased computational costs, and over-representation of certain behaviour, is imperative.

Additionally, to explore long-term relational patterns, a correlational analysis is conducted on weekly demand of $168$ hours spanning the entire four-year dataset. Visualised in Figure \ref{fig:204week_correlations}, this broader perspective further highlights the presence of yearly periodicities and recurring cycles of stronger inter-week correlations, providing a comprehensive understanding of the multi-scale temporal dependencies within the heat demand data.
This exhibited dependency of the nature of demand for a lag of 24 hours and 168 hours is statistically demonstrated in Table \ref{tab:feature_correlations} through Spearman correlations.

 \aboverulesep=0ex
 \belowrulesep=0ex

\begin{table}[htbp] 
\caption{Available endogenous and exogenous features at an hourly resolution along with the respective Spearman correlation coefficient with demand.}
\label{tab:feature_correlations}

\newcolumntype{C}{>{\centering\arraybackslash}X}
\newcolumntype{L}{>{\raggedright\arraybackslash}X}

\renewcommand{\arraystretch}{1.2} 

\begin{tabularx}{\textwidth}{ 
    >{\hsize=0.7\hsize}C | 
    >{\hsize=1.6\hsize}L | 
    >{\hsize=0.9\hsize}C | 
    >{\hsize=0.9\hsize}C | 
    >{\hsize=0.9\hsize}C   
}
\toprule 
\multirow{2}{*}{\textbf{Feature Type}} & \multirow{2}{*}{\textbf{Feature}} & \multicolumn{3}{c}{\textbf{Spearman Correlation Coefficient ($\rho$)}} \\
\cmidrule(l){3-5} 
&  &  DMA A &  DMA B & DMA C \\
\hline
\multirow{2}{*}{Consumption} & Lag of 24 hours ($\textbf{c}_{24}$) & 0.973 & 0.975 & 0.970 \\
&  Lag of 168 hours ($\textbf{c}_{168}$)  & 0.923  & 0.920 & 0.918 \\
\hline
\multirow{14}{*}{Weather} & Temperature	($\textbf{t}_{\text{amb}}$) & -0.907 & -0.914 & -0.911 \\
 & Minimum temperature ($\textbf{t}_{\text{min}}$) & -0.902 & -0.910 & -0.907 \\
 & Maximum temperature ($\textbf{t}_{\text{max}}$) & -0.921 & -0.927 & -0.924 \\
 & Feels-like temperature ($\textbf{t}_{\text{feels}}$) & -0.915 & -0.922 & -0.919 \\
 & Rainfall (1 hour, 3 hours) & -0.046/0.021 & -0.050/0.017 & -0.051/0.016 \\
 & Snow (1 hour, 3 hours) & 0.130/0.106 & 0.131/0.106 & 0.130/0.106 \\
 & Wind speed & 0.121 & 0.123 & 0.116 \\
 & Wind degree & -0.114 & -0.115 & -0.112 \\
 & Wind gust & 0.065 & 0.064 & 0.064 \\
 & Humidity & 0.263 & 0.255 & 0.255 \\
 & Dew point ($\textbf{t}_{\text{dew}}$) & -0.847 & -0.857 & -0.854 \\
 & Pressure & -0.073 & -0.062 & -0.062 \\
 & Cloud cover & 0.223 & 0.213 & 0.210 \\
\hline
\multirow{2}{*}{Time} & Hour of day   & \multicolumn{3}{c}{\multirow{2}{*}{Cyclically encoded to sine-cosine pair with respective periodicity}} \\  
                      & Day of week   & \multicolumn{1}{c}{} & \multicolumn{1}{c}{} & \multicolumn{1}{c}{} \\ 
\hline
\bottomrule
\end{tabularx}
\end{table}

Beyond the auto-regressive nature of heat demand, exogenous factors, particularly meteorological conditions, are well-known to influence heat energy consumption significantly. To quantify these relationships within our dataset, an initial assessment is conducted by calculating the Spearman correlation coefficient $\rho$ between the hourly heat demand and a comprehensive set of weather variables. Table \ref{tab:feature_correlations} summarizes these correlations across the three \acrshort{dma}s. We denote key features using a formal notation, with historical consumption lagged by $h$ hours as $\textbf{c}_{h}$. We denote weather features as $\textbf{t}_{x}$ with specific weather features of ambient, minimum, maximum, feels-like, and dew point temperatures taking the form $\textbf{t}_{\text{amb}}$, $\textbf{t}_{\text{min}}$, $\textbf{t}_{\text{max}}$, $\textbf{t}_{\text{feels}}$, and $\textbf{t}_{\text{dew}}$ respectively.

As evident from Table \ref{tab:feature_correlations}, temperature-related variables (ambient temperature, minimum/maximum temperature, and feels-like temperature) exhibit a strong and consistent negative correlation with heat demand across all \acrshort{dma}s, with $\rho$ values typically below -0.9. The dew point $\textbf{t}_{\text{dew}}$ also shows a strong negative correlation (around -0.85). Conversely, other weather variables such as rainfall, snow accumulation, wind speed, wind direction, wind gust, pressure, and cloud cover demonstrate considerably weaker correlations with hourly heat demand.  

Given the prominent influence of temperature-related variables and dew point, a more nuanced analysis is performed to understand how these key weather features correlate with heat demand not just in their raw form, but also across their different underlying temporal components. To achieve this, both the heat demand time series and the time series of these selected key weather features ($\textbf{t}_{\text{amb}}$, $\textbf{t}_{\text{min}}$, $\textbf{t}_{\text{max}}$, $\textbf{t}_{\text{feels}}$, $\textbf{t}_{\text{dew}}$) are subjected to additive seasonal decomposition to derive their trend, seasonal, and residual components. Subsequently, Spearman correlations $\rho$ are computed between the corresponding components of heat demand and each key weather feature.

\begin{figure}[htbp]
    \centering 

    \begin{subfigure}[b]{\linewidth}
        \centering

        \includegraphics[width=\linewidth]{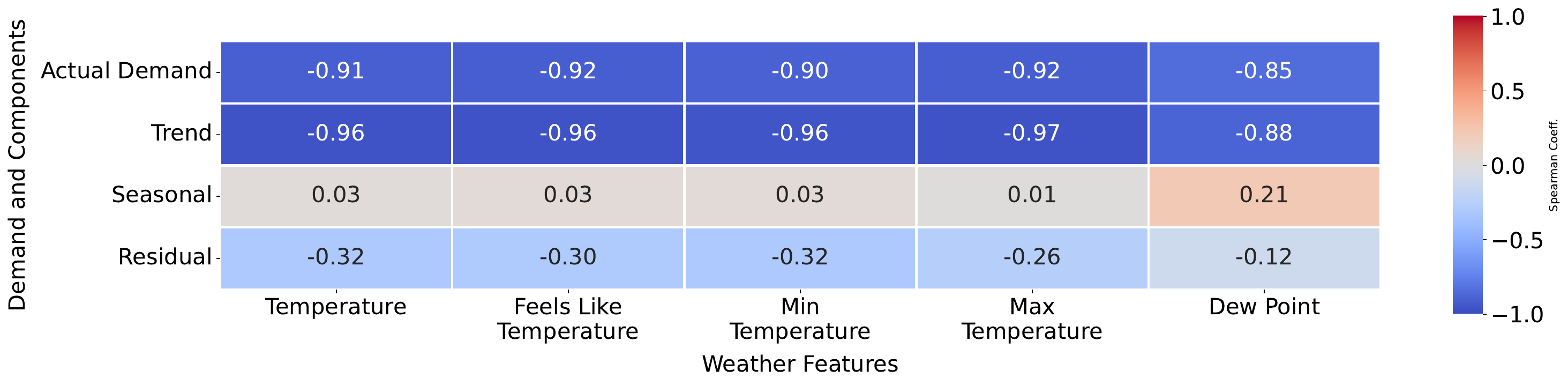}
        \caption{Calculated Spearman Correlation between the heat demand and weather variables for \acrshort{dma} A.}
        \label{fig:dma_a_weather}
        
    \end{subfigure}
    
    \hfill
    
    \begin{subfigure}[b]{\linewidth}
        \centering
        
        \includegraphics[width=\linewidth]{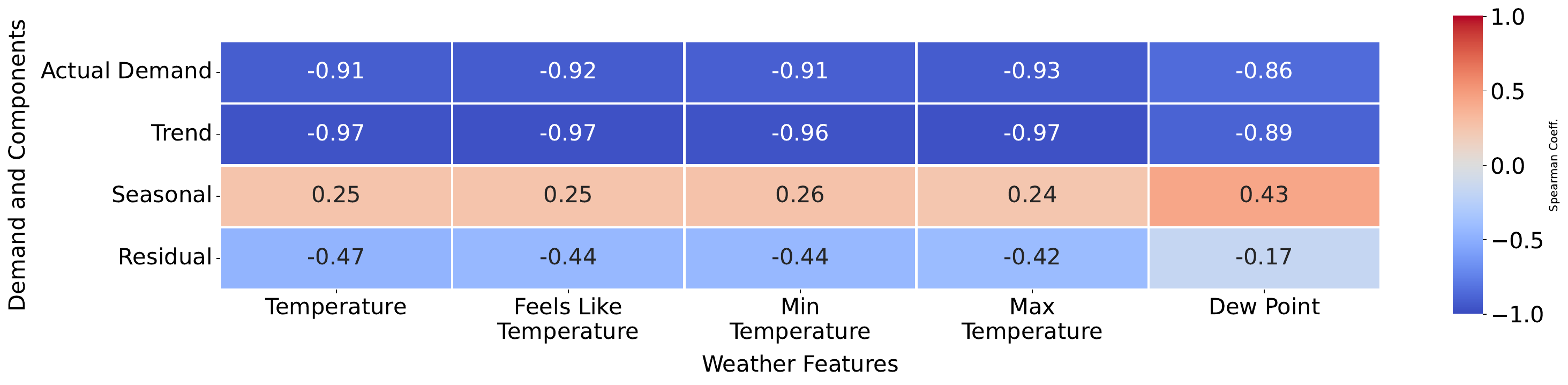}
        \caption{Calculated Spearman Correlation between the heat demand and weather variables for \acrshort{dma} B.}
        \label{fig:dma_b_weather}
        
    \end{subfigure}
    
    \hfill
    
    \begin{subfigure}[b]{\linewidth}
        \centering
        
        \includegraphics[width=\linewidth]{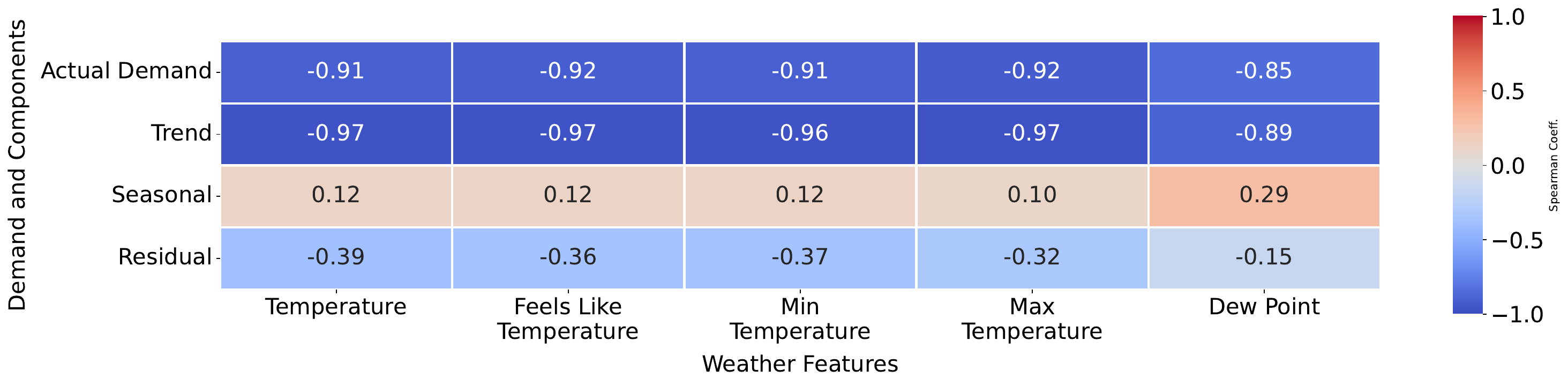}
        \caption{Calculated Spearman Correlation between the heat demand and weather variables for \acrshort{dma} C.}
        \label{fig:dma_c_weather}
        
    \end{subfigure}

    \caption{Spearman correlation coefficients between heat demand and key weather features across different \acrshort{dma}s. The heatmap(s) are structured in four rows, displaying correlations between: (1) actual heat demand and actual weather data; (2) their respective trend components; (3) their respective seasonal components; and (4) their respective residual components.}
    \label{fig:demand_weather_correlations}
    
\end{figure}

These multi-component correlations are visualised in Figure \ref{fig:demand_weather_correlations}. The heatmaps in Figure \ref{fig:demand_weather_correlations} are structured in four rows, displaying correlations between the actual heat demand and actual weather data, their respective trend components, their respective seasonal components, and their respective residual components in that order.
Analysis of Figure \ref{fig:demand_weather_correlations} reveals several key patterns in the relationship between heat demand and selected weather variables across their different temporal components. The first row, representing the correlation between actual heat demand and actual weather data, are the initial findings presented in Table \ref{tab:feature_correlations}.
Examination of the trend components (second row) of the heatmap indicates an exceptionally strong negative correlation between the long-term trend of heat demand and the trends of temperature-related weather variables, including dew point. This robust relationship signifies that underlying, long-term shifts in heat consumption are closely mirrored by corresponding long-term variations in the trend of the key meteorological indicators.
The correlations involving seasonal components (third row) present a more varied picture. For \acrshort{dma} A and \acrshort{dma} C, the seasonal components of various temperature metrics show negligible correlation with the seasonal component of heat demand. In \acrshort{dma} B, a weak correlation is observed for the seasonal components. The seasonal component of dew point, however, distinguishes itself from temperature by exhibiting weak to moderate correlations with seasonal heat demand across the \acrshort{dma}s.
Finally, the correlations between the residual components (fourth row) are generally diminished compared to the trend or actual data correlations. For the temperature-related weather variables, the residual correlations, while still negative, are significantly attenuated, falling into the weak to moderate correlation range. This implies that after accounting for trend, the coupling between the remaining short-term, irregular fluctuations in heat demand and these weather variables is less pronounced, although some relationship persists.

\begin{figure}[htbp]
\centering
\includegraphics[width=1.0\linewidth]{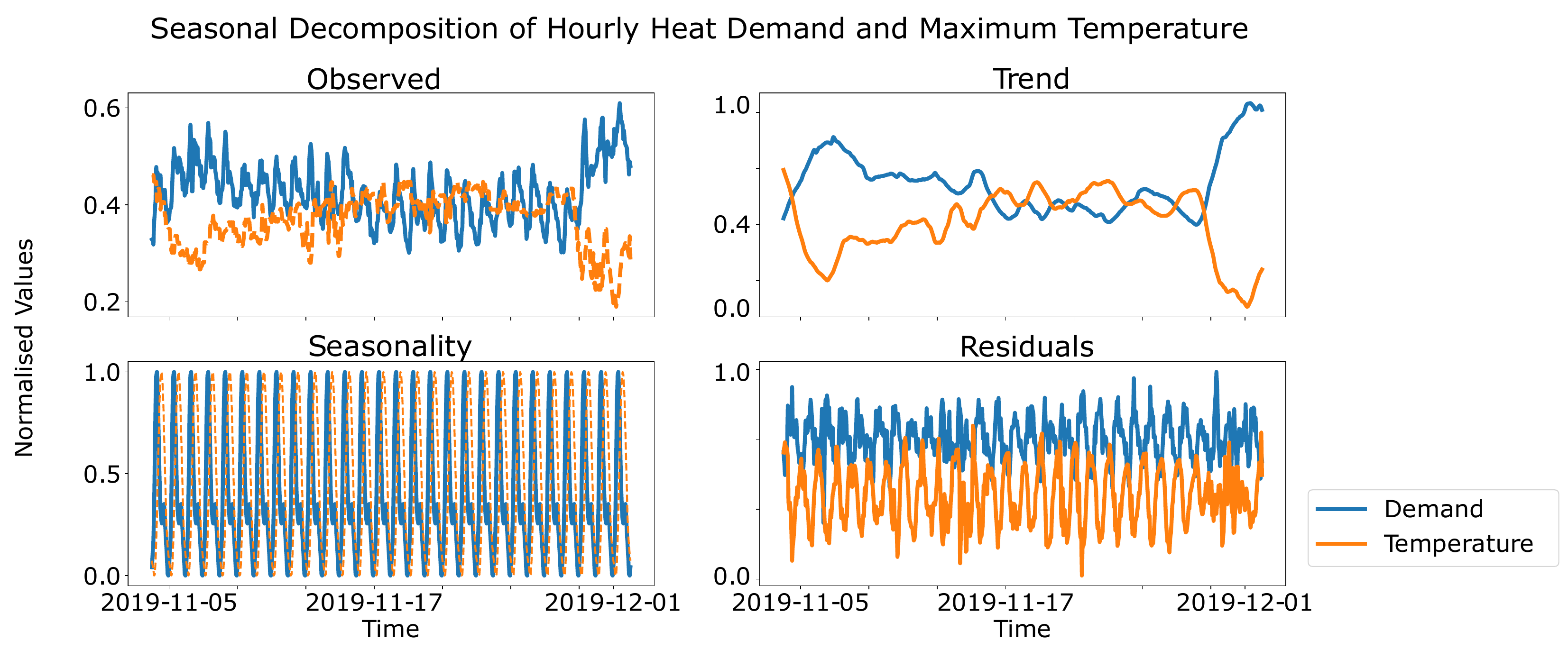}
\caption{Seasonal decomposed demand data and maximum temperature sampled hourly for a specific DMA B for a period of one month showcasing the explicit relationship between respective components of demand and exogenous feature.}
\label{fig:SD_qualitative}
\end{figure}

A qualitative illustration of these decomposed relationships is provided in Figure \ref{fig:SD_qualitative}, which showcases the interplay between heat demand and maximum temperature across their actual and decomposed components over a one-month period. This visualization highlights how the seasonal component of heat demand primarily reflects its own inherent historical seasonality. Concurrently, the inverse relationship between the respective trend components of heat demand and maximum temperature is clearly discernible. Furthermore, such a decomposed visualization can offer insights into the varying nature and consistency of the temperature's influence on demand across different timescales and components.

Additionally, calendrical features also affect heat demand. The inherent cyclical nature of consumption, demonstrated by analyses in Figure \ref{fig:204week_correlations} and Figure \ref{fig:4week_correlation}, necessitates appropriate representation of temporal indicators. Consequently, features like time of day and day of week are transformed into sine-cosine pairs to effectively model their respective daily, weekly, and yearly periodic patterns. To further enhance the feature set, specific indicators for public holidays and other special occasions are included to capture their distinct, though often infrequent, impact on demand. A distinction is maintained between these explicit holiday features and regular weekend patterns, as weekends are intrinsically represented within the cyclical encoding of the 'day of week'.

\subsection{Feature Representation}
The representation of input features for the forecasting framework begins with structuring the temporal data. 
For each input feature $k$ (as summarized in Table \ref{tab:feature_correlations}), by considering the historical values, an input feature vector is denoted as $\mathbf{s}_k(t) = [s_{t-N_{history}+1}, s_{t-N_{history}+2}, \ldots, s_t]$, comprising $N_{history}$ historical observations up to the current time $t$. 
This historical information, across a curated set of $N_{features}$ input variables, is leveraged to predict a target vector $\mathbf{y}(t) = [y_{t+1}, y_{t+2} \ldots, y_{t+N_{horizon}}]$, which contains $N_{horizon}$ future values of the target variable, which is the hourly heat demand.


Each of the $N_{features}$ input feature vectors is transformed from the time domain into a two-dimensional time-frequency representation known as a wavelet scalogram through the \acrshort{cwt} approach \cite{cwt}.
A scalogram illustrates the frequency components inherent in a signal and, critically, localizes these components in time. 
The \acrshort{cwt} operates by convolving the input signal with a mother wavelet, $\psi_{a, \tau}(\mathbf{s}_k)$, which is a prototype function systematically scaled by a factor $a$ and translated by $\tau$.
The resulting scalogram for the feature $k$, denoted as $\mathbf{S}_k$, has dimensions $M \times N_{\text{history}}$, where $M$ represents the number of discrete scales used in the wavelet analysis.
The \acrshort{cwt} effectively maps the temporal dynamics of $\mathbf{s}_k$ into a matrix where rows correspond to frequency scales and columns correspond to time steps within the $N_{\text{history}}$ window.
Finally, the $N_{\text{features}}$ individual scalograms, ${\mathbf{S}_1, \mathbf{S}_2, \dots, \mathbf{S}_{N_{\text{features}}}}$, are concatenated along a third dimension to form a single three-dimensional input tensor, $\mathbf{X}$.
This input tensor, $\mathbf{X} \in \mathbb{R}^{M \times N_{\text{history}} \times N_{\text{features}}}$, serves as the comprehensive, image-like representation capturing the time-frequency components at a given instance.
The model is tasked with learning a mapping from $\mathbf{X}$ to predict a target vector $\mathbf{y}(t) = [y_{t+1}, y_{t+2} \ldots, y_{t+N_{horizon}}]$.

The \acrshort{cwt} is selected over the \acrfull{dwt} for its potential to provide high-fidelity time-frequency representations \cite{Arts2022}. The efficacy of the wavelet transform is highly dependent on the choice of the mother wavelet $\psi(\cdot)$, as different signals possess unique intrinsic characteristics. Therefore, for each of the $N_{\text{features}}$ input time series, an optimal mother wavelet is quantitatively selected.

\subsection{Forecasting Model}

\begin{figure}[htbp]
\centering
\includegraphics[width=1.0\linewidth]{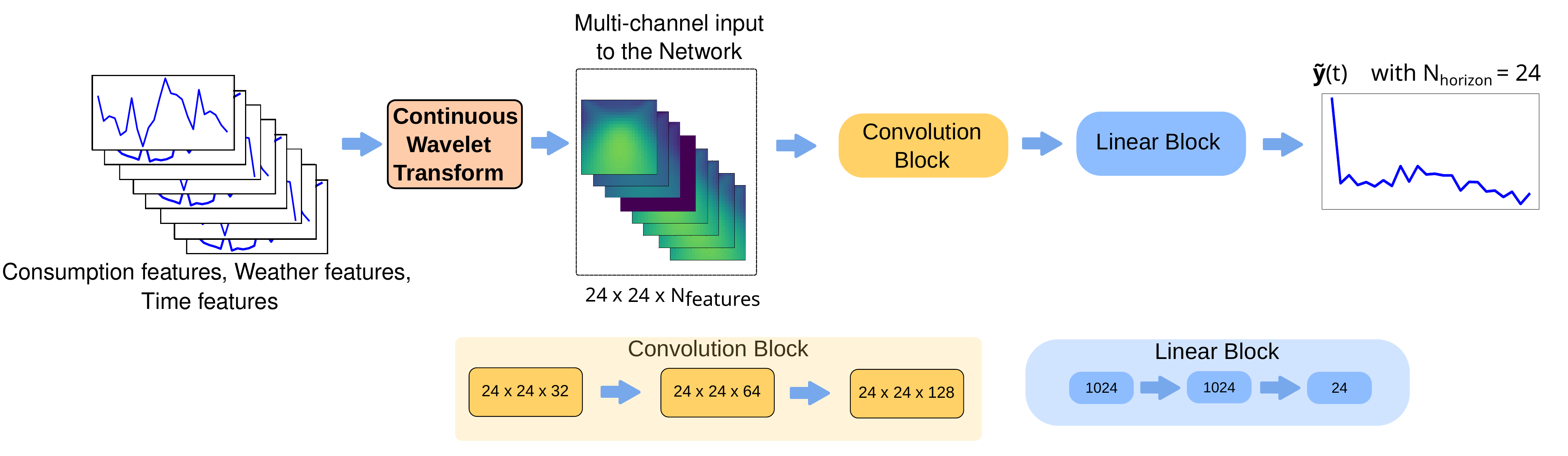}
\caption{Wavelet CNN forecasting network with image-like multi-channel representation of scalograms as inputs for day ahead hourly demand forecasting.}
\label{fig:block_diagram}
\end{figure}

Following the transformation of the input time series into the time-frequency domain tensor $\mathbf{X} \in \mathbb{R}^{M \times N_{\text{history}} \times N_{\text{features}}}$, a \acrshort{cnn} based architecture is employed to learn the complex patterns and map them to the desired forecast horizon. The architecture of the proposed model, depicted in Figure \ref{fig:block_diagram}, is specifically designed to interpret the image-like scalogram data for a multi-step-ahead regression task. The overall model represents a mapping function $F_{\text{model}}$ with learnable parameters $\theta$, such that:

\begin{equation}
\mathbf{\hat{y}}(t) = F_{\text{model}}(\mathbf{X}; \theta)
\end{equation}

The architectural choices are directly informed by the characteristics of the decomposed heat demand data and its relationship with exogenous features. Since the \acrshort{cwt} explicitly localizes frequency components in time, the use of 2D convolutional kernels allows the model to capture local temporal-periodic patterns within each feature. Crucially, by stacking scalograms depth-wise, the kernels simultaneously learn cross-feature interactions, such as the coupling between temperature and demand at time $t$.
Pooling layers introduce spatial downsampling, effectively reducing temporal resolution and spectral resolution \cite{pooling}. In \acrshort{dhs}, accurately predicting the timing of the morning peak is as critical as predicting its magnitude. Therefore, to reduce spatial invariance, we exclude pooling operations, preserving high-fidelity time-frequency localization, allowing the network to maintain phase coherence between the input features and the forecast vector for preserving the relationship between time of day and demand patterns due to spatial equivariance from convolution.
Consequently, we retain the full spatial resolution through the convolutional stack and rely on a high-dimensional flattening operation that allows subsequent dense layers to capture the most important dependencies as opposed to weak and noisy patterns. To mitigate the risk of overfitting inherent in this high-dimensional representation, we introduce a Dropout layer between the linear layers, ensuring the model learns robust, generalizable features rather than memorizing noise.

The initial module, which is a stack of $L$ 2D convolutional layers, is responsible for feature extraction from the input tensor $\mathbf{X}$. Its purpose is to identify and learn hierarchical patterns within the scalograms, which correspond to specific significant time-frequency events in the original signals. 
The input to the model $\mathbf{H}^{(0)} = \mathbf{X}$ goes through $l$-th convolutional layer with set of $C^{(l)}$ filters applied across the input feature maps from the previous layer, $\mathbf{H}^{(l-1)}$. Each filter performs a convolution operation to produce an output feature map, and a \acrfull{relu} activation function is applied element-wise to introduce non-linearity. The operation at each convolutional layer $l$ can be represented as:

\begin{equation}
\mathbf{H}^{(l)} = \text{ReLU}(\text{Conv2D}^{(l)}(\mathbf{H}^{(l-1)}))
\end{equation}

where $\mathbf{H}^{(l)} \in \mathbb{R}^{M \times N_{\text{history}} \times C^{(l)}}$ is the tensor of output feature maps.


After the final convolutional layer, the resulting three-dimensional feature map tensor $\mathbf{H}^{(L)}$ is flattened into a one-dimensional vector, $\mathbf{v}_{\text{flat}}$, to serve as the input for the linear regression heads.

\begin{equation}
\mathbf{v}_{\text{flat}} = \text{Flatten}(\mathbf{H}^{(L)})
\end{equation}

The dimensionality of this vector is $M \cdot N_{\text{history}} \cdot C^{(L)}$. This vector is then processed by a \acrshort{mlp}, which performs the final regression. The \acrshort{mlp} consists of a series of fully connected dense layers. Each dense layer computes a weighted sum of its inputs and applies an activation function. The transformation for a dense layer $j$ is given by:

\begin{equation}
\mathbf{d}^{(j)} = \sigma(\mathbf{W}^{(j)}\mathbf{d}^{(j-1)} + \mathbf{b}^{(j)})
\end{equation}

where $\mathbf{d}^{(j-1)}$ is the output from the previous layer (with $\mathbf{d}^{(0)} = \mathbf{v}_{\text{flat}}$), $\mathbf{W}^{(j)}$ and $\mathbf{b}^{(j)}$ are the weight matrix and bias vector, and $\sigma$ represents the Leaky ReLU activation function. The regression head projects $\mathbf{v}_{\text{flat}}$ to dense layers with $d_{neurons}$ neurons respectively, followed by the final prediction layer with $N_{horizon}$ to predict the heat demand $\mathbf{\Tilde{y}}(t)$.

\subsection{Experimental Setup}
\label{sec:experimental_setup}

\subsubsection{Dataset}
The study utilized a historical dataset of district heating consumption and associated features spanning four years, from January 1, 2016, to December 31, 2019, as mentioned in Section \ref{sec:data_overview}. The model is configured to forecast the subsequent $N_{\text{horizon}} = 24$ hours of heat demand based on a historical context window of $N_{\text{history}} = 24$ hours of historical data from $N_{\text{features}}$ input variables. A chronological data split is employed to simulate a realistic forecasting scenario, where the final 52 weeks (364 days) are designated as the held-out test set. 
The year-long test ensures adequate representation of different seasons, psychographic behaviours such as vacation periods or schooling periods, and at least one instance of every holiday, thus making the evaluation robust and comprehensive.
The preceding data is further divided chronologically, with 80\% allocated for training and the subsequent 20\% for the validation dataset. During training, the model iterates through temporal chronology in batches. This partitioning scheme yielded 872 training, 219 validation, and 364 test samples, each representing a 24-hour forecast window based on the input features, each with a context length of 24 hours.

To validate generalisability, this evaluation protocol is replicated for the public datasets with identical model configurations ($N_{\text{horizon}}=24, N_{\text{history}}=24$). The Flensburg dataset \cite{stadtwerke_flensburg_gmbh_2019_2562658}, which spans 8 years from January 1, 2017, to December 31, 2024, serving for a multi-year performance assessment of demand forecasting of a German city. The Aalborg dataset \cite{aalborg}, spanning $\approx3$ years from January 2, 2018, to December 31, 2020, represents residential heat consumption of 3021 residential units. For both public datasets, the remaining historical data preceding the test set adheres to the same chronological 80/20 split for training and validation, ensuring that no data leakage occurs across any experimental configuration.

\subsubsection{Architecture}

The core of the framework consists of a stack of three two-dimensional convolutional layers with 32, 64, and 128 filters, respectively. All convolutional layers utilize a $3 \times 3$ kernel, a stride of 1, and same padding to preserve the spatial dimensions of the input tensor. The linear layer comprises two hidden dense layers, each with 1024 neurons, activated by the Leaky ReLU function with a slope of $0.3$ to ensure robust gradient flow, followed by a Dropout layer with a probability of $0.1$. The final layer is a dense output layer with 24 neurons to produce the final forecast. Neither Pooling nor L2 regularization (weight decay) is employed in the final model configuration. The hyperparameter search process and ablation studies regarding pooling, dropout, and regularization are documented in the Appendix. The summary of the model is presented in Table \ref{tab:model_architecture}.

\begin{table}[h!]
\centering
\caption{Detailed architecture of the proposed CNN model with a total of $76, 669, 976 $ parameters. The input tensor dimensions correspond to (M $\times$ $N_{history}$ $\times$ $N_{features}$).}

\label{tab:model_architecture}
\begin{tabular}{l|c|c|c|c|c}
\hline 
\textbf{Layer} & \textbf{Operator} & \textbf{Output Dimension} & \textbf{Kernel} & \textbf{Stride} & \textbf{Padding} \\ \hline
Input          & -                 & 24 $\times$ 24 $\times$ $N_{\text{features}}$     & -               & -               & -                \\ \hline
Convolution 1         & Conv2D + ReLU     & 24 $\times$ 24 $\times$ 32    & 3 $\times$ 3    & 1               & 1 (Same)         \\ \hline
Convolution 2         & Conv2D + ReLU     & 24 $\times$ 24 $\times$ 64    & 3 $\times$ 3    & 1               & 1 (Same)         \\ \hline
Convolution 3         & Conv2D + ReLU     & 24 $\times$ 24 $\times$ 128   & 3 $\times$ 3    & 1               & 1 (Same)         \\ \hline
Flatten        & Flatten           & 73728                         & -               & -               & -                \\ \hline
Dense 1        & Dense + LeakyReLU & 1024                          & -               & -               & -                \\ \hline
Dense 2        & Dense + LeakyReLU & 1024                          & -               & -               & -                \\ \hline
Dropout        & Dropout ($p=0.1$) & 1024                          & -               & -               & -                \\ \hline
Output         & Dense             & 24                            & -               & -               & -                \\ \hline
\end{tabular}
\end{table}

The model's trainable parameters are optimized using the Adam algorithm with learning rate $\eta = 0.001$. A learning rate scheduler that decays the learning rate on a plateau by a factor of $0.9$ is also employed. The training objective is the minimization of the \acrfull{mse} loss function, which quantifies the squared difference between the predicted and actual demand values. The model is trained in batches of 32 samples for a maximum of 1000 epochs. To prevent overfitting and enhance model generalisation, an early stopping mechanism is integrated into the training loop. This mechanism monitored the \acrshort{mse} loss on the validation set and terminated the training if no improvement was observed for a patience of 50 consecutive epochs. Upon termination, the model weights corresponding to the epoch with the lowest validation loss are restored for the final evaluation on the test set. Ablation studies on the architecture, the hyperparameter search space, the hardware and the compute efficiency are documented in Appendix \ref{architecture_design_choice}.

\subsubsection{Evaluation Strategy}

The forecasting accuracy of the models is assessed using the two standard metrics of \acrfull{mae} and \acrfull{mape}. The \acrshort{mae} measures the average absolute deviation of the forecasts, while the \acrshort{mape} offers a scale-independent perspective by expressing the error as a percentage of the actual values. To assist in evaluating the effect of feature addition on outliers, \acrshort{mse} is also reported. These metrics are formally defined as:

\begin{align}
    \text{MAE} &= \frac{1}{n} \sum_{i=1}^{n} |y_i - \Tilde{y}_i| \\
    \text{MAPE} &= \frac{1}{n} \sum_{i=1}^{n} \left| \frac{y_i - \Tilde{y}_i}{y_i} \right|\\
    \text{MSE} &= \frac{1}{n} \sum_{i=1}^{n} \left( y_i - \Tilde{y}_i \right)^2
\end{align}

where $n$ is the total number of forecasted data points, $y_i$ represents the ground truth value, and $\Tilde{y}_i$ is its corresponding model prediction. Due to the cycling magnitude of demand between summer and winter, \acrshort{mape}, a relative error metric, allows for a viable comparison between predictions made during different seasons, while \acrshort{mae} provides a grounded comparison due to its absolute nature. We also keep in mind the instability of \acrshort{mape}, due to the possibility of dividing by zero. Together, these metrics characterize the hourly accuracy of the predictions against the observed demand.

\subsubsection{Experimental Overview}

A series of structured experiments is designed to systematically evaluate and optimize the proposed forecasting framework. The evaluation protocol is multifaceted, aiming to first determine the optimal feature composition and feature representation. For a rigorous benchmark, the finalised model is evaluated against a comprehensive suite of established methods. The experimental process begins with establishing a baseline model that relies solely on endogenous (autoregressive) demand features. This initial experiment (Experiment I) compares the predictive utility of a 24-hour lag ($\mathbf{c}_{24}$) against a combined feature set that also includes a weekly lag ($\mathbf{c}_{24}, \mathbf{c}_{168}$), with a specific focus on performance during periods of pattern transition as seen in Figure \ref{fig:successive_correlation}, and Figure \ref{fig:4week_correlation}. This foundational feature set will be denoted as $\mathcal{M}_C$.
The framework's performance is then evaluated in Experiment II by systematically incorporating exogenous meteorological variables. We evaluate the impact of adding key weather features identified through the initial data analysis, both individually and in various combinations, to the identified endogenous baseline $\mathcal{M}_C$ feature set. This process is conducted to determine the most informative and non-redundant set of weather features.

To further enhance the model's predictive capabilities, in Experiment III we investigate an advanced feature engineering strategy based on time series decomposition. This approach involves augmenting the raw time series features with their constituent components.

\begin{table}[htbp]
\centering
\caption{Definition of raw and decomposed feature sets.  The superscript $d$ in $\mathbf{c}^d_{24}$, $\mathbf{c}^d_{168}$, and $\mathbf{t}^d_{x}$ indicates the representations include decomposed components.}
\label{tab:decomposition_notation_centered}
\renewcommand{\arraystretch}{1.5} 
\newcolumntype{C}[1]{>{\centering\arraybackslash}p{#1}} 

\begin{tabular}{c|c|c|c}
\toprule
\textbf{Feature Type} & \textbf{Original Vector} & \textbf{Decomposed Components} & \textbf{Decomposed Vector} \\
\midrule
\multirow{2}{*}{Consumption} & $\mathbf{c}_{24} \in \mathbb{R}^{24 \times 1}$ & \multirow{2}{*}{Seasonal, Trend, Residual} & $\mathbf{c}^d_{24} = \left([\mathbf{c}_{24}] \bigcup  [\mathbf{c}^{trend}_{24}, \mathbf{c}^{seas}_{24}, \mathbf{c}^{resid}_{24}]\right) \in \mathbb{R}^{24 \times 4}$ \\
 & $\mathbf{c}_{168} \in \mathbb{R}^{24 \times 1}$ & & $\mathbf{c}^d_{168} =\left([\mathbf{c}_{168}] \bigcup  [\mathbf{c}^{trend}_{168}, \mathbf{c}^{seas}_{168}, \mathbf{c}^{resid}_{168}]\right) \in \mathbb{R}^{24 \times 4}$ \\
\hline
Weather & $\mathbf{t}_{x} \in \mathbb{R}^{24 \times 1}$ & Trend, Residual & $\mathbf{t}^d_{x} =\left([\mathbf{t}_{x}] \bigcup  [\mathbf{t}^{trend}_{x}, \mathbf{t}^{resid}_{x}]\right) \in \mathbb{R}^{24 \times 3}$ \\
\bottomrule
\end{tabular}
\end{table}

As seen in Table \ref{tab:decomposition_notation_centered}, consumption features are augmented with their seasonal, trend, and residual components, while weather-related features are augmented with only their trend and residual components. The seasonal component of weather features is ignored, as the inherent seasonality of heat demand is sufficiently represented by the seasonal components of its own historical features, an observation supported by the analysis in Figure \ref{fig:SD_qualitative}.  The efficacy of expanding the representation of each demand and weather is compared, and the best features and their representation are chosen. Subsequently in Experiment IV, we examine the effect of adding explicit cyclical calendrical features and evaluate multiple strategies for embedding information about public holidays. Such a step-by-step feature evaluation is essential because, in real-world forecasting, even marginal improvements in average accuracy can translate into better decisions and operational benefits. A careful curation of features helps reduce large errors on atypical or difficult-to-predict days, and also help understand the bounds in modelling the information at hand, and when mispredictions can occur due to lack of information. While simpler feature sets may perform adequately on the majority of cases, the ability to mitigate these extreme deviations justifies the depth of analysis, even when overall performance gains appear modest.

For a comprehensive comparative evaluation of the proposed framework within the current landscape of time series forecasting, as Experiment V, a benchmarking study is performed. The evaluation schema is designed to compare our approach against a wide spectrum of models, ranging from classical statistical methods to state-of-the-art deep learning architectures. The selected benchmarks include a traditional statistical model (\acrshort{sarimax}) \cite{sarimax}, a tree-based ensemble method (XGBoost) \cite{xgboost}, a canonical recurrent neural network (\acrshort{lstm}) \cite{lstm_orig}, and several modern Transformer-based models with encoder-only and encoder-decoder architectures renowned for long-sequence forecasting, namely Informer \cite{haoyietal-informer-2021}, Autoformer \cite{wu2021autoformer}, Preformer \cite{preformer}, PatchTST \cite{Yuqietal-2023-PatchTST}, and TimesNet \cite{wu2023timesnet}. Time series mixing models that have proven effective for time series forecasting, such as DLinear \cite{dlinear}, TSMixer\ \cite{tsmixer}, and TimeMixer \cite{timemixer} are also employed for benchmarking. Furthermore, to assess performance against the emerging paradigm of large pretrained foundation models, fine-tuned versions of \acrshort{ttm} \cite{ekambaram2024tinytimemixersttms} and Chronos-2 \cite{ansari2025chronos2} are included in the comparison. To ensure a fair comparison, a unified training protocol was enforced across all baselines. While architecture specific hyperparameters were individually optimized via grid search to ensure peak performance, the computational budget was standardized. All models were trained with a fixed batch size of 32, utilized the same ReduceLROnPlateau scheduler, and were subject to a maximum of 1000 epochs with an identical early-stopping patience of 50 epochs. The detailed hyperparameter search space and final configurations for each model are provided in Appendix \ref{benchmarks}.

Finally, in Experiment VI, we conduct a longitudinal performance analysis using a rolling block forecasting strategy. This experiment is designed to assess the temporal robustness of the proposed model and its ability to adapt to accumulating historical data over multiple years. In this setup, the model is initially trained on the first available year of data to forecast the subsequent year. The training is then expanded to the first two years and is evaluated in the third year. This iterative process continues until the final year of the dataset is predicted. This rigorous rolling evaluation is performed across all three datasets from Brønderslev, Flensburg, and Aalborg, providing a comprehensive view of the model's stability over long-term operational horizons.
\section{Results and Discussion}

\subsection{Feature Composition}


\subsubsection{Endogenous Features - Experiment I}

Our initial feature evaluation focuses on the impact of endogenous consumption features. We compare the models ability to forecast with the baseline using only the consumption lagged by 24-hours ($\textbf{c}_{24}$) as a feature against the incorporation of the weekly lag feature to the daily lag feature [$\textbf{c}_{24}$, $\textbf{c}_{168}$]. As shown in the aggregate results in Table \ref{tab:model_performance_consumption}, the addition of the $\textbf{c}_{168}$ feature offers only a slight improvement in metrics for DMA B \& DMA C but not for DMA A.

\begin{table}[ht]
\centering
\caption{Comparison of model performance (\acrshort{mae}, \acrshort{mape}, and \acrshort{mse}) for different endogenous features sets.}
\label{tab:model_performance_consumption}
\begin{tabular}{@{}l|l|c|c|c@{}}
\toprule
\textbf{\acrshort{dma}} & \textbf{Features} & \textbf{\acrshort{mae}} (kWh) & \textbf{\acrshort{mape}} (\%) & \textbf{\acrshort{mse}} ($\times 10^{4}$ kWh$^2$)\\
\hline
\multirow{2}{*}{\textit{\acrshort{dma} A}} 
 & $\textbf{c}_{24}$ & $160.78 \pm 114.56$ & $8.1 \pm 4.9$ & $5.65 \pm 9.09$ \\
 & $\textbf{c}_{24}, \textbf{c}_{168}$ & $165.98 \pm 119.89$ & $8.4 \pm 5.2$ & $5.95 \pm 9.35$ \\
\hline
\multirow{2}{*}{\textit{\acrshort{dma} B}} 
 & $\textbf{c}_{24}$ & $300.87 \pm 206.56$ & $7.9 \pm 4.6$ & $18.32 \pm 25.44$ \\
 & $\textbf{c}_{24}, \textbf{c}_{168}$ & $293.57 \pm 203.57$ & $7.7 \pm 4.6$ & $17.44 \pm 24.22$ \\
\hline
\multirow{2}{*}{\textit{\acrshort{dma} C}} 
 & $\textbf{c}_{24}$ & $264.69 \pm 172.27$ & $7.7 \pm 4.3$ & $13.74 \pm 18.52$ \\
 & $\textbf{c}_{24}, \textbf{c}_{168}$ & $260.04 \pm 172.59$ & $7.5 \pm 4.2$ & $13.43 \pm 18.64$ \\
\bottomrule
\end{tabular}
\end{table}

\begin{figure}[htbp]
    \centering 

    \begin{subfigure}[b]{0.3\linewidth}
        \centering
        \includegraphics[width=\linewidth]{"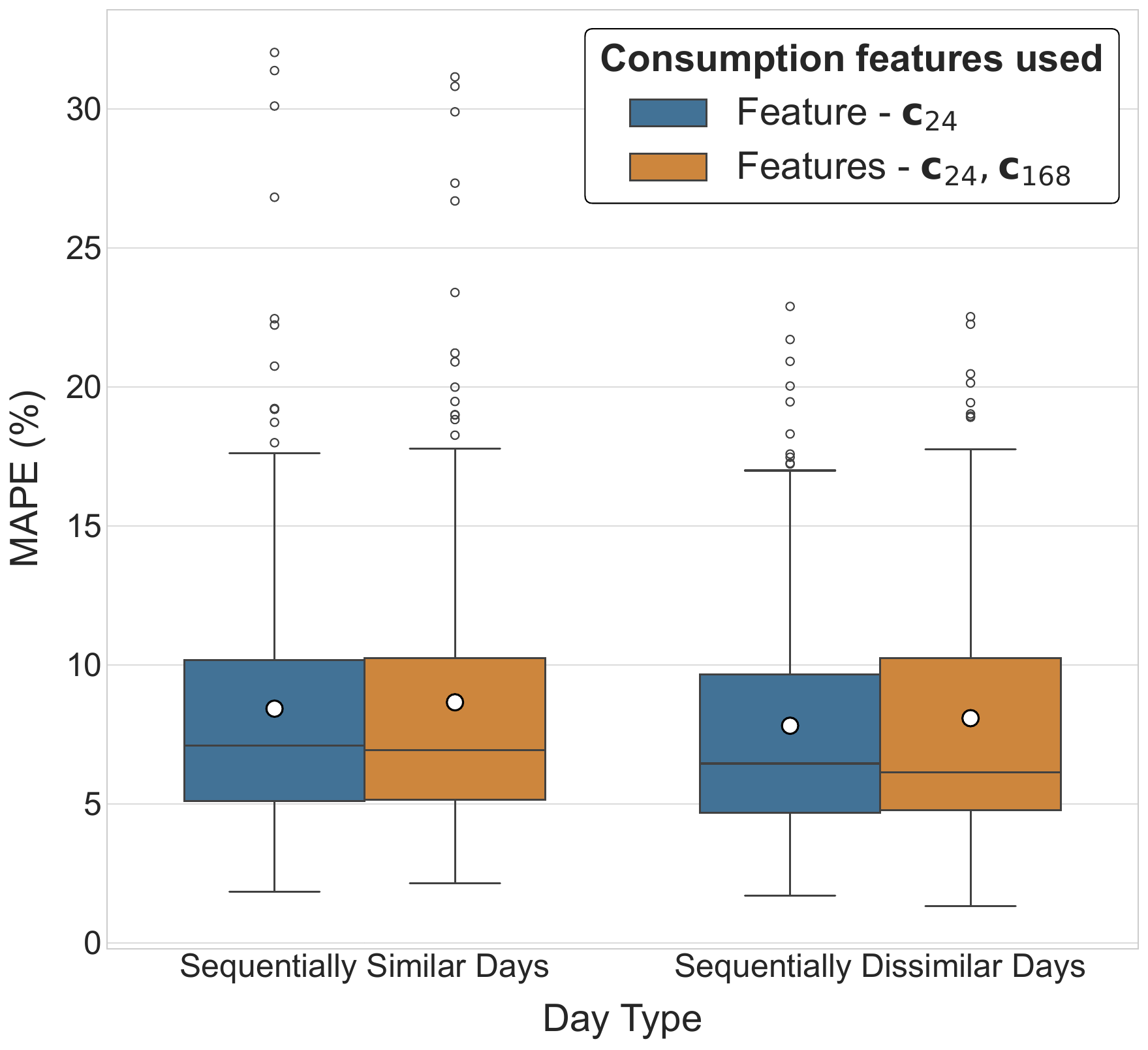"}
        \caption{\acrshort{dma} A}
        \label{fig:dma_a_cons_feature_analysis}
    \end{subfigure}    
    \begin{subfigure}[b]{0.3\linewidth}
        \centering
        \includegraphics[width=\linewidth]{"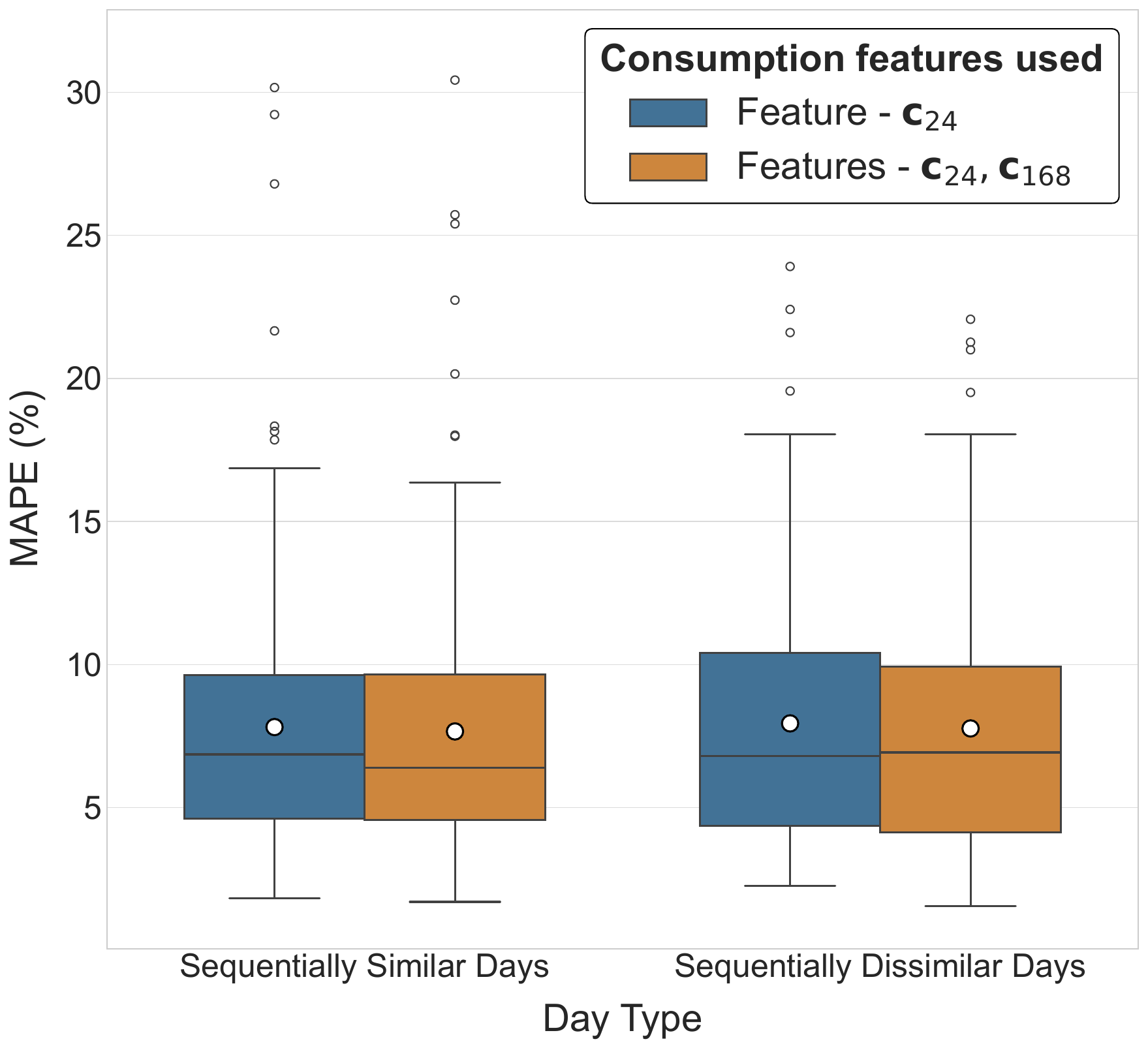"}
        \caption{\acrshort{dma} B}
        \label{fig:dma_b_cons_feature_analysis}
    \end{subfigure}    
    \begin{subfigure}[b]{0.3\linewidth}
        \centering
        \includegraphics[width=\linewidth]{"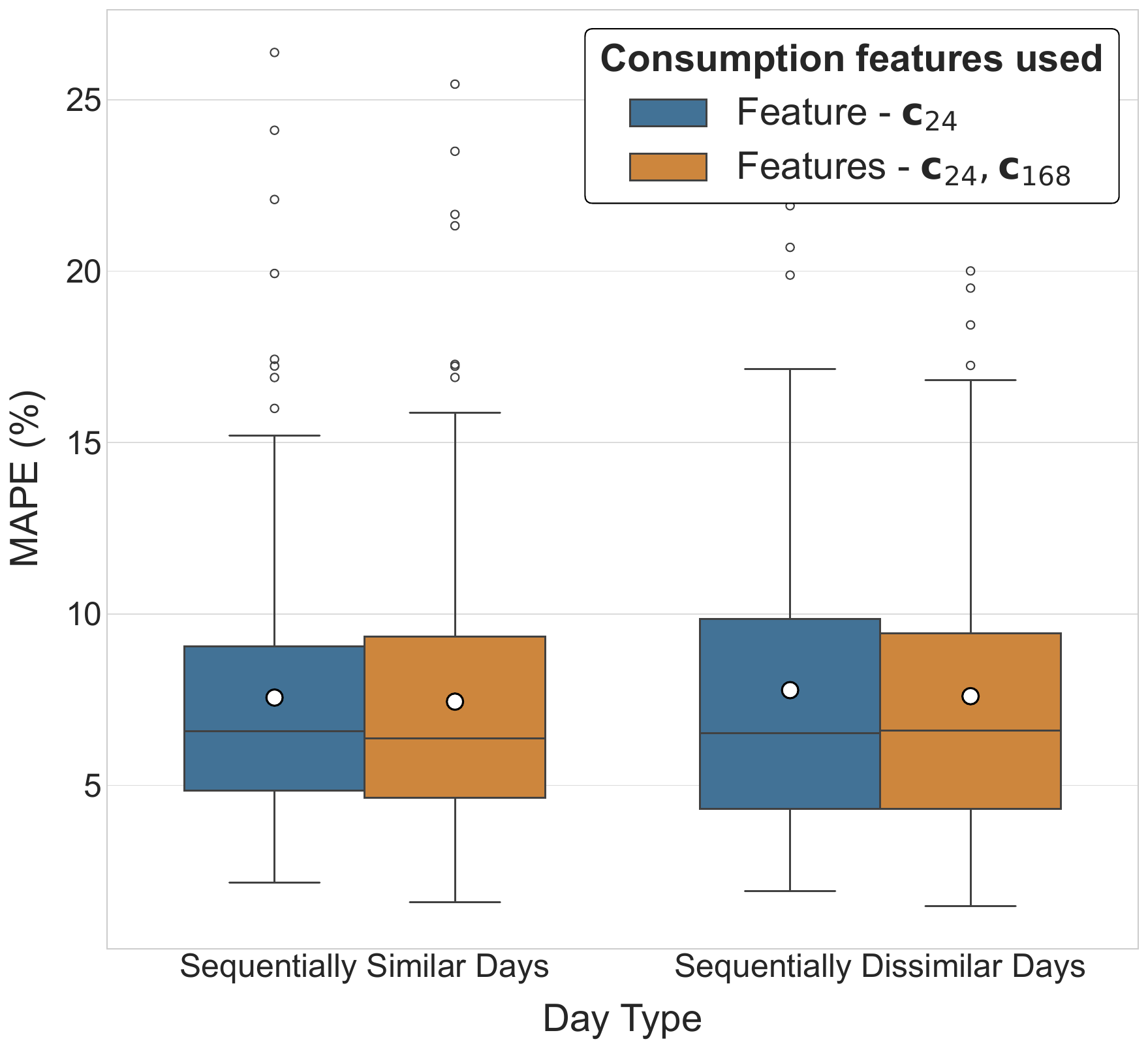"}
        \caption{\acrshort{dma} C}
        \label{fig:dma_c_cons_feature_analysis}
    \end{subfigure}

    \caption{Distribution of \acrshort{mape} for models using only a 24-hour consumption lag ($\textbf{c}_{24}$) versus both 24-hour and 168-hour lags ($\textbf{c}_{24}$, $\textbf{c}_{168}$). Performance is compared on days with sequentially similar patterns versus days with pattern disruptions across the three \acrshort{dmas}.}
    \label{fig:cons_feature_analysis}
    
\end{figure}

We persist with the feature set $\textbf{c}_{24}$, $\textbf{c}_{168}$ owing to a crucial nuance in the performance required for downstream application of the forecasts, especially for holidays and rare events. To investigate further, we stratified the forecast errors based on day-to-day pattern similarity, as detailed in Figure \ref{fig:cons_feature_analysis}. The analysis categorizes days into two groups of sequentially similar days, where the consumption pattern is expected to follow that of the preceding day (e.g., a Tuesday following a Monday), and sequentially dissimilar days, where a pattern break occurs as in weekend-weekday transition (Sunday to Monday), and occurrences of holidays during weekdays.

The quantitative Figure \ref{fig:cons_feature_analysis} reveals that while both models perform comparably on sequentially similar days, the inclusion of the weekly lag ($\textbf{c}_{168}$) provides an advantage on sequentially dissimilar days. The reduced number of outliers is tied to the reduction of \acrshort{mse}. For these transitional periods, the model leveraging both $\textbf{c}_{24}$ and $\textbf{c}_{168}$ achieves a lower median \acrshort{mape} and a smaller interquartile range, while reducing outliers in the prediction. This demonstrates that the $\textbf{c}_{168}$ feature acts as a vital stabilizing anchor, improving forecast robustness precisely when the daily pattern captured by $\textbf{c}_{24}$ is least reliable. Despite its limited impact on overall average metrics, its role in mitigating large errors during pattern shifts without affecting the performance on other samples justifies its inclusion in our feature set. We denote the endogenous feature set of  $[\textbf{c}_{24}$, $\textbf{c}_{168}]$ as $\mathcal{M}_{C}$. The inclusion of $\textbf{c}_{168}$ also provides us with an avenue to model demand features for rare events with relevant historical demand, which could have a variable lag. 


\subsubsection{Exogenous Features -  Experiment II}

The feature analysis is subsequently extended to incorporate exogenous variables, specifically the weather features shortlisted from the correlation analysis presented in Table \ref{tab:feature_correlations}. To quantify their impact, these features are systematically added to the baseline consumption feature set ($\mathcal{M}_{C}$) from  Experiment I, both individually and in various combinations. The quantitative outcomes of these experiments are summarized for each \acrshort{dma} in Table \ref{tab:dma_a_weather_features}, Table \ref{tab:dma_b_weather_features}, and Table \ref{tab:dma_c_weather_features}.

\begin{table}[htbp]
\centering
\caption{Performance metrics for various weather feature combinations for \acrshort{dma} A, sorted by \acrshort{mae} in descending order within each section. The base model, $\mathcal{M}_C$, includes consumption features. The best-performing model is highlighted in bold.}
\label{tab:dma_a_weather_features}
\begin{tabular}{l|c|c|c}
\hline
\textbf{Model Features} & \textbf{\acrshort{mae}} (kWh) & \textbf{\acrshort{mape}} (\%) & \textbf{\acrshort{mse}} (kWh$^2 \times 10^4$) \\
\hline
 $\mathcal{M}_C$ & $166.0 \pm 119.9$ & $8.40 \pm 5.20$ & $5.95 \pm 9.35$  \\
\hline
$\mathcal{M}_C, \mathbf{t}_{\text{dew}}$ & $149.1 \pm 100.8$ & $7.87 \pm 4.83$ & $4.73 \pm 7.72$ \\
$\mathcal{M}_C, \mathbf{t}_{\text{min}}$ & $130.4 \pm 88.5$ & $6.72 \pm 3.85$ & $3.65 \pm 6.55$ \\
$\mathcal{M}_C, \mathbf{t}_{\text{feels}}$ & $129.1 \pm 94.2$ & $6.58 \pm 3.90$ & $3.71 \pm 6.85$ \\
$\mathcal{M}_C, \mathbf{t}_{\text{max}}$ & $127.0 \pm 87.7$ & $6.41 \pm 3.44$ & $3.58 \pm 6.84$ \\
$\mathcal{M}_C, \mathbf{t}_{\text{amb}}$ & $122.7 \pm 80.0$ & $6.45 \pm 3.50$ & $3.21 \pm 6.16$ \\
\hline
$\mathcal{M}_C, \mathbf{t}_{\text{feels}}, \mathbf{t}_{\text{min}}$ & $126.2 \pm 92.2$ & $6.46 \pm 3.76$ & $3.60 \pm 7.30$ \\
$\mathcal{M}_C, \mathbf{t}_{\text{max}}, \mathbf{t}_{\text{feels}}$ & $123.4 \pm 85.4$ & $6.22 \pm 3.29$ & $3.33 \pm 6.18$ \\
$\mathcal{M}_C, \mathbf{t}_{\text{amb}}, \mathbf{t}_{\text{min}}$ & $121.1 \pm 80.5$ & $6.32 \pm 3.51$ & $3.17 \pm 6.20$ \\
$\mathcal{M}_C, \mathbf{t}_{\text{max}}, \mathbf{t}_{\text{min}}$ & $121.1 \pm 82.6$ & $6.28 \pm 3.59$ & $3.23 \pm 6.40$ \\
$\mathcal{M}_C, \mathbf{t}_{\text{max}}, \mathbf{t}_{\text{amb}}$ & $120.1 \pm 82.2$ & $6.21 \pm 3.51$ & $3.17 \pm 6.18$ \\
$\mathcal{M}_C, \mathbf{t}_{\text{feels}}, \mathbf{t}_{\text{amb}}$ & $117.6 \pm 78.6$ & $6.18 \pm 3.46$ & $3.03 \pm 6.08$ \\
\hline
$\mathcal{M}_C, \mathbf{t}_{\text{max}}, \mathbf{t}_{\text{feels}}, \mathbf{t}_{\text{min}}$ & $121.4 \pm 82.3$ & $6.47 \pm 4.04$ & $3.19 \pm 6.20$ \\
$\mathcal{M}_C, \mathbf{t}_{\text{max}}, \mathbf{t}_{\text{amb}}, \mathbf{t}_{\text{min}}$ & $119.3 \pm 84.9$ & $6.13 \pm 3.49$ & $3.23 \pm 6.27$ \\
$\mathcal{M}_C, \mathbf{t}_{\text{max}}, \mathbf{t}_{\text{feels}}, \mathbf{t}_{\text{amb}}$ & $118.9 \pm 81.3$ & $6.24 \pm 3.54$ & $3.10 \pm 6.20$ \\
\textbf{$\mathcal{M}_C, \mathbf{t}_{\text{feels}}, \mathbf{t}_{\text{amb}}, \mathbf{t}_{\text{min}}$} & \textbf{116.0 $\pm$ 78.7} & \textbf{6.12 $\pm$ 3.48} & \textbf{2.97 $\pm$ 6.02} \\
\hline
$\mathcal{M}_C, \mathbf{t}_{\text{max}}, \mathbf{t}_{\text{feels}}, \mathbf{t}_{\text{amb}}, \mathbf{t}_{\text{min}}$ & $117.7 \pm 81.4$ & $6.15 \pm 3.51$ & $3.06 \pm 6.22$ \\
\hline
\end{tabular}
\end{table}

\begin{table}[htbp]
\centering
\caption{Performance metrics for various weather feature combinations for \acrshort{dma} B, sorted by \acrshort{mae} in descending order within each section. The base model, $\mathcal{M}_C$, includes consumption features. The best-performing model is highlighted in bold.}
\label{tab:dma_b_weather_features}
\begin{tabular}{l|c|c|c}
\hline
\textbf{Model Features} & \textbf{\acrshort{mae}} (kWh) & \textbf{\acrshort{mape}} (\%) & \textbf{\acrshort{mse}} (kWh$^2 \times 10^4$) \\
\hline
 $\mathcal{M}_C$ & $293.6 \pm 203.6$ & $7.70 \pm 4.60$ & $17.44 \pm 24.22$ \\
\hline
$\mathcal{M}_C, \mathbf{t}_{\text{dew}}$ & $250.4 \pm 151.5$ & $6.85 \pm 4.15$ & $12.30 \pm 14.71$ \\
$\mathcal{M}_C, \mathbf{t}_{\text{min}}$ & $221.6 \pm 119.7$ & $6.11 \pm 3.35$ & $9.13 \pm 9.74$ \\
$\mathcal{M}_C, \mathbf{t}_{\text{feels}}$ & $216.8 \pm 127.1$ & $5.83 \pm 3.11$ & $9.10 \pm 11.18$ \\
$\mathcal{M}_C, \mathbf{t}_{\text{max}}$ & $212.6 \pm 121.0$ & $5.64 \pm 2.78$ & $8.71 \pm 9.78$ \\
\textbf{$\mathcal{M}_C, \mathbf{t}_{\text{amb}}$} & \textbf{195.2 $\pm$ 104.3} & \textbf{5.21 $\pm$ 2.57} & \textbf{7.22 $\pm$ 7.43} \\
\hline
$\mathcal{M}_C, \mathbf{t}_{\text{feels}}, \mathbf{t}_{\text{min}}$ & $204.5 \pm 120.1$ & $5.45 \pm 2.88$ & $8.17 \pm 9.88$ \\
$\mathcal{M}_C, \mathbf{t}_{\text{max}}, \mathbf{t}_{\text{feels}}$ & $201.8 \pm 113.8$ & $5.38 \pm 2.72$ & $7.95 \pm 8.90$ \\
$\mathcal{M}_C, \mathbf{t}_{\text{feels}}, \mathbf{t}_{\text{amb}}$ & $200.3 \pm 107.6$ & $5.47 \pm 2.86$ & $7.59 \pm 8.09$ \\
$\mathcal{M}_C, \mathbf{t}_{\text{max}}, \mathbf{t}_{\text{min}}$ & $199.9 \pm 111.3$ & $5.37 \pm 2.79$ & $7.72 \pm 8.62$ \\
$\mathcal{M}_C, \mathbf{t}_{\text{max}}, \mathbf{t}_{\text{amb}}$ & $196.2 \pm 103.5$ & $5.45 \pm 3.00$ & $7.25 \pm 7.89$ \\
$\mathcal{M}_C, \mathbf{t}_{\text{amb}}, \mathbf{t}_{\text{min}}$ & $195.5 \pm 106.6$ & $5.25 \pm 2.67$ & $7.34 \pm 7.93$ \\
\hline
$\mathcal{M}_C, \mathbf{t}_{\text{max}}, \mathbf{t}_{\text{feels}}, \mathbf{t}_{\text{min}}$ & $206.4 \pm 105.2$ & $5.91 \pm 3.81$ & $7.81 \pm 7.78$ \\
$\mathcal{M}_C, \mathbf{t}_{\text{feels}}, \mathbf{t}_{\text{amb}}, \mathbf{t}_{\text{min}}$ & $201.6 \pm 112.7$ & $5.50 \pm 3.05$ & $7.83 \pm 8.85$ \\
$\mathcal{M}_C, \mathbf{t}_{\text{max}}, \mathbf{t}_{\text{amb}}, \mathbf{t}_{\text{min}}$ & $201.5 \pm 114.4$ & $5.52 \pm 2.91$ & $7.82 \pm 10.18$ \\
$\mathcal{M}_C, \mathbf{t}_{\text{max}}, \mathbf{t}_{\text{feels}}, \mathbf{t}_{\text{amb}}$ & $199.5 \pm 114.1$ & $5.37 \pm 2.72$ & $7.74 \pm 8.91$ \\
\hline
$\mathcal{M}_C, \mathbf{t}_{\text{max}}, \mathbf{t}_{\text{feels}}, \mathbf{t}_{\text{amb}}, \mathbf{t}_{\text{min}}$ & $195.5 \pm 106.1$ & $5.26 \pm 2.72$ & $7.35 \pm 7.76$ \\
\hline
\end{tabular}
\end{table}

\begin{table}[htbp]
\centering
\caption{Performance metrics for various weather feature combinations for \acrshort{dma} C, sorted by \acrshort{mae} in descending order within each section. The base model, $\mathcal{M}_C$, includes consumption features. The best-performing model is highlighted in bold.}
\label{tab:dma_c_weather_features}
\begin{tabular}{l|c|c|c}
\hline
\textbf{Model Features} & \textbf{\acrshort{mae}} (kWh) & \textbf{\acrshort{mape}} (\%) & \textbf{\acrshort{mse}} (kWh$^2 \times 10^4$) \\
\hline
 $\mathcal{M}_C$ & $260.0 \pm 172.6$ & $7.50 \pm 4.20$ & $13.43 \pm 18.64$ \\
\hline
$\mathcal{M}_C, \mathbf{t}_{\text{dew}}$ & $241.0 \pm 144.5$ & $7.19 \pm 4.31$ & $11.15 \pm 13.50$ \\
$\mathcal{M}_C, \mathbf{t}_{\text{feels}}$ & $192.6 \pm 116.7$ & $5.63 \pm 2.89$ & $7.24 \pm 9.59$ \\
$\mathcal{M}_C, \mathbf{t}_{\text{max}}$ & $192.1 \pm 103.5$ & $5.74 \pm 2.86$ & $6.89 \pm 7.84$ \\
$\mathcal{M}_C, \mathbf{t}_{\text{min}}$ & $189.7 \pm 106.1$ & $5.64 \pm 2.81$ & $6.80 \pm 8.35$ \\
$\mathcal{M}_C, \mathbf{t}_{\text{amb}}$ & $176.7 \pm 91.4$ & $5.33 \pm 2.62$ & $5.80 \pm 6.45$ \\
\hline
$\mathcal{M}_C, \mathbf{t}_{\text{max}}, \mathbf{t}_{\text{min}}$ & $188.6 \pm 101.1$ & $5.64 \pm 2.78$ & $6.58 \pm 7.43$ \\
$\mathcal{M}_C, \mathbf{t}_{\text{max}}, \mathbf{t}_{\text{feels}}$ & $187.8 \pm 106.8$ & $5.53 \pm 2.66$ & $6.71 \pm 7.99$ \\
$\mathcal{M}_C, \mathbf{t}_{\text{feels}}, \mathbf{t}_{\text{min}}$ & $185.9 \pm 101.7$ & $5.62 \pm 2.98$ & $6.50 \pm 7.54$ \\
$\mathcal{M}_C, \mathbf{t}_{\text{max}}, \mathbf{t}_{\text{amb}}$ & $180.5 \pm 97.2$ & $5.35 \pm 2.62$ & $6.13 \pm 6.82$ \\
$\mathcal{M}_C, \mathbf{t}_{\text{amb}}, \mathbf{t}_{\text{min}}$ & $177.1 \pm 94.1$ & $5.26 \pm 2.56$ & $5.89 \pm 6.60$ \\
\textbf{$\mathcal{M}_C, \mathbf{t}_{\text{feels}}, \mathbf{t}_{\text{amb}}$} & \textbf{173.4 $\pm$ 92.7} & \textbf{5.24 $\pm$ 2.71} & \textbf{5.61 $\pm$ 6.54} \\
\hline
$\mathcal{M}_C, \mathbf{t}_{\text{max}}, \mathbf{t}_{\text{amb}}, \mathbf{t}_{\text{min}}$ & $182.3 \pm 93.2$ & $5.63 \pm 3.16$ & $6.05 \pm 6.51$ \\
$\mathcal{M}_C, \mathbf{t}_{\text{max}}, \mathbf{t}_{\text{feels}}, \mathbf{t}_{\text{min}}$ & $182.3 \pm 101.3$ & $5.43 \pm 2.79$ & $6.30 \pm 7.23$ \\
$\mathcal{M}_C, \mathbf{t}_{\text{max}}, \mathbf{t}_{\text{feels}}, \mathbf{t}_{\text{amb}}$ & $179.7 \pm 93.2$ & $5.53 \pm 3.07$ & $5.96 \pm 6.54$ \\
$\mathcal{M}_C, \mathbf{t}_{\text{feels}}, \mathbf{t}_{\text{amb}}, \mathbf{t}_{\text{min}}$ & $178.9 \pm 93.3$ & $5.45 \pm 3.01$ & $5.90 \pm 6.43$ \\
\hline
$\mathcal{M}_C, \mathbf{t}_{\text{max}}, \mathbf{t}_{\text{feels}}, \mathbf{t}_{\text{amb}}, \mathbf{t}_{\text{min}}$ & $178.5 \pm 92.4$ & $5.43 \pm 2.83$ & $5.87 \pm 6.25$ \\
\hline
\end{tabular}
\end{table}

A primary and consistent finding across all three regions is that the inclusion of any single temperature-related feature yields a substantial improvement in forecasting accuracy relative to the consumption-only baseline.The least improvement is for the addition of $\mathbf{t}_{\text{dew}}$, ranging from approximately $7.3\%$ for \acrshort{dma} C to $14.7\%$ for \acrshort{dma} B. The feature $\mathbf{t}_{\text{amb}}$ improved the results significantly, by $26.1\%$ for \acrshort{dma} A, and approximately $32\%$ to $33.5\%$ for \acrshort{dma} C and B, respectively. In accordance with the correlation analysis, $\mathbf{t}_{\text{amb}}$ consistently emerged as the most impactful single feature, whereas $\mathbf{t}_{\text{dew}}$ offered the least predictive power, albeit still providing a marked improvement.

The evaluation is then extended to feature combinations, motivated by the hypothesis that different temperature metrics would capture unique environmental nuances. For instance, $\mathbf{t}_{\text{feels}}$ incorporates factors such as humidity, and wind speed, while the pairing of $\mathbf{t}_{\text{min}}$ and $\mathbf{t}_{\text{max}}$ could potentially model the diurnal thermal dynamics of the heating network. For \acrshort{dma} A and \acrshort{dma} C, the inclusion of complementary features to different extent yielded the lowest errors, suggesting the model successfully captures subtler environmental dynamics. However, for \acrshort{dma} B, the single feature $\mathbf{t}_{\text{amb}}$ outperformed complex combinations, though the degradation with more features was minimal ($0.04\%$ to $0.15\%$ difference). Nevertheless, it is important to note that $\mathbf{t}_{\text{amb}}$ provides a strong reference for weather feature to augment endogenous features and further augmentation to the endogenous features can still provide an improvement in performance although small and to varied effects. In fact, augmenting with all four primary temperature features led to a slight degradation in performance relative to the optimal feature model. However, it is possible that this apparent redundancy stems from the entanglement of trend and seasonal information within the raw signals, which masks the distinct contributions of each variable. Therefore, rather than discarding the complex feature sets at this stage, we carry this comparative analysis into the next phase.

\subsubsection{Seasonal Decomposition of Features -  Experiment III}
A comprehensive comparison between models using raw features versus those augmented with decomposed components as described in Table \ref{tab:decomposition_notation_centered} is presented in Table \ref{tab:all_dma_original_decomposed_comparison}.
The results provide conclusive evidence with a uniform reduction in error across all metrics when employing decomposed feature sets. For every combination of weather features, the model with decomposed features significantly outperforms its raw counterpart. The magnitude of this improvement is substantial. For instance, in \acrshort{dma} C, the optimal raw feature model achieved an \acrshort{mae} of $173.4$ kWh, whereas the decomposed variant reduced this to $152.6$ kWh, a relative improvement of nearly $12\%$. Similar gains are observed across all districts, with \acrshort{dma} A achieving a remarkable \acrshort{mae} of $101.7$ kWh with decomposition. This consistency validates the hypothesis that explicit disentanglement of signal components simplifies the learning task for the model, allowing it to effectively map separate trend, seasonal, and residual drivers.

Crucially, decomposition alters the feature selection landscape. In Experiment II, adding multiple correlated temperature variables often resulted in degraded performance particularly in \acrshort{dma} B. However, after decomposition, the feature set [$\mathcal{M}_C, \mathbf{t}_{\text{max}}, \mathbf{t}_{\text{feels}}, \mathbf{t}_{\text{amb}}$]) emerge as the top performers for \acrshort{dma} B and C. This suggests that while raw correlated features introduce noise, their decomposed components contain distinct, complementary information that the model can successfully leverage to effectively unlock the value of these complex feature sets.

\begin{table}[htbp]
\centering
\caption{Unified comparison of model performance across all \acrshort{dmas} for models with raw features versus those with decomposed features.}
\label{tab:all_dma_original_decomposed_comparison}
\renewcommand{\arraystretch}{1.1} 
\setlength{\tabcolsep}{4pt} 
\begin{tabular}{l|l|c|c|c|c|c|c}
\toprule
\multirow{2}{*}{\textbf{\acrshort{dma}}} & \multirow{2}{*}{\textbf{Features}} & \multicolumn{2}{c|}{\textbf{\acrshort{mse}} (kWh$^2 \times 10^4$)} & \multicolumn{2}{c|}{\textbf{\acrshort{mape}} (\%)} & \multicolumn{2}{c}{\textbf{\acrshort{mae}} (kWh)} \\
\cline{3-4} \cline{5-6} \cline{7-8}

& & Original & Decomposed & Original & Decomposed & Original & Decomposed \\

\hline

\multirow{15}{*}{\textit{\acrshort{dma} A}} 
& $\mathcal{M}_C, \mathbf{t}_{\text{max}}$ & 3.58 & 2.64 & 6.41 & 5.47 & 127.0 & 106.1 \\
& $\mathcal{M}_C, \mathbf{t}_{\text{feels}}$ & 3.71 & 2.68 & 6.58 & 5.53 & 129.1 & 105.7 \\
& $\mathcal{M}_C, \mathbf{t}_{\text{amb}}$ & 3.21 & 2.39 & 6.45 & 5.50 & 122.7 & 102.8 \\
& $\mathcal{M}_C, \mathbf{t}_{\text{min}}$ & 3.65 & 2.64 & 6.72 & 5.67 & 130.4 & 107.1 \\
\cline{2-8}
& $\mathcal{M}_C, \mathbf{t}_{\text{max}}, \mathbf{t}_{\text{feels}}$ & 3.33 & 2.69 & 6.22 & 5.68 & 123.4 & 108.6 \\
& $\mathcal{M}_C, \mathbf{t}_{\text{max}}, \mathbf{t}_{\text{amb}}$ & 3.17 & 2.52 & 6.21 & 5.55 & 120.1 & 105.1 \\
& $\mathcal{M}_C, \mathbf{t}_{\text{max}}, \mathbf{t}_{\text{min}}$ & 3.23 & 2.55 & 6.28 & 5.48 & 121.1 & 104.9 \\
& $\mathcal{M}_C, \mathbf{t}_{\text{feels}}, \mathbf{t}_{\text{amb}}$ & 3.03 & 2.56 & 6.18 & 5.66 & 117.6 & 106.3 \\
& $\mathcal{M}_C, \mathbf{t}_{\text{feels}}, \mathbf{t}_{\text{min}}$ & 3.60 & 2.98 & 6.46 & 6.04 & 126.2 & 113.9 \\
& $\mathcal{M}_C, \mathbf{t}_{\text{amb}}, \mathbf{t}_{\text{min}}$ & 3.17 & 2.69 & 6.32 & 5.76 & 121.1 & 108.6 \\
\cline{2-8}
& $\mathcal{M}_C, \mathbf{t}_{\text{max}}, \mathbf{t}_{\text{feels}}, \mathbf{t}_{\text{amb}}$ & 3.10 & 2.47 & 6.24 & 5.49 & 118.9 & 104.7 \\
& $\mathcal{M}_C, \mathbf{t}_{\text{max}}, \mathbf{t}_{\text{feels}}, \mathbf{t}_{\text{min}}$ & 3.19 & 2.50 & 6.47 & 5.41 & 121.4 & 103.8 \\
& $\mathcal{M}_C, \mathbf{t}_{\text{max}}, \mathbf{t}_{\text{amb}}, \mathbf{t}_{\text{min}}$ & 3.23 & 2.43 & 6.13 & 5.37 & 119.3 & 103.0 \\
& $\mathcal{M}_C, \mathbf{t}_{\text{feels}}, \mathbf{t}_{\text{amb}}, \mathbf{t}_{\text{min}}$ & 2.97 & \textbf{2.45} & 6.12 & \textbf{5.37} & 116.0 & \textbf{101.7} \\
\cline{2-8}

\hline

\multirow{15}{*}{\textit{\acrshort{dma} B}}
& $\mathcal{M}_C, \mathbf{t}_{\text{max}}$ & 8.71 & 6.83 & 5.64 & 4.86 & 212.6 & 184.2 \\
& $\mathcal{M}_C, \mathbf{t}_{\text{feels}}$ & 9.10 & 6.83 & 5.83 & 4.82 & 216.8 & 184.0 \\
& $\mathcal{M}_C, \mathbf{t}_{\text{amb}}$ & 7.22 & 5.39 & 5.21 & 4.59 & 195.2 & 166.7 \\
& $\mathcal{M}_C, \mathbf{t}_{\text{min}}$ & 9.13 & 6.58 & 6.11 & 4.90 & 221.6 & 182.4 \\
\cline{2-8}
& $\mathcal{M}_C, \mathbf{t}_{\text{max}}, \mathbf{t}_{\text{feels}}$ & 7.95 & 6.03 & 5.38 & 4.79 & 201.8 & 175.7 \\
& $\mathcal{M}_C, \mathbf{t}_{\text{max}}, \mathbf{t}_{\text{amb}}$ & 7.25 & 5.81 & 5.45 & 4.80 & 196.2 & 173.4 \\
& $\mathcal{M}_C, \mathbf{t}_{\text{max}}, \mathbf{t}_{\text{min}}$ & 7.72 & 6.35 & 5.37 & 4.84 & 199.9 & 180.1 \\
& $\mathcal{M}_C, \mathbf{t}_{\text{feels}}, \mathbf{t}_{\text{amb}}$ & 7.59 & 5.74 & 5.47 & 4.71 & 200.3 & 172.7 \\
& $\mathcal{M}_C, \mathbf{t}_{\text{feels}}, \mathbf{t}_{\text{min}}$ & 8.17 & 6.87 & 5.45 & 5.13 & 204.5 & 188.0 \\
& $\mathcal{M}_C, \mathbf{t}_{\text{amb}}, \mathbf{t}_{\text{min}}$ & 7.34 & 5.81 & 5.25 & 4.77 & 195.5 & 175.1 \\
\cline{2-8}
& $\mathcal{M}_C, \mathbf{t}_{\text{max}}, \mathbf{t}_{\text{feels}}, \mathbf{t}_{\text{amb}}$ & 7.74 & \textbf{5.35} & 5.37 & \textbf{4.46} & 199.5 & \textbf{165.5} \\
& $\mathcal{M}_C, \mathbf{t}_{\text{max}}, \mathbf{t}_{\text{feels}}, \mathbf{t}_{\text{min}}$ & 7.81 & 5.77 & 5.91 & 4.58 & 206.4 & 171.2 \\
& $\mathcal{M}_C, \mathbf{t}_{\text{max}}, \mathbf{t}_{\text{amb}}, \mathbf{t}_{\text{min}}$ & 7.82 & 5.90 & 5.52 & 4.66 & 201.5 & 174.1 \\
& $\mathcal{M}_C, \mathbf{t}_{\text{feels}}, \mathbf{t}_{\text{amb}}, \mathbf{t}_{\text{min}}$ & 7.83 & 5.63 & 5.50 & 4.62 & 201.6 & 169.4 \\
\cline{2-8}

\hline

\multirow{15}{*}{\textit{\acrshort{dma} C}}
& $\mathcal{M}_C, \mathbf{t}_{\text{max}}$ & 6.89 & 5.27 & 5.74 & 4.99 & 192.1 & 165.1 \\
& $\mathcal{M}_C, \mathbf{t}_{\text{feels}}$ & 7.24 & 5.17 & 5.63 & 4.74 & 192.6 & 159.0 \\
& $\mathcal{M}_C, \mathbf{t}_{\text{amb}}$ & 5.80 & 4.86 & 5.33 & 4.75 & 176.7 & 158.4 \\
& $\mathcal{M}_C, \mathbf{t}_{\text{min}}$ & 6.80 & 5.64 & 5.64 & 5.09 & 189.7 & 169.3 \\
\cline{2-8}
& $\mathcal{M}_C, \mathbf{t}_{\text{max}}, \mathbf{t}_{\text{feels}}$ & 6.71 & 4.75 & 5.53 & 4.78 & 187.8 & 156.4 \\
& $\mathcal{M}_C, \mathbf{t}_{\text{max}}, \mathbf{t}_{\text{amb}}$ & 6.13 & 5.17 & 5.35 & 5.11 & 180.5 & 166.8 \\
& $\mathcal{M}_C, \mathbf{t}_{\text{max}}, \mathbf{t}_{\text{min}}$ & 6.58 & 5.04 & 5.64 & 4.86 & 188.6 & 161.9 \\
& $\mathcal{M}_C, \mathbf{t}_{\text{feels}}, \mathbf{t}_{\text{amb}}$ & 5.61 & 5.00 & 5.24 & 4.93 & 173.4 & 161.9 \\
& $\mathcal{M}_C, \mathbf{t}_{\text{feels}}, \mathbf{t}_{\text{min}}$ & 6.50 & 5.34 & 5.62 & 5.08 & 185.9 & 166.2 \\
& $\mathcal{M}_C, \mathbf{t}_{\text{amb}}, \mathbf{t}_{\text{min}}$ & 5.89 & 5.25 & 5.26 & 4.85 & 177.1 & 163.8 \\
\cline{2-8}
& $\mathcal{M}_C, \mathbf{t}_{\text{max}}, \mathbf{t}_{\text{feels}}, \mathbf{t}_{\text{amb}}$ & 5.96 & \textbf{4.43} & 5.53 & \textbf{4.65} & 179.7 & \textbf{152.6} \\
& $\mathcal{M}_C, \mathbf{t}_{\text{max}}, \mathbf{t}_{\text{feels}}, \mathbf{t}_{\text{min}}$ & 6.30 & 4.72 & 5.43 & 4.72 & 182.3 & 155.9 \\
& $\mathcal{M}_C, \mathbf{t}_{\text{max}}, \mathbf{t}_{\text{amb}}, \mathbf{t}_{\text{min}}$ & 6.05 & 5.41 & 5.63 & 5.28 & 182.3 & 170.6 \\
& $\mathcal{M}_C, \mathbf{t}_{\text{feels}}, \mathbf{t}_{\text{amb}}, \mathbf{t}_{\text{min}}$ & 5.90 & 4.60 & 5.45 & 4.67 & 178.9 & 154.1 \\
\cline{2-8}
\bottomrule
\end{tabular}
\end{table}

Despite the varying optimal configurations, a critical outcome of decomposition is the significant narrowing of the performance spread between feature sets. For instance, in \acrshort{dma} A, the difference between the single best feature ($\mathbf{t}_{\text{amb}}$) and the complex optimal set is merely $1.1$ kWh ($102.8$ vs $101.7$ kWh). This characteristic highlights the framework's operational flexibility. It implies that while complex combinations offer a statistical edge, the decomposition enables the model to extract the bulk of the predictive signal from even minimal data. In real-world scenarios where specific meteorological information may be unavailable or faulty, the model remains highly reliable using only ambient temperature. However, to rigorously evaluate the framework's maximum potential for this benchmarking study, we proceed with the specific optimal decomposed feature set identified for each district: [$\mathcal{M}^d_C, \mathbf{t}^d_{\text{feels}}, \mathbf{t}^d_{\text{amb}}, \mathbf{t}^d_{\text{min}}$] for \acrshort{dma} A, and [$\mathcal{M}^d_C, \mathbf{t}^d_{\text{max}}, \mathbf{t}^d_{\text{feels}}, \mathbf{t}^d_{\text{amb}}$] for \acrshort{dma} B and C.



\subsubsection{Time and Calendrical Features -  Experiment IV}

The final stage of our feature engineering analysis assessed the impact of including explicit temporal information. We augmented our best-performing decomposed models from Experiment III with cyclical features representing the day of the week, encoded as a sine-cosine pair ($T_{\text{cyc}}$). The comparative results are presented in Table \ref{tab:time_component_impact}.

Contrary to common assumptions in time-series forecasting, the inclusion of explicit cyclical time features resulted in a significant degradation in performance across all three districts. For \acrshort{dma} B and C, the \acrshort{mae} increased by approximately $6.5\%$ and $5.5\%$, respectively, with Wilcoxon Signed-Rank Test p-values ($p < 0.05$) confirming that this degradation is not due to random chance, while for \acrshort{dma} A, the error increased from $101.7$ kWh to $103.0$ kWh.

Through the model architecture, wherein the scalograms encode the time-frequency information, and the lack of pooling layers which maintain positional information, along the with nature of the objective function of forecasting fixed time intervals from 00:00 to 23:00, we induce the capacity to capture the temporal aspect. 
Although counter-intuitive, adding explicit $T_{\text{cyc}}$ vectors seems to introduce redundant information that effectively acts as noise, leading to reduced generalisation. Consequently, we discard explicit cyclical time encoding for the final model configuration, preferring the simpler and more robust architecture that derives temporal context directly from the signal dynamics.

\begin{table}[htbp]
\centering
\caption{Performance comparison of models with and without cyclical time features ($+T_{\text{cyc}}$). The temporal features are added to the best performing decomposed models: [$\mathcal{M}^d_C, \mathbf{t}^d_{\text{feels}}, \mathbf{t}^d_{\text{amb}}, \mathbf{t}^d_{\text{min}}$] for \textit{\acrshort{dma} A}, and [$\mathcal{M}^d_C, \mathbf{t}^d_{\text{max}}, \mathbf{t}^d_{\text{feels}}, \mathbf{t}^d_{\text{amb}}$] for \textit{\acrshort{dma} B} and \textit{\acrshort{dma} C}.}
\label{tab:time_component_impact}
\renewcommand{\arraystretch}{1.2} 
\begin{tabular}{l|l|c|c|c|>{\centering\arraybackslash}p{3cm}}
\toprule
\textbf{\acrshort{dma}} & \textbf{Features} & \textbf{\acrshort{mse}} (kWh$^2 \times 10^4$) & \textbf{\acrshort{mape}} (\%) & \textbf{\acrshort{mae}} (kWh) & \textbf{Wilcoxon Signed-Rank Test p-value}  \\
\hline
\multirow{2}{*}{\textit{\acrshort{dma} A}} & $\mathcal{M}^d_C, \mathbf{t}^d_{\text{feels}}, \mathbf{t}^d_{\text{amb}}, \mathbf{t}^d_{\text{min}}$ & 2.45 & 5.37 & 101.7  & \multirow{2}{*}{$> 0.05$}\\
\cline{2-5}
 & With Time Features ($+T_{\text{cyc}}$) & 2.44 & 5.50 & 103.0 & \\
\hline
\multirow{2}{*}{\textit{\acrshort{dma} B}} & $\mathcal{M}^d_C, \mathbf{t}^d_{\text{max}}, \mathbf{t}^d_{\text{feels}}, \mathbf{t}^d_{\text{amb}}$ & 5.35 & 4.46 & 165.5  & \multirow{2}{*}{$< 0.05$} \\
\cline{2-5}
 & With Time Features ($+T_{\text{cyc}}$) & 5.93 & 4.93 & 176.3 & \\
\hline
\multirow{2}{*}{\textit{\acrshort{dma} C}} & $\mathcal{M}^d_C, \mathbf{t}^d_{\text{max}}, \mathbf{t}^d_{\text{feels}}, \mathbf{t}^d_{\text{amb}}$ & 4.43 & 4.65 & 152.6  & \multirow{2}{*}{0.035} \\
\cline{2-5}
 & With Time Features ($+T_{\text{cyc}}$) & 5.06 & 4.90 & 161.0 & \\
\bottomrule
\end{tabular}
\end{table}

To address the challenge of modelling non-continuous deterministic variables, we investigated the effect of incorporating explicit holiday features. We integrate holiday information through three distinct strategies. First, a categorical ($\mathbf{h}_{categorical}$) one-hot encoded feature indicating holiday and non-holiday days for each hour of day. Second, to capture holiday-specific behaviour, a lagged ($\mathbf{h}_{lagged}$) feature replacing the consumption value from the same day of the previous week ($\mathbf{c}_{168}$) with the consumption from the previous occurrence of the same holiday (e.g., Christmas 2018 data used for Christmas 2019). This preserves the principle behind the use $\mathbf{c}_{168}$ which is to provide the model with representative demand behaviour for a specific day. This dynamic replacement accounts for the fact that festive holidays deviate significantly from typical weekly patterns. Finally, a combined ($\mathbf{h}_{lagged}, \mathbf{h}_{categorical}$) feature set including both the categorical and lagged holiday features.

\begin{table}[htbp]
\centering
\caption{Performance comparison of different holiday handling strategies. The model with their best performance features are compared against three approaches: adding holiday as a categorical feature $\mathbf{h}_{categorical}$, replacing the feature $C_{168}$ with the consumption values of the previous occurrence of the same holiday $\mathbf{h}_{lagged}$, and combining both strategies $\mathbf{h}_{categorical}$ and $\mathbf{h}_{lagged}$.}
\label{tab:holiday_handling_impact}
\renewcommand{\arraystretch}{1.2} 
\begin{tabular}{l|l|c|c|c}
\toprule
\textbf{\acrshort{dma}} & \textbf{Holiday Handling Method} & \textbf{\acrshort{mse}} ($\times 10^{4}$ kWh$^2$) & \textbf{\acrshort{mape}} (\%) & \textbf{\acrshort{mae}} (kWh) \\
\hline

\multirow{4}{*}{\acrshort{dma} A} 
& $\mathcal{M}^d_C, \mathbf{t}^d_{\text{feels}}, \mathbf{t}^d_{\text{amb}}, \mathbf{t}^d_{\text{min}}$ & $2.45 \pm 5.47$ & $5.4 \pm 3.3$ & $101.73 \pm 71.75$ \\
& $\mathcal{M}^d_C, \mathbf{t}^d_{\text{feels}}, \mathbf{t}^d_{\text{amb}}, \mathbf{t}^d_{\text{min}}, \mathbf{h}_{categorical}$ & $2.46 \pm 5.58$ & $5.5 \pm 3.4$ & $103.42 \pm 71.32$ \\
& $\mathcal{M}^d_C, \mathbf{t}^d_{\text{feels}}, \mathbf{t}^d_{\text{amb}}, \mathbf{t}^d_{\text{min}}, \mathbf{h}_{lagged}$ & $2.51 \pm 5.54$ & $5.6 \pm 3.3$ & $104.84 \pm 71.03$ \\
& $\mathcal{M}^d_C, \mathbf{t}^d_{\text{feels}}, \mathbf{t}^d_{\text{amb}}, \mathbf{t}^d_{\text{min}}, \mathbf{h}_{lagged}, \mathbf{h}_{categorical}$ & $\mathbf{2.36 \pm 5.42}$ & $\mathbf{5.3 \pm 3.3}$ & $\mathbf{99.76 \pm 72.71}$ \\

\hline

\multirow{4}{*}{\acrshort{dma} B} 
& $\mathcal{M}^d_C, \mathbf{t}^d_{\text{max}}, \mathbf{t}^d_{\text{feels}}, \mathbf{t}^d_{\text{amb}}$ & $5.35 \pm 5.99$ & $4.5 \pm 2.3$ & $165.48 \pm 88.42$ \\
& $\mathcal{M}^d_C, \mathbf{t}^d_{\text{max}}, \mathbf{t}^d_{\text{feels}}, \mathbf{t}^d_{\text{amb}}, \mathbf{h}_{categorical}$ & $5.63 \pm 6.22$ & $4.8 \pm 3.0$ & $171.97 \pm 90.57$ \\
& $\mathcal{M}^d_C, \mathbf{t}^d_{\text{max}}, \mathbf{t}^d_{\text{feels}}, \mathbf{t}^d_{\text{amb}}, \mathbf{h}_{lagged}$ & $\mathbf{5.31 \pm 6.00}$ & $\mathbf{4.4 \pm 2.2}$ & $\mathbf{164.81 \pm 88.36}$ \\
& $\mathcal{M}^d_C, \mathbf{t}^d_{\text{max}}, \mathbf{t}^d_{\text{feels}}, \mathbf{t}^d_{\text{amb}}, \mathbf{h}_{lagged}, \mathbf{h}_{categorical}$ & $5.45 \pm 6.22$ & $4.7 \pm 2.7$ & $169.12 \pm 87.10$ \\

\hline

\multirow{4}{*}{\acrshort{dma} C} 
& $\mathcal{M}^d_C, \mathbf{t}^d_{\text{max}}, \mathbf{t}^d_{\text{feels}}, \mathbf{t}^d_{\text{amb}}$ & $\mathbf{4.43 \pm 4.81}$ & $\mathbf{4.6 \pm 2.3}$ & $\mathbf{152.60 \pm 74.10}$ \\
& $\mathcal{M}^d_C, \mathbf{t}^d_{\text{max}}, \mathbf{t}^d_{\text{feels}}, \mathbf{t}^d_{\text{amb}}, \mathbf{h}_{categorical}$ & $4.40 \pm 4.75$ & $4.8 \pm 2.6$ & $153.08 \pm 72.30$ \\
& $\mathcal{M}^d_C, \mathbf{t}^d_{\text{max}}, \mathbf{t}^d_{\text{feels}}, \mathbf{t}^d_{\text{amb}}, \mathbf{h}_{lagged}$ & $7.82 \pm 7.86$ & $6.3 \pm 3.0$ & $207.81 \pm 101.98$ \\
& $\mathcal{M}^d_C, \mathbf{t}^d_{\text{max}}, \mathbf{t}^d_{\text{feels}}, \mathbf{t}^d_{\text{amb}}, \mathbf{h}_{lagged}, \mathbf{h}_{categorical}$ & $4.56 \pm 5.21$ & $4.8 \pm 2.7$ & $154.59 \pm 75.79$ \\
\bottomrule
\end{tabular}
\end{table}




\begin{figure}[htbp]
    \centering 

    \begin{subfigure}[b]{0.90\linewidth}
        \centering
\includegraphics[width=1.0\linewidth]{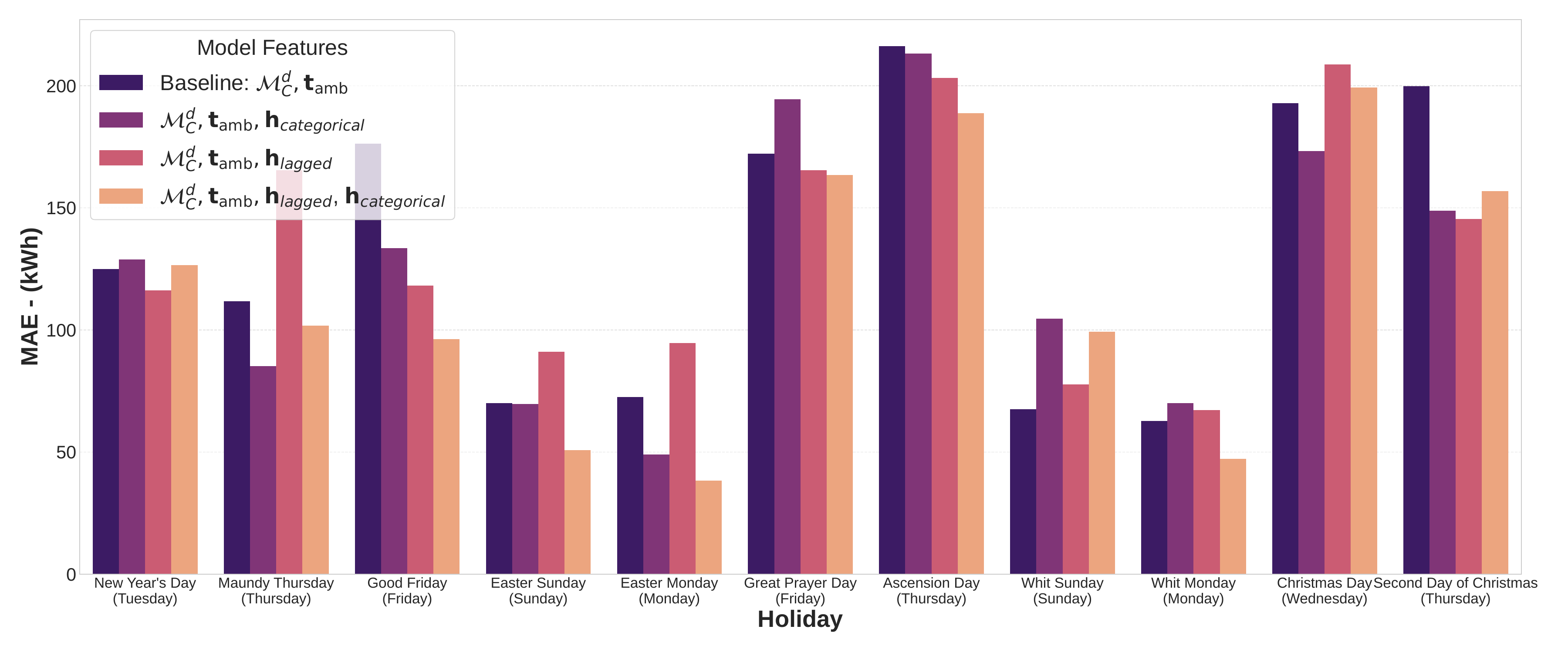}
        \caption{\acrshort{dma} A}
        \label{fig:dma_a_holiday_bar}
    \end{subfigure}

    \begin{subfigure}[b]{0.90\linewidth}
        \centering
\includegraphics[width=1.0\linewidth]{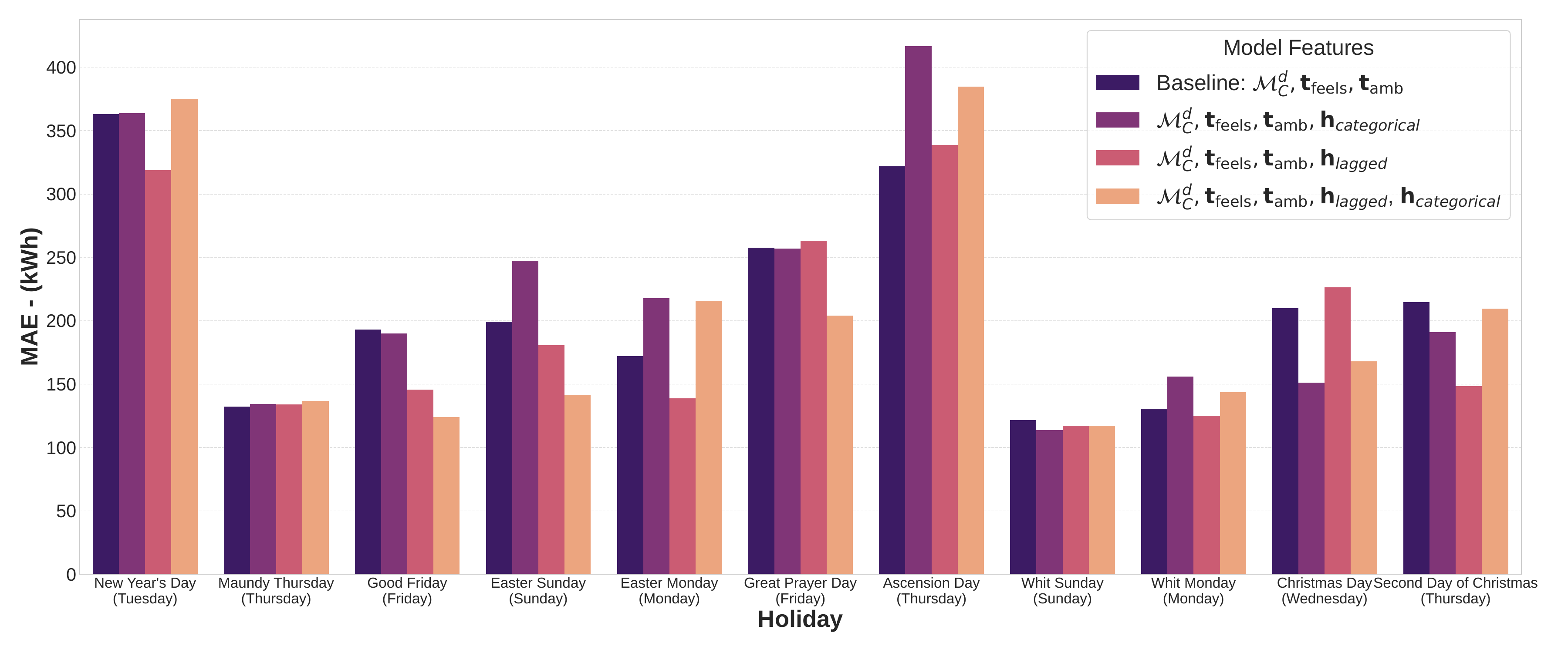}
        \caption{\acrshort{dma} B}
        \label{fig:dma_b_holiday_bar}
    \end{subfigure}

    \begin{subfigure}[b]{0.95\linewidth}
        \centering
\includegraphics[width=1.0\linewidth]{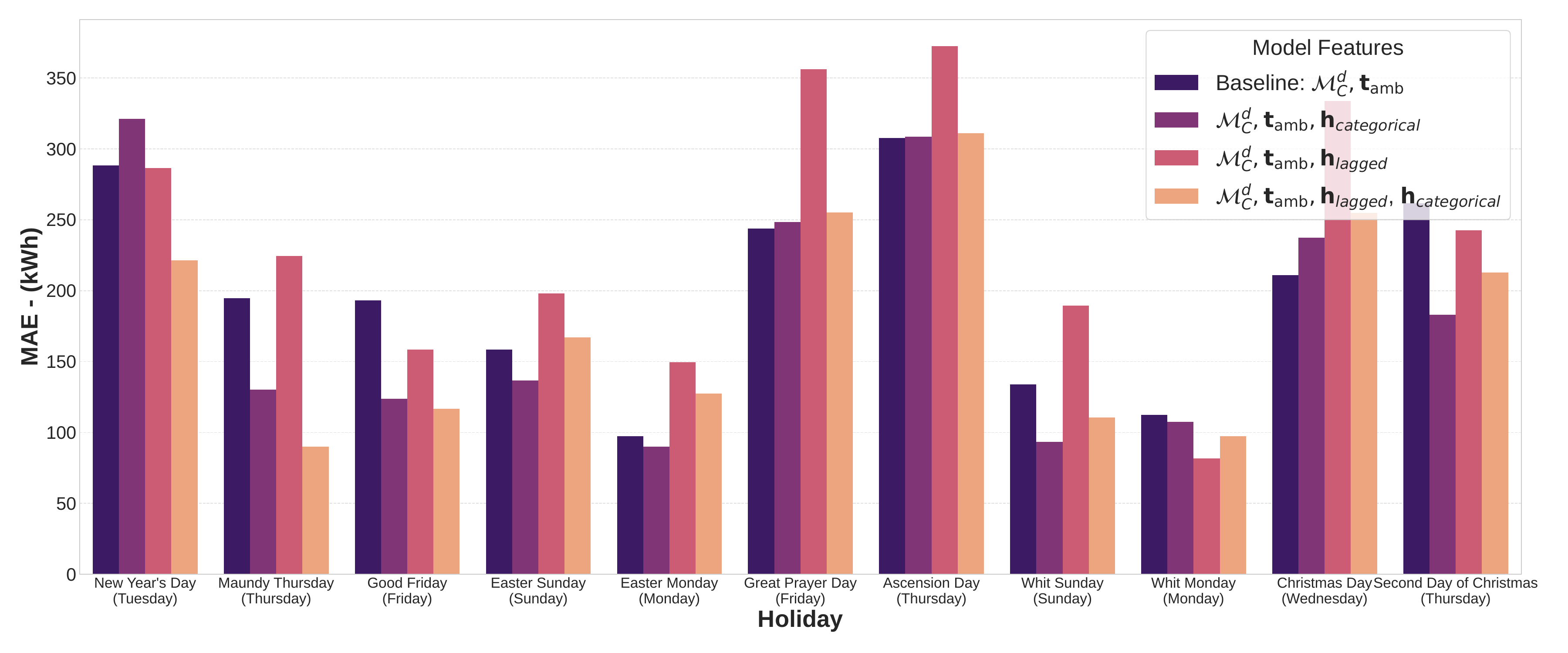}
        \caption{\acrshort{dma} C}
        \label{fig:dma_c_holiday_bar}
    \end{subfigure}

    \caption{Model Error (\acrshort{mae}) on public holidays in Denmark comparing different embeddings of holiday features across three \acrshort{dmas} (a) \acrshort{dma} A, (b) \acrshort{dma} B, and (c) \acrshort{dma} C.}

    \label{fig:dma_holiday_bar}
    
\end{figure}

\begin{figure}[htbp]
    \centering 

    \begin{subfigure}[b]{0.45\linewidth}    
        \centering
\includegraphics[width=1.0\linewidth]{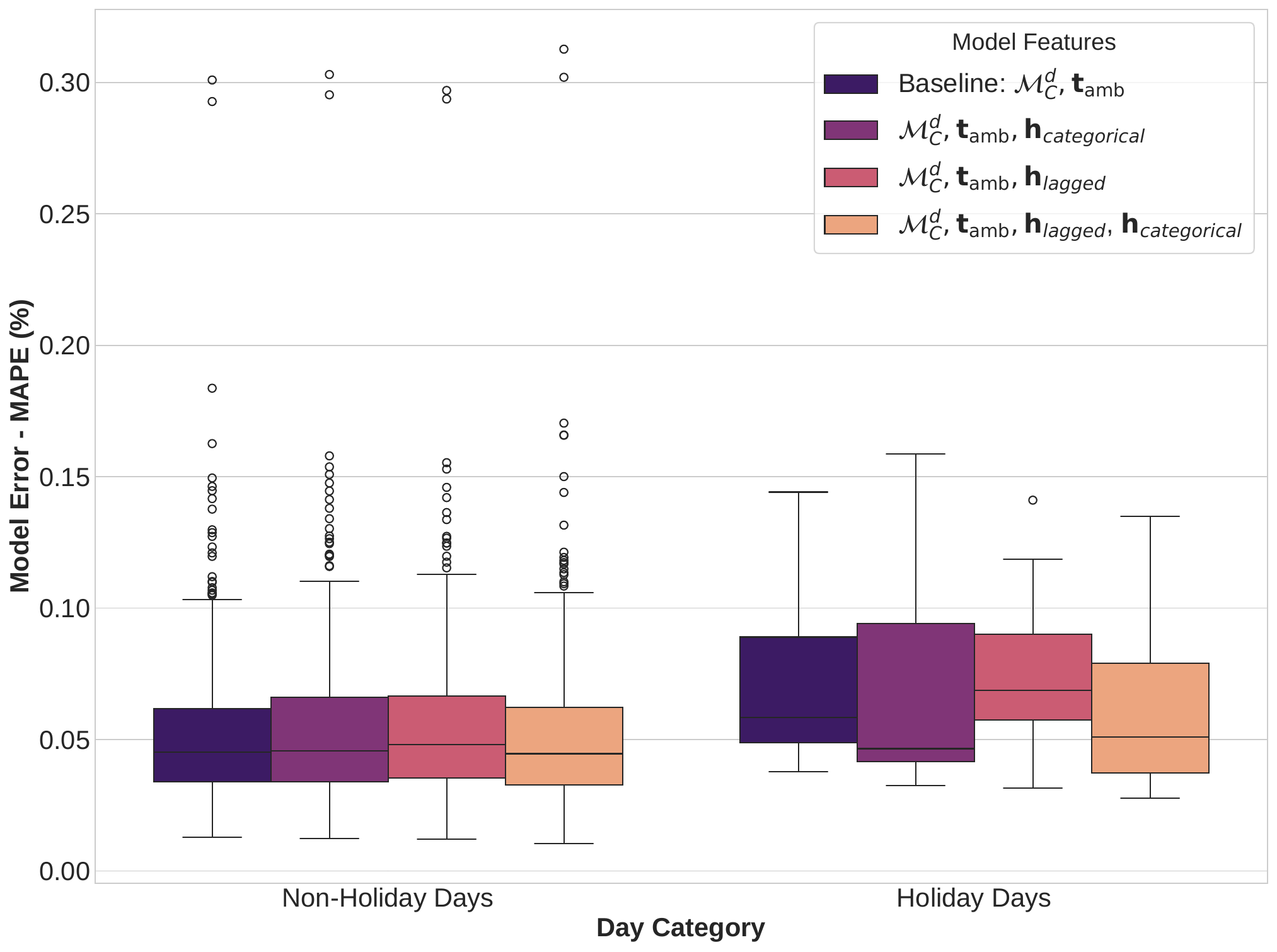}
        \caption{\acrshort{dma} A}
        \label{fig:dma_a_holiday_box}
    \end{subfigure}    
    \begin{subfigure}[b]{0.45\linewidth}
        \centering
\includegraphics[width=1.0\linewidth]{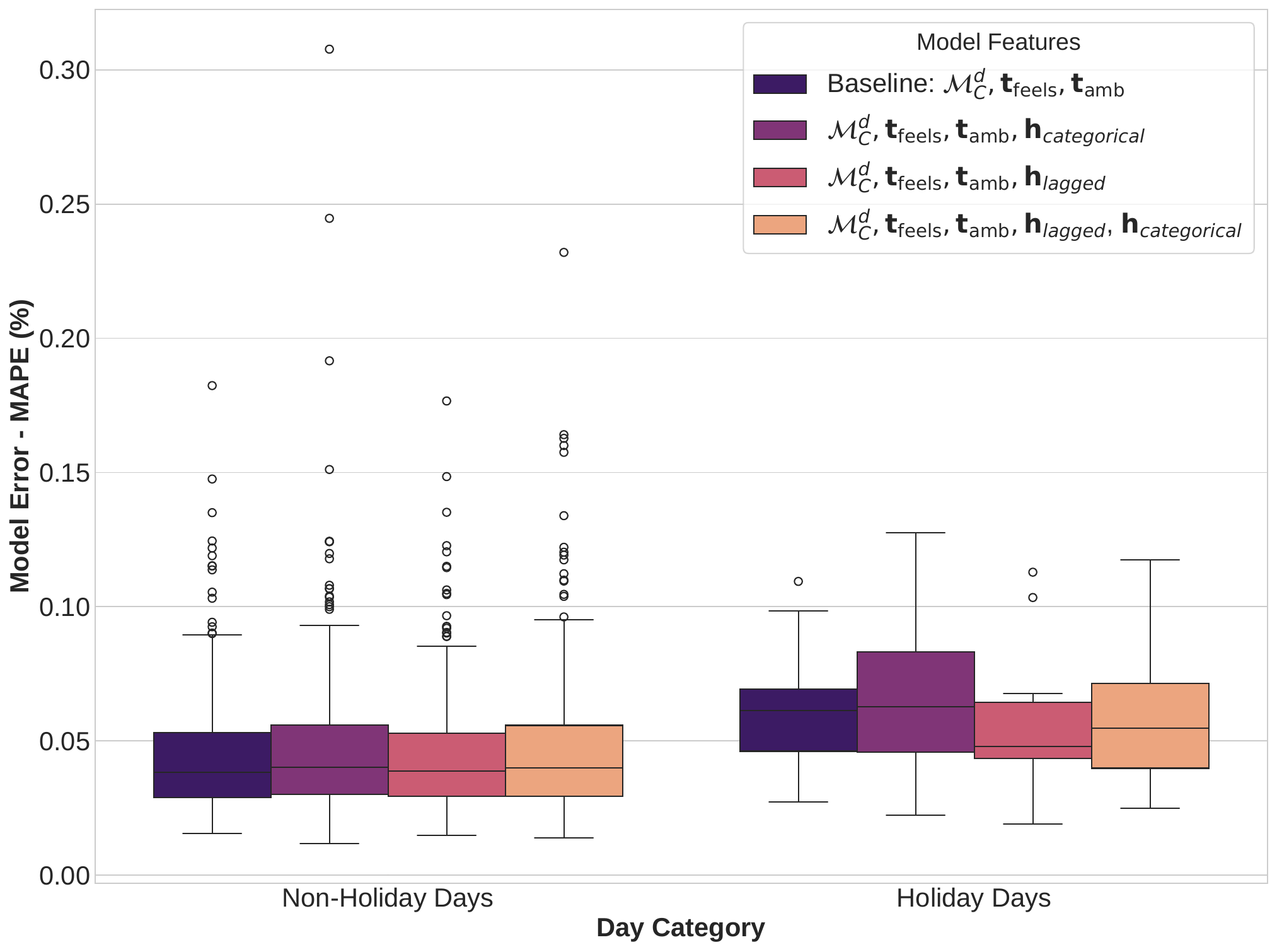}
        \caption{\acrshort{dma} B}
        \label{fig:dma_b_holiday_box}
    \end{subfigure} 

\vspace{0.25cm}
    
    \begin{subfigure}[b]{0.5\linewidth}
        \centering
\includegraphics[width=1.0\linewidth]{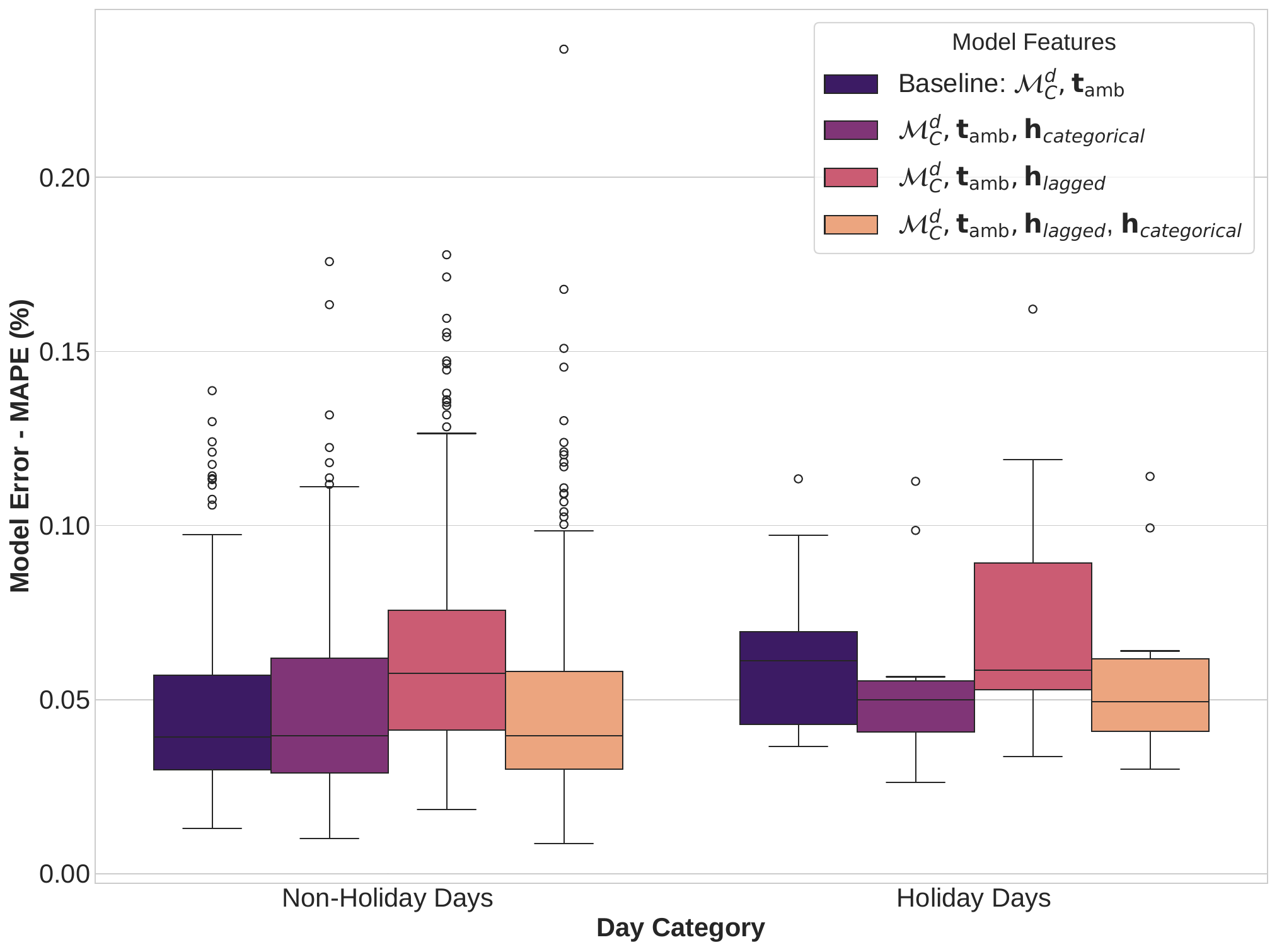}
        \caption{\acrshort{dma} C}
        \label{fig:dma_c_holiday_box}
    \end{subfigure}

    \caption{Distribution of \acrshort{mape} across the three \acrshort{dmas} comparing different embeddings of holiday features for holiday and non-holiday samples with comparison to the baseline without holiday features. The box plots in (a), (b), and (c) show the model error for \acrshort{dma} A, \acrshort{dma} B, and \acrshort{dma} C, respectively.}

    \label{fig:dma_holiday_box}
    
\end{figure}

The overall quantitative performance across the entire test set for these approaches is summarized in Table \ref{tab:holiday_handling_impact}. At first glance, the improvements appear marginal (e.g., reducing MAE by $<1$ kWh). However, given that public holidays constitute less than $4\%$ of the annual data, aggregate metrics heavily dilute the impact of these features. To truly assess their utility, a granular analysis of specific holiday periods is required.

A detailed breakdown of model performance on individual public holidays is provided in Figure \ref{fig:dma_holiday_bar}. These results reveal that no single holiday embedding strategy is universally superior and that the optimal approach is highly stochastic. However, at least one of the feature sets with holiday information outperforms the one without for almost all holidays. The difference in performance is also not consistent across holidays. High-impact holidays such as Christmas and New Year's Day, the baseline model consistently exhibits the highest error, often failing to anticipate the demand fluctuation. The introduction of explicit holiday features significantly mitigates this. For instance, in \acrshort{dma} A and C, the categorical embedding ($\mathbf{h}_{categorical}$) substantially reduces \acrshort{mae} during the Christmas period. In \acrshort{dma} B, the lagged feature ($\mathbf{h}_{lagged}$) proves superior.

Furthermore, the box plots in Figure \ref{fig:dma_holiday_box} provide reassurance regarding model stability. The error distribution for non-holidays remains virtually identical across all configurations, confirming that the addition of sparse holiday features does not introduce noise or degrade performance on standard operational days. In contrast, for holiday days, the proposed methods frequently lower the median error and compress the interquartile range compared to the baseline. While this confirms the model's ability to learn rare event positively, this also raises question due to the stochasticity of forecasting performance. This stochasticity could arise from the different types of holidays, the day of occurrence of the event, season of occurrence, etc., and requires more attention.

\subsubsection{Comparison with State-of-the-Art Methods -  Experiment V}

To rigorously evaluate the performance of our proposed model, we conducted a comprehensive benchmark across all three \acrshort{dmas}. The quantitative results, tabulated in Table \ref{tab:heat_quantitative_metrics}, unequivocally demonstrate the superior performance of our proposed approach.
The proposed model consistently outperforms all baselines, achieving the lowest error metrics across every district. In \acrshort{dma} A, the proposed framework achieves an \acrshort{mae} of $99.76$ kWh, representing a massive reduction of approximately $42.5\%$ compared to the best-performing baseline, TimeMixer ($173.63$ kWh). Similar dominance is observed in the other districts, with error reductions of roughly $36\%$ for \acrshort{dma} B ($164.81$ vs $257.28$ kWh) and $43\%$ for \acrshort{dma} C ($152.60$ vs $269.38$ kWh) relative to the strongest competitors.

Interestingly, the benchmark reveals a shift in the landscape of time-series forecasting. While Transformer-based models like PatchTST and TimesNet perform well, simpler MLP-based architectures such as DLinear, TSMixer, and TimeMixer consistently emerge as the strongest baselines, surpassing both classical methods (SARIMAX, XGBoost) and complex attention mechanisms (Informer, Autoformer). This reinforces recent findings in the literature questioning the necessity of Transformers for certain time-series tasks. However, even these advanced Mixer architectures struggle to match the precision of our proposed Wavelet-CNN framework.

\begin{figure}[htbp]
    \centering
    \begin{subfigure}{0.65\linewidth}
        \includegraphics[width=\linewidth]{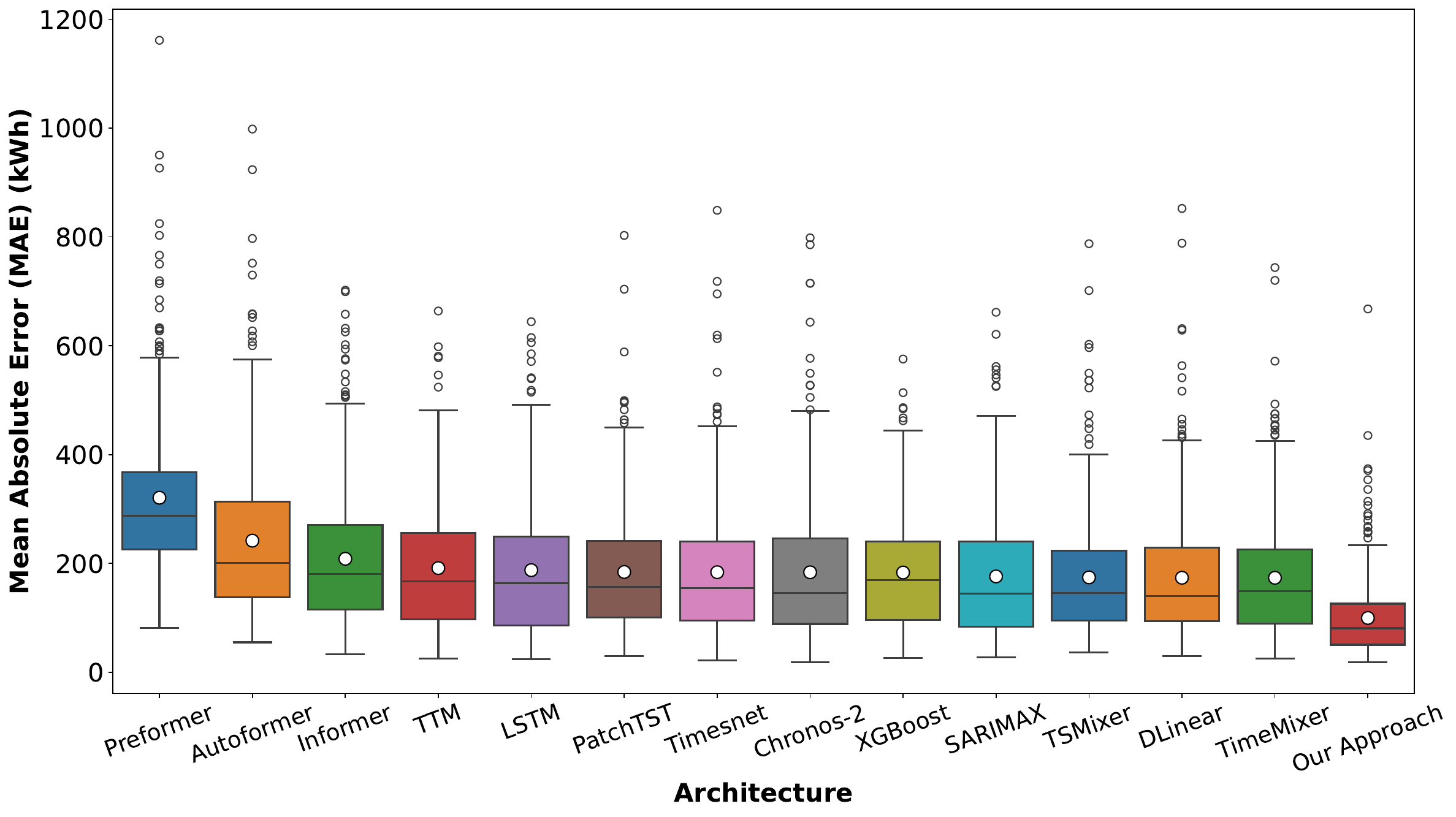}
        \caption{\acrshort{dma} A}
        \label{fig:dma_a_sota}
    \end{subfigure}
    
    \vspace{0.5cm} 
    \begin{subfigure}{0.65\linewidth}
        \includegraphics[width=\linewidth]{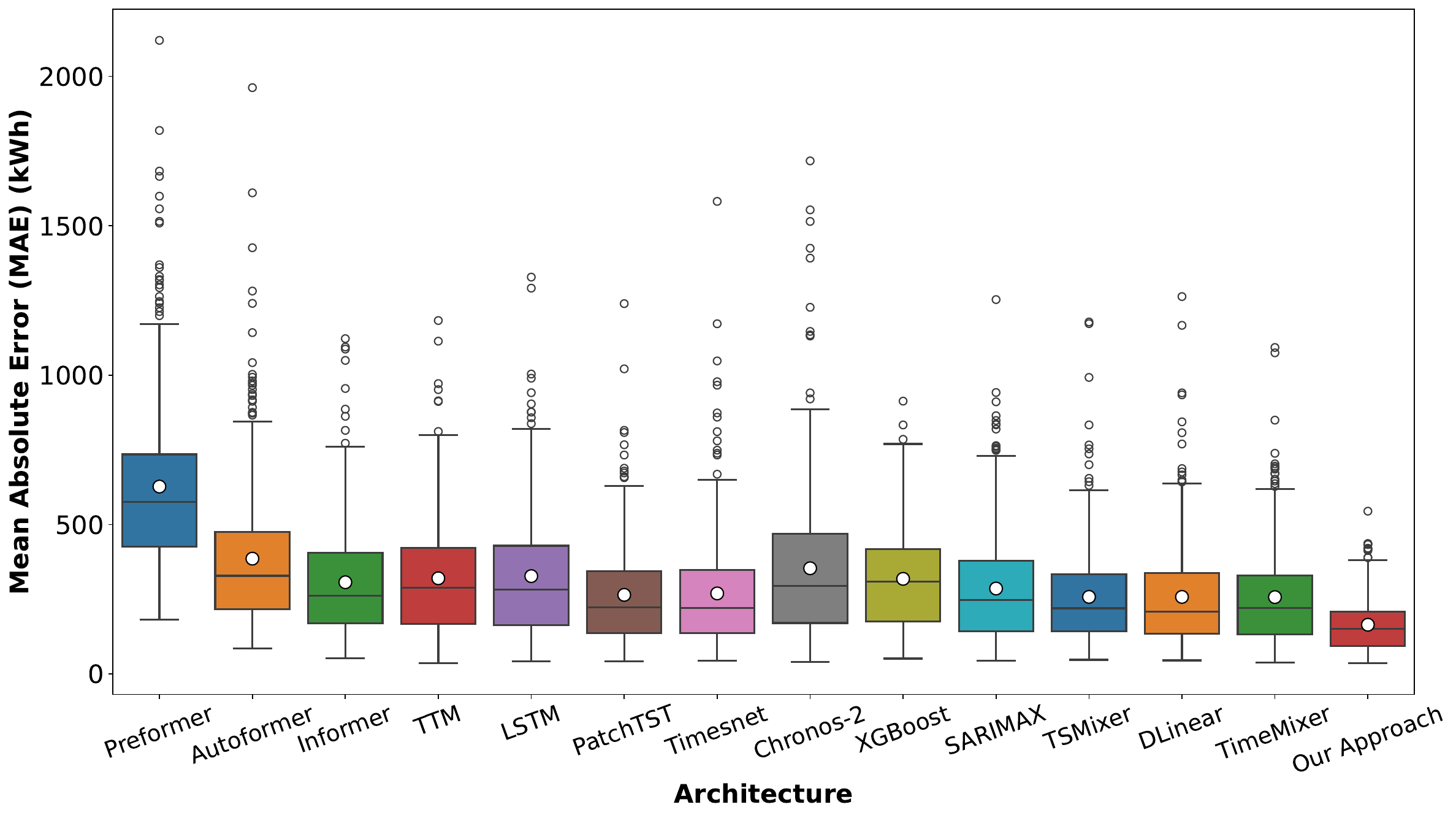}
        \caption{\acrshort{dma} B}
        \label{fig:dma_b_sota}
    \end{subfigure}
    
    \vspace{0.5cm} 
    \begin{subfigure}{0.65\linewidth}
        \includegraphics[width=\linewidth]{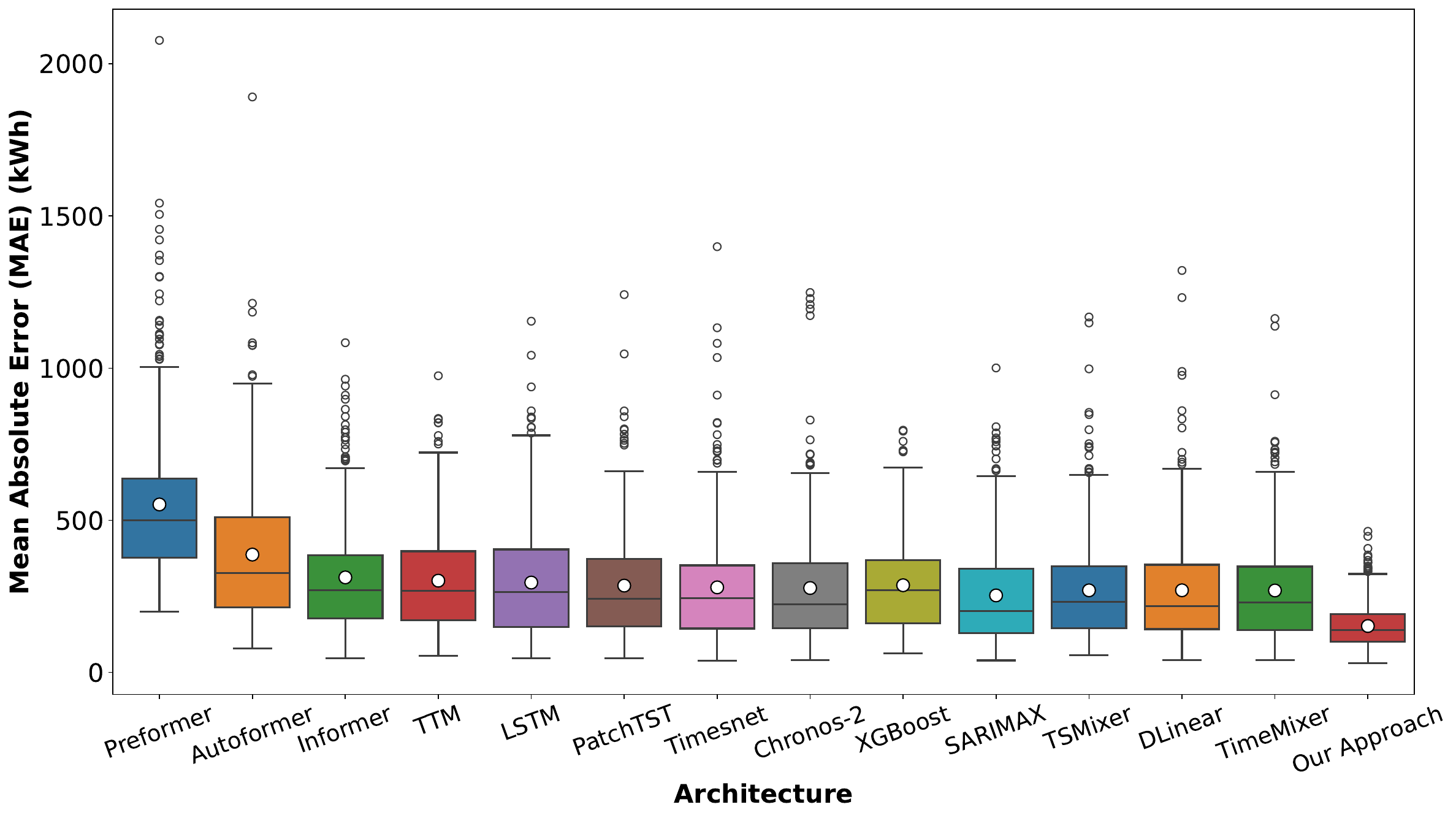}
        \caption{\acrshort{dma} C}
        \label{fig:dma_c_sota}
    \end{subfigure}
    
    \caption{Quantitative comparison of \acrshort{mae} of different models for the respective \acrshort{dmas} of (a) \acrshort{dma} A, (b) \acrshort{dma} B, and (c) \acrshort{dma} C. The proposed approach (far right) demonstrates lower median error and reduced variance compared to \acrshort{dl} and statistical baselines.}
    \label{fig:sota_box_plots}
\end{figure}

The box plots in Figure \ref{fig:sota_box_plots} further substantiate these findings. Our model (far right) not only has the lowest mean error (white dot) but also the most compact interquartile range and a better handling of outliers. In contrast, models like Preformer and Autoformer exhibit extreme variance, making them unreliable for grid operations where worst-case performance is critical. The foundation models (Chronos-2 and Fine-tuned TTM), while promising, currently lag behind purpose-built architectures for this specific high-frequency domain.

\begin{table}[htbp]
\caption{Quantitative results comparing existing statistical, state-of-the-art methods, and foundation models for hourly heat demand forecasting with our proposed approach for a forecasting horizon of 24 hours.     
\acrshort{mae} is in \text{kWh}, \acrshort{mape} in \%, and \acrshort{mse} in $10^{5}\text{kWh}^2$.}
\begin{center}
\begin{tabular}{c|c|c|c|c|c|c|c|c|c} 
\hline
\multirow{2}{*}{\textbf{Architecture}} & \multicolumn{3}{c|}{\textbf{\acrshort{dma} A}} & \multicolumn{3}{c|}{\textbf{\acrshort{dma} B}} & \multicolumn{3}{c}{\textbf{\acrshort{dma} C}} \\
\cline{2-10}
& \textbf{\acrshort{mae}} & \textbf{\acrshort{mape}} & \textbf{\acrshort{mse}} 
& \textbf{\acrshort{mae}} & \textbf{\acrshort{mape}} & \textbf{\acrshort{mse}} 
& \textbf{\acrshort{mae}} & \textbf{\acrshort{mape}} & \textbf{\acrshort{mse}} \\
\hline
Preformer     & $320.68$ & $21.37$ & $1.69$ & $627.20$ & $21.12$ & $6.37$ & $552.04$ & $19.34$ & $4.99$ \\
Autoformer    & $241.86$ & $11.52$ & $1.03$ & $385.94$ & $12.51$ & $2.62$ & $387.36$ & $11.97$ & $2.61$ \\
Informer      & $208.51$ & $9.42$  & $0.79$ & $306.92$ & $9.30$  & $1.75$ & $312.38$ & $9.23$  & $1.77$ \\
TTM Fine-tuned& $191.47$ & $10.08$ & $0.73$ & $320.25$ & $8.28$  & $1.98$ & $301.66$ & $8.49$  & $1.71$ \\
LSTM          & $187.62$ & $9.56$  & $0.71$ & $327.49$ & $8.70$  & $2.06$ & $295.44$ & $8.32$  & $1.67$ \\
PatchTST      & $184.27$ & $8.44$  & $0.65$ & $264.71$ & $8.04$  & $1.39$ & $285.42$ & $8.34$  & $1.57$ \\
TimesNet      & $183.89$ & $8.09$  & $0.67$ & $269.62$ & $8.08$  & $1.48$ & $279.60$ & $7.91$  & $1.55$ \\
Chronos-2     & $183.71$ & $9.38$  & $0.73$ & $353.80$ & $8.89$  & $2.60$ & $277.50$ & $7.81$  & $1.57$ \\
XGBoost       & $183.25$ & $9.81$  & $0.67$ & $318.31$ & $8.58$  & $1.85$ & $286.56$ & $8.37$  & $1.52$ \\
SARIMAX       & $176.08$ & $8.96$  & $0.61$ & $285.97$ & $7.20$  & $1.69$ & $253.16$ & $7.13$  & $1.28$ \\
TSMixer       & $174.32$ & $7.91$  & $0.60$ & $258.00$ & $7.83$  & $1.31$ & $269.89$ & $7.89$  & $1.42$ \\
DLinear       & $173.71$ & $7.93$  & $0.60$ & $257.88$ & $7.90$  & $1.32$ & $270.02$ & $7.91$  & $1.45$ \\
TimeMixer     & $173.63$ & $7.81$  & $0.60$ & $257.28$ & $7.80$  & $1.31$ & $269.38$ & $7.80$  & $1.44$ \\
\textbf{Proposed} & $\mathbf{99.76}$ & $\mathbf{5.26}$ & $\mathbf{0.24}$ 
                  & $\mathbf{164.81}$ & $\mathbf{4.44}$ & $\mathbf{0.53}$ 
                  & $\mathbf{152.60}$ & $\mathbf{4.65}$ & $\mathbf{0.44}$ \\
\hline
\end{tabular}
\label{tab:heat_quantitative_metrics}
\end{center}
\end{table}

\begin{figure}[htbp]
    \centering
    \begin{subfigure}{0.45\linewidth}
        \includegraphics[width=\linewidth]{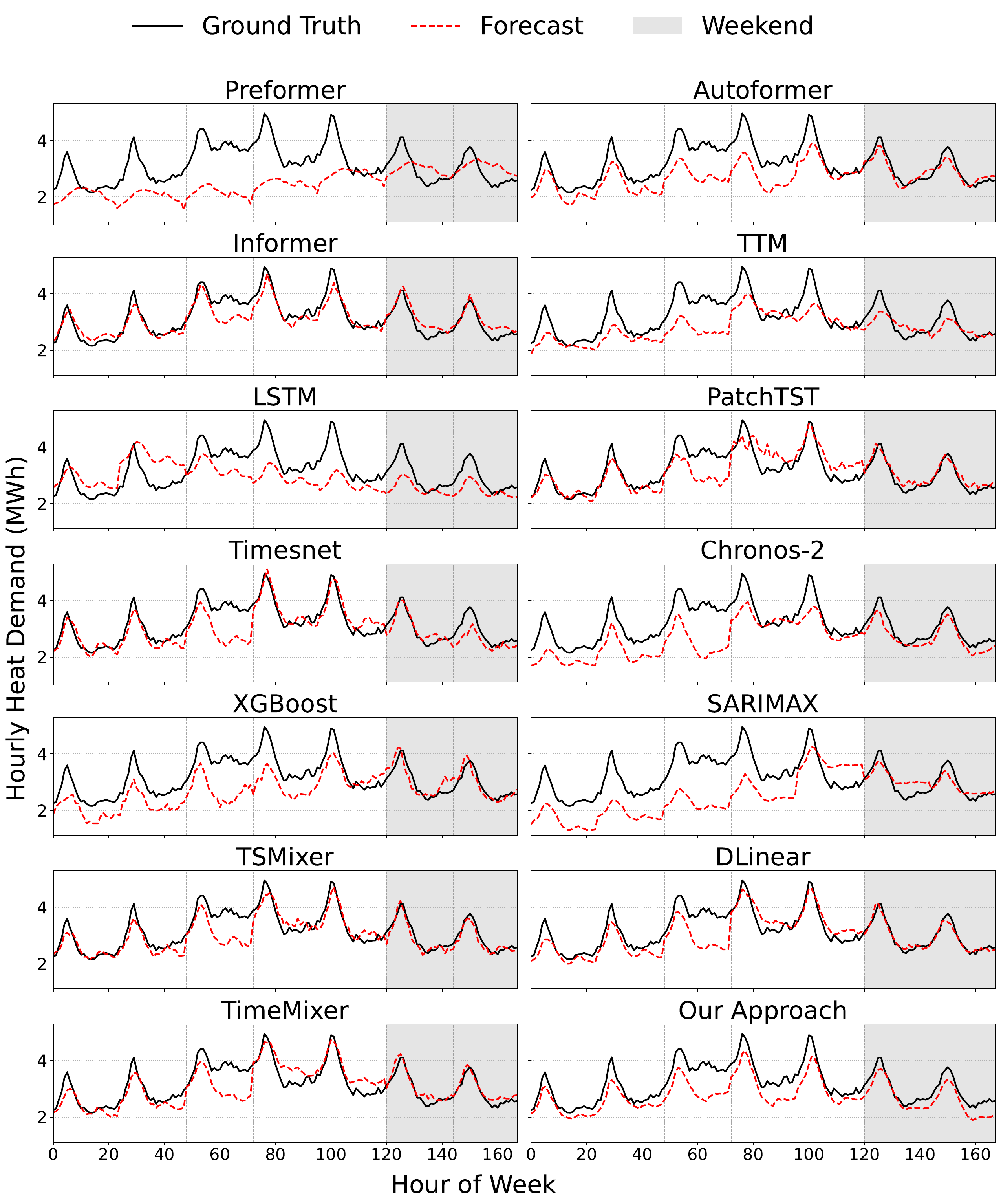}
        \caption{\acrshort{dma} A}
        \label{fig:all_models_worst_dma_a}
    \end{subfigure}
    \vspace{0.5cm} 
    \begin{subfigure}{0.45\linewidth}
        \includegraphics[width=\linewidth]{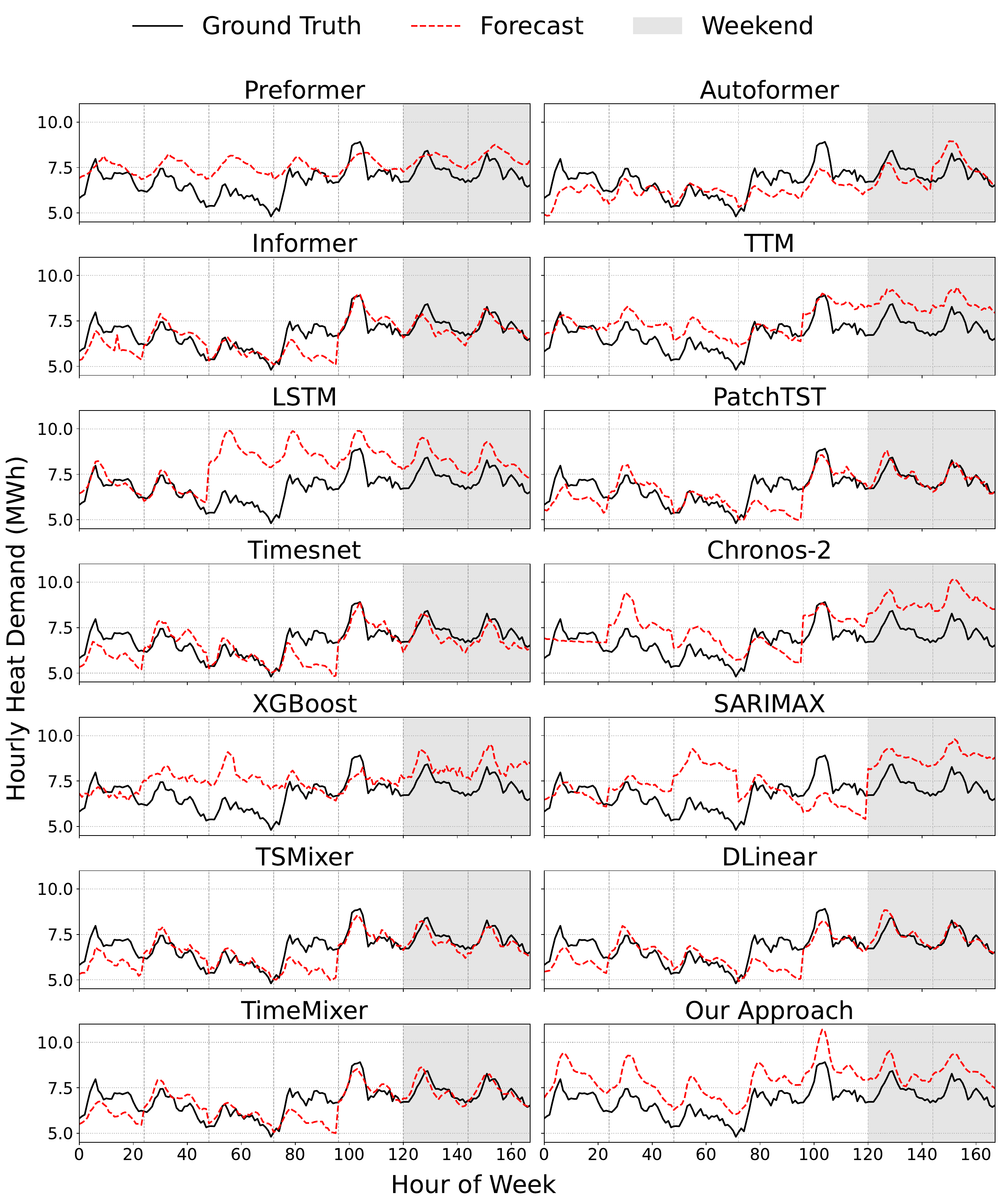}
        \caption{\acrshort{dma} B}
        \label{fig:all_models_worst_dma_b}
    \end{subfigure}
    \vspace{0.5cm} 

    \begin{subfigure}{0.45\linewidth}
        \includegraphics[width=\linewidth]{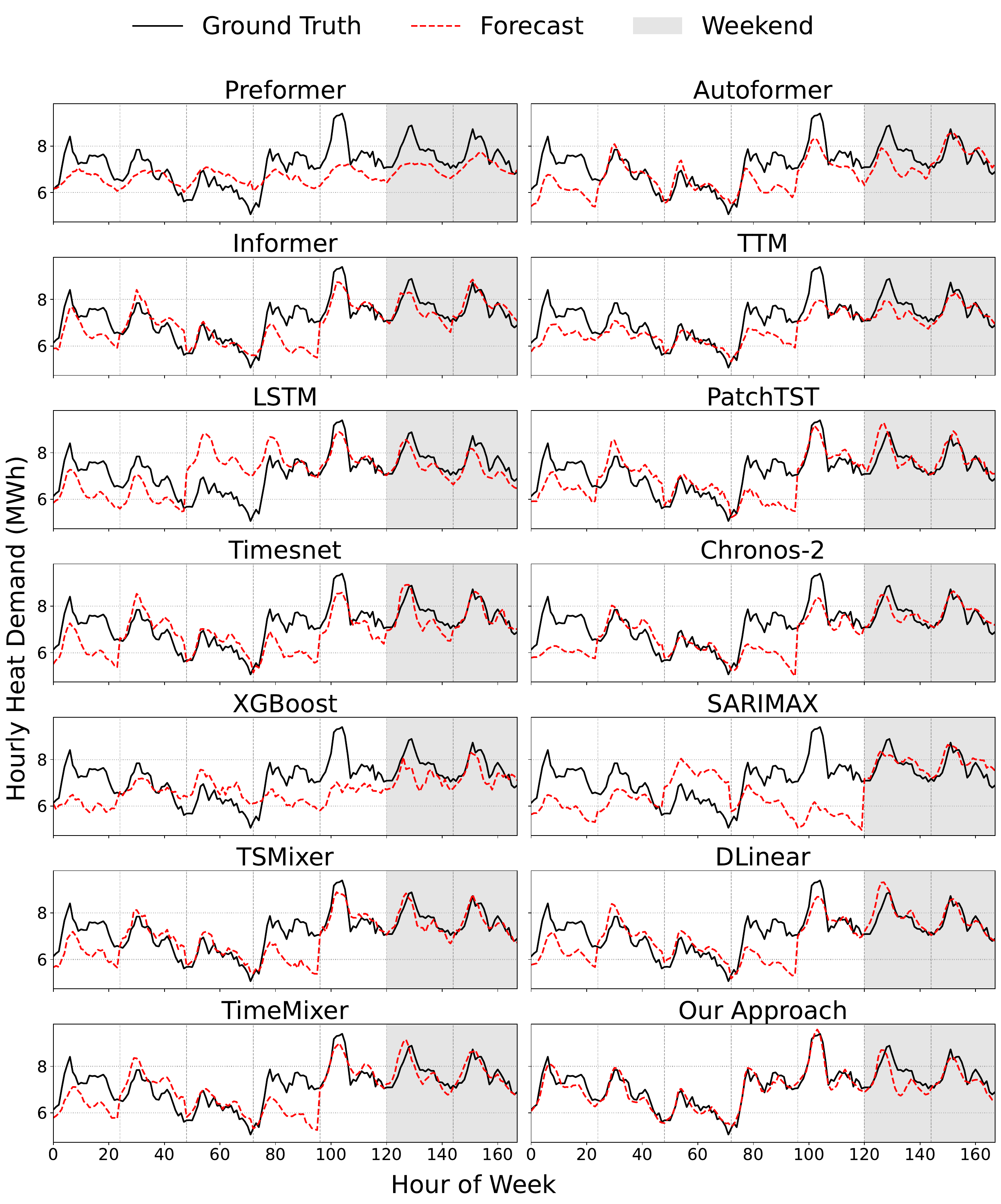}
        \caption{\acrshort{dma} C}
        \label{fig:all_models_worst_dma_c}
    \end{subfigure}
    
    \caption{Qualitative forecast comparison for the week with the highest error across all three districts: (a) \acrshort{dma} A (2019-04-08 to 2019-04-14), (b) \acrshort{dma} B (2019-01-14 to 2019-01-20), and (c) \acrshort{dma} C (2019-01-14 to 2019-01-20). These plots highlight model performance during highly volatile demand periods, showcasing the robustness of our approach compared to baselines.}
    \label{fig:forecast_worst_all_dmas_all_models}
\end{figure}

\begin{figure}[htbp]
    \centering
    \begin{subfigure}{0.45\linewidth}
        \includegraphics[width=\linewidth]{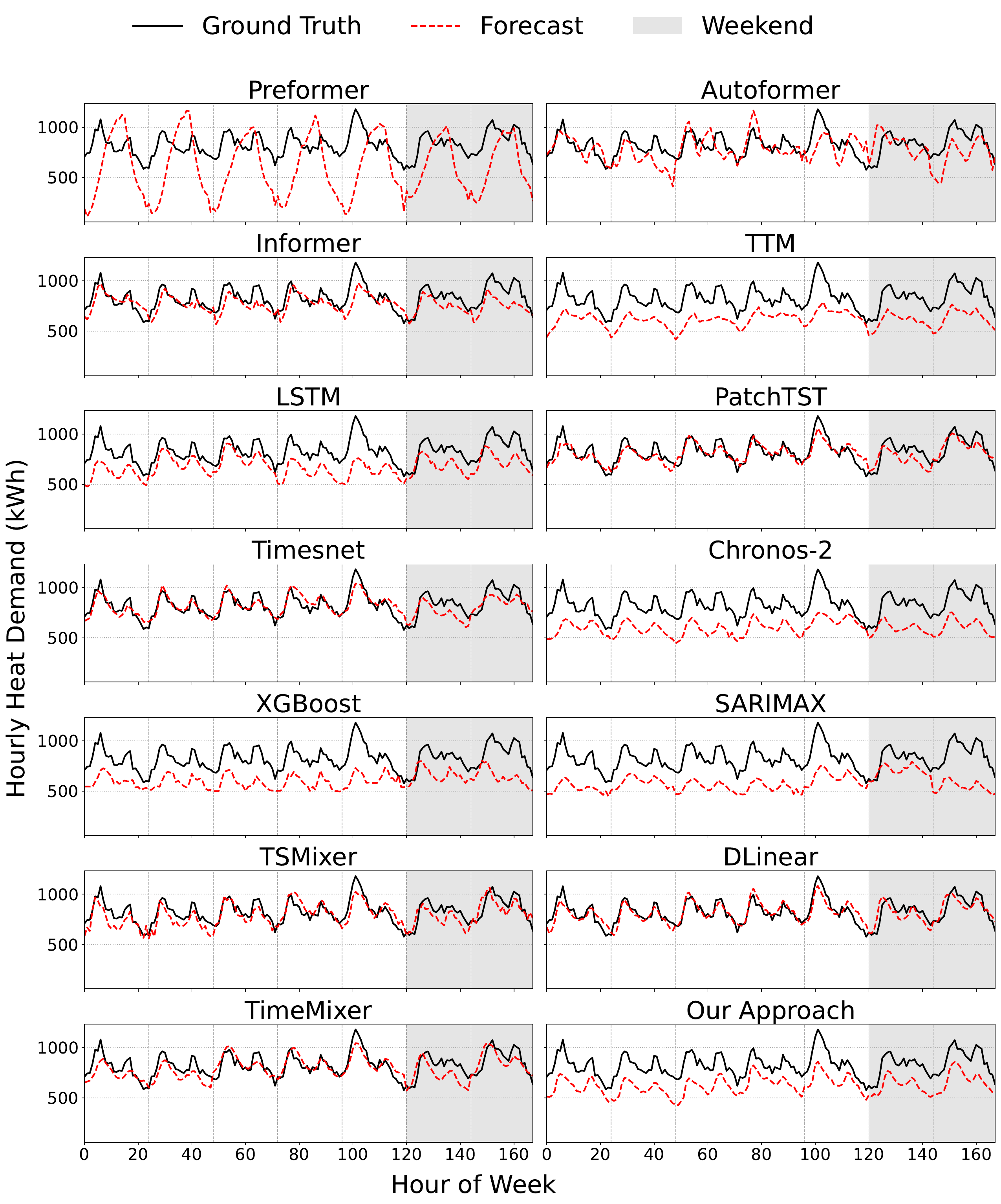}
        \caption{\acrshort{dma} A}
        \label{fig:all_models_best_dma_a}
    \end{subfigure}
    \vspace{0.5cm} 
    \begin{subfigure}{0.45\linewidth}
        \includegraphics[width=\linewidth]{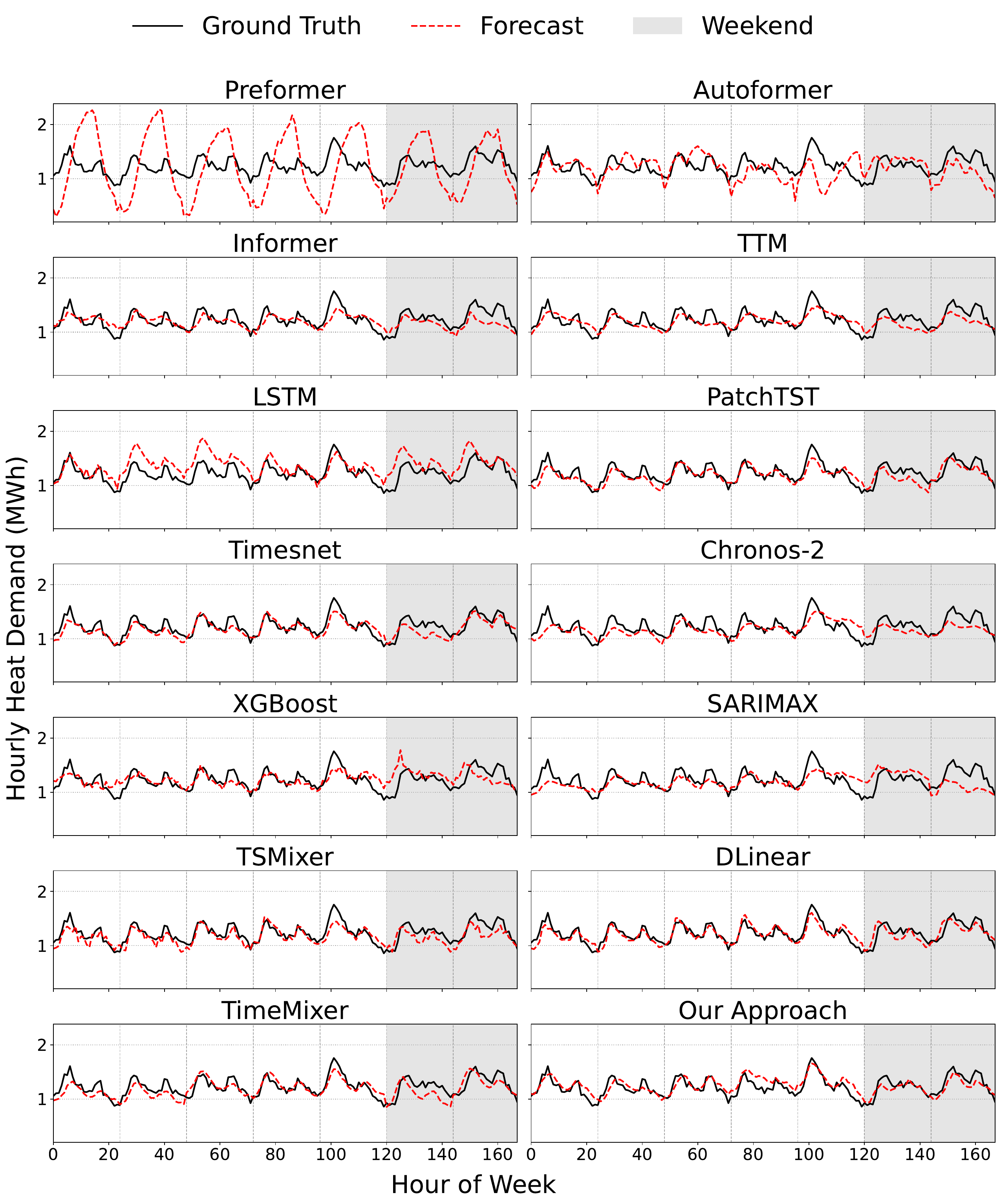}
        \caption{\acrshort{dma} B}
        \label{fig:all_models_best_dma_b}
    \end{subfigure}
    \vspace{0.5cm} 
    
    \begin{subfigure}{0.48\linewidth}
        \includegraphics[width=\linewidth]{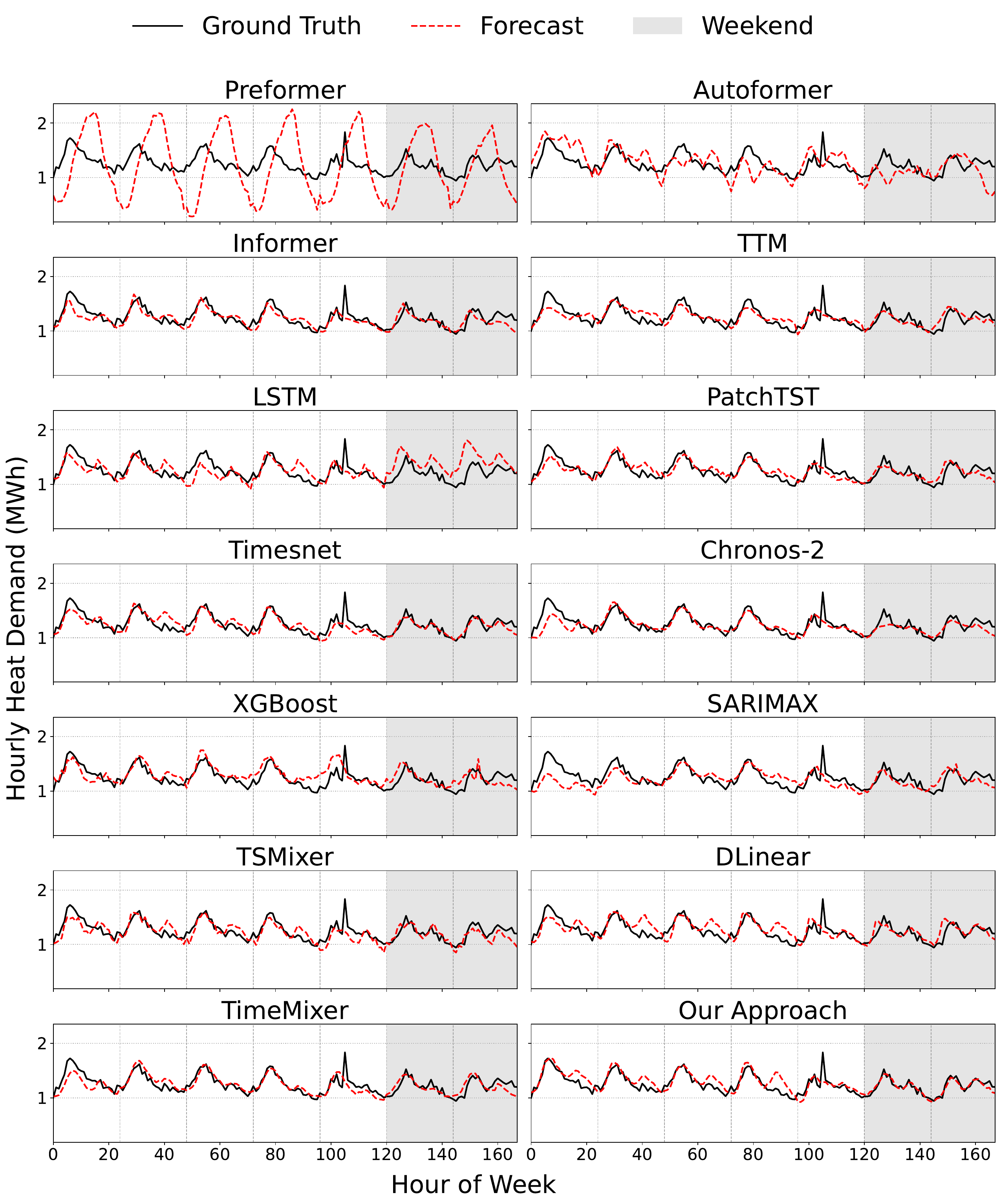}
        \caption{\acrshort{dma} C}
        \label{fig:all_models_best_dma_c}
    \end{subfigure}
    
    \caption{Qualitative forecast comparison for the week with the lowest error across all three districts: (a) \acrshort{dma} A (2019-08-05 to 2019-08-11), (b) \acrshort{dma} B (2019-08-05 to 2019-08-11), and (c) \acrshort{dma} C (2019-07-15 to 2019-07-21). These plots show performance during stable, predictable demand, where all models perform reasonably well.}
    \label{fig:forecast_best_all_dmas_all_models}
\end{figure}

\begin{figure}[htbp]
    \centering
    \begin{subfigure}{0.32\linewidth}
        \includegraphics[width=\linewidth]{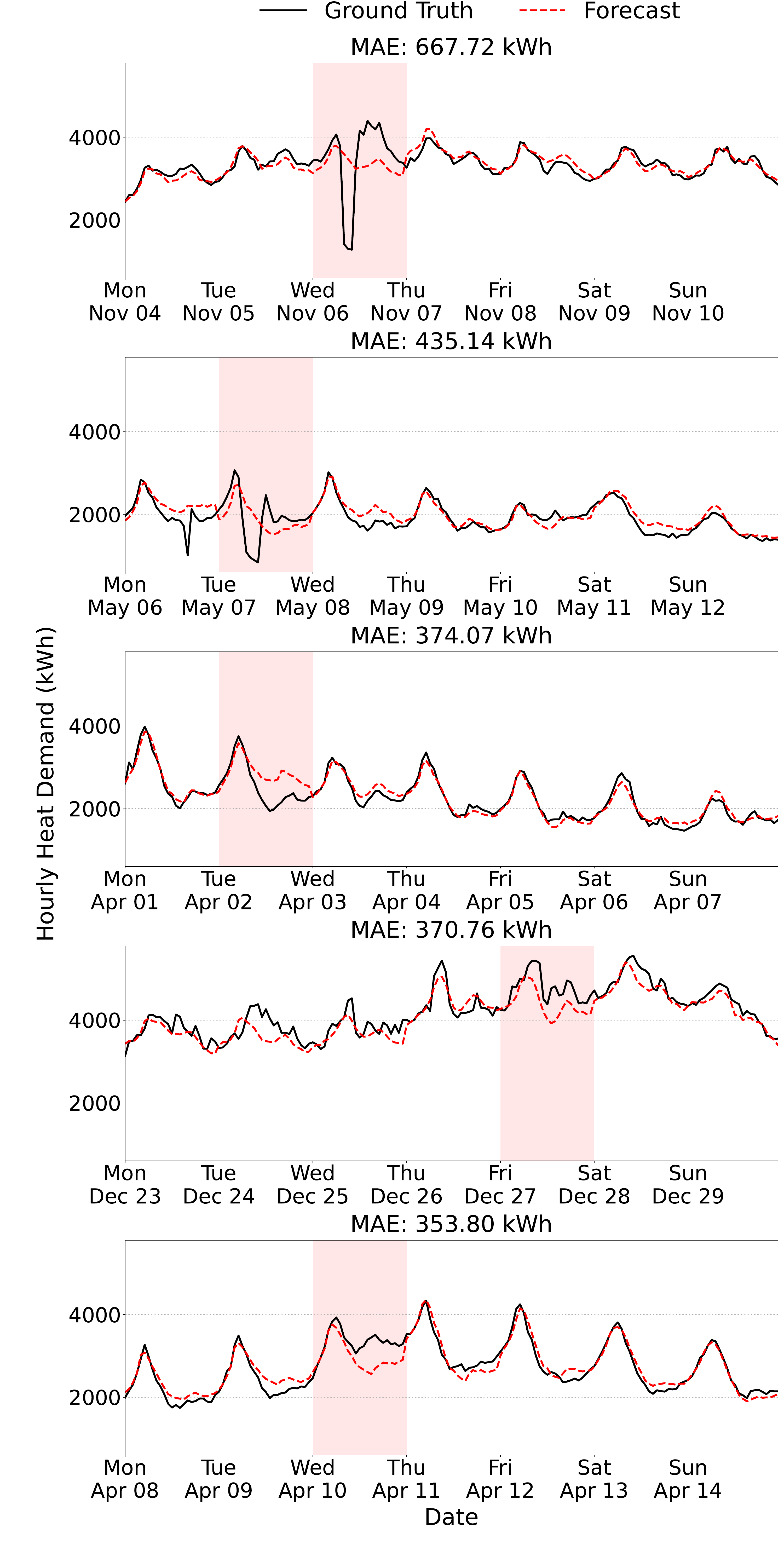}
        \caption{\acrshort{dma} A}
        \label{fig:worst_dma_a}
    \end{subfigure}
    \vspace{0.5cm} 
    \begin{subfigure}{0.32\linewidth}
        \includegraphics[width=\linewidth]{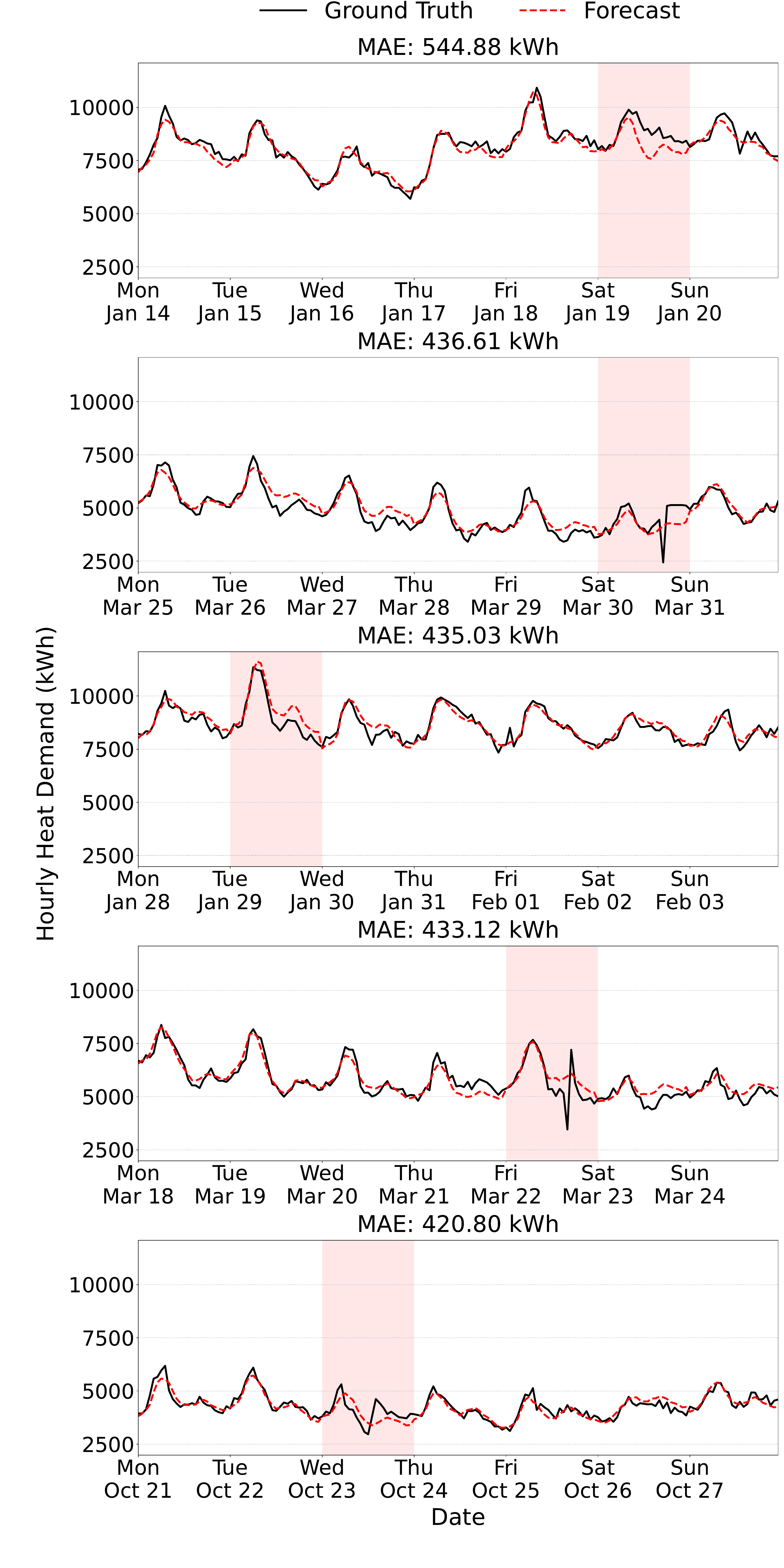}
        \caption{\acrshort{dma} B}
        \label{fig:worst_dma_b}
    \end{subfigure}
    \vspace{0.5cm} 
    \begin{subfigure}{0.32\linewidth}
        \includegraphics[width=\linewidth]{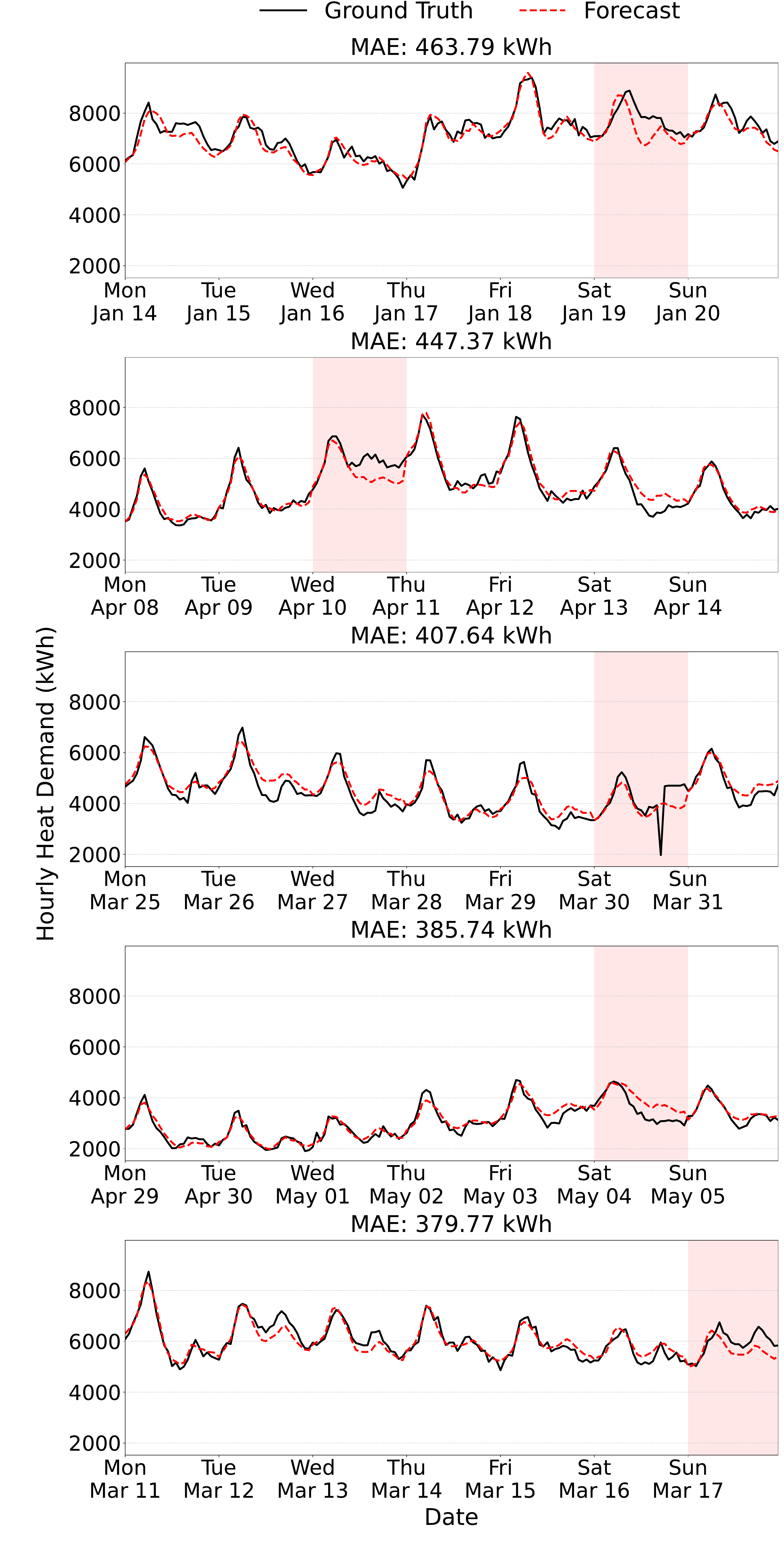}
        \caption{\acrshort{dma} C}
        \label{fig:worst_dma_c}
    \end{subfigure}
    
    \caption{Qualitative forecast comparison for the top five days of worst forecasts across all three districts: (a) \acrshort{dma} A, (b) \acrshort{dma} B, and (c) \acrshort{dma} C.}
    \label{fig:forecast_worst_all_dmas}
\end{figure}

\begin{figure}[htbp]
    \centering
    \begin{subfigure}{0.32\linewidth}
        \includegraphics[width=\linewidth]{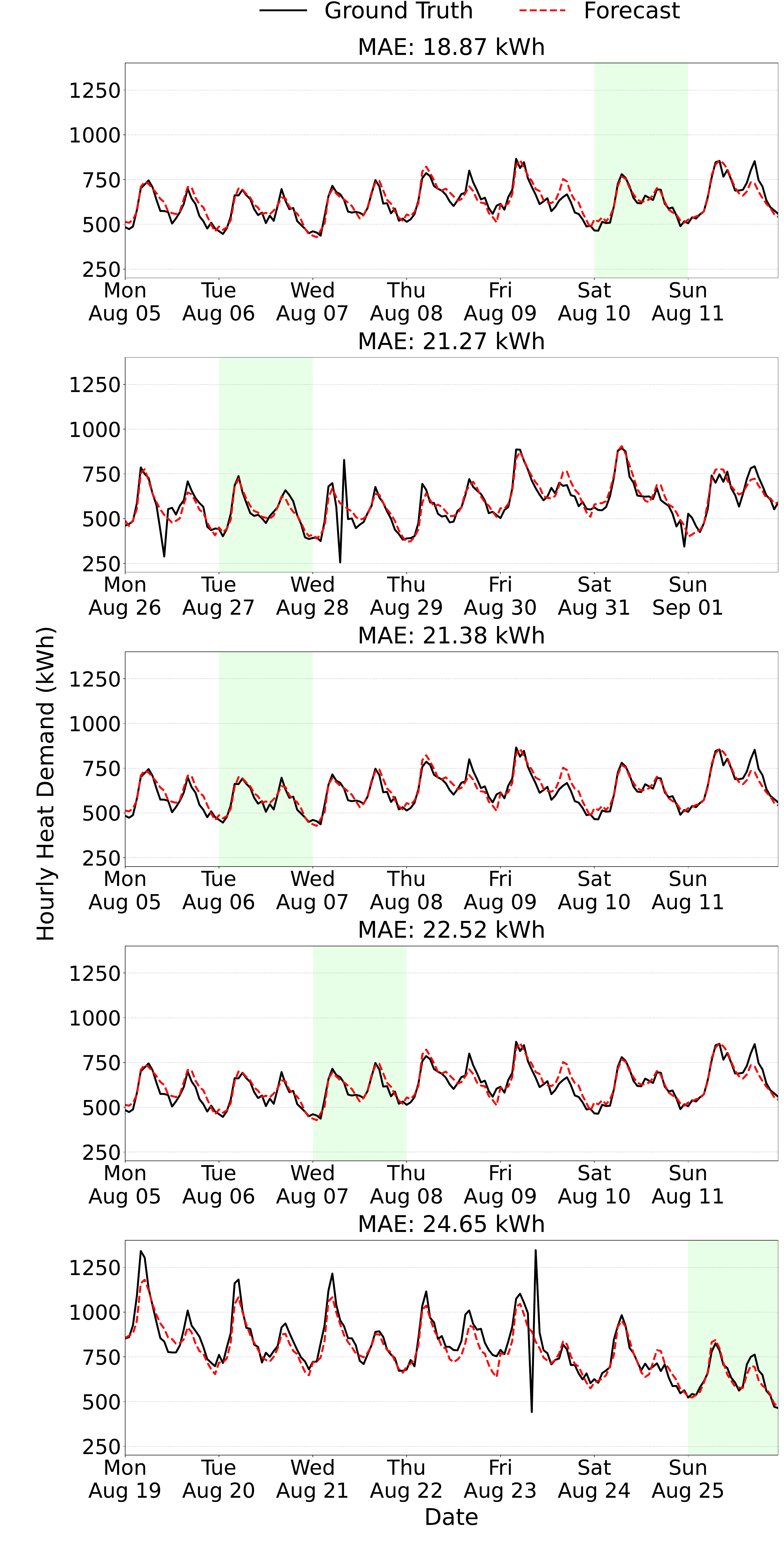}
        \caption{\acrshort{dma} A}
        \label{fig:best_dma_a}
    \end{subfigure}
    \vspace{0.5cm} 
    \begin{subfigure}{0.32\linewidth}
        \includegraphics[width=\linewidth]{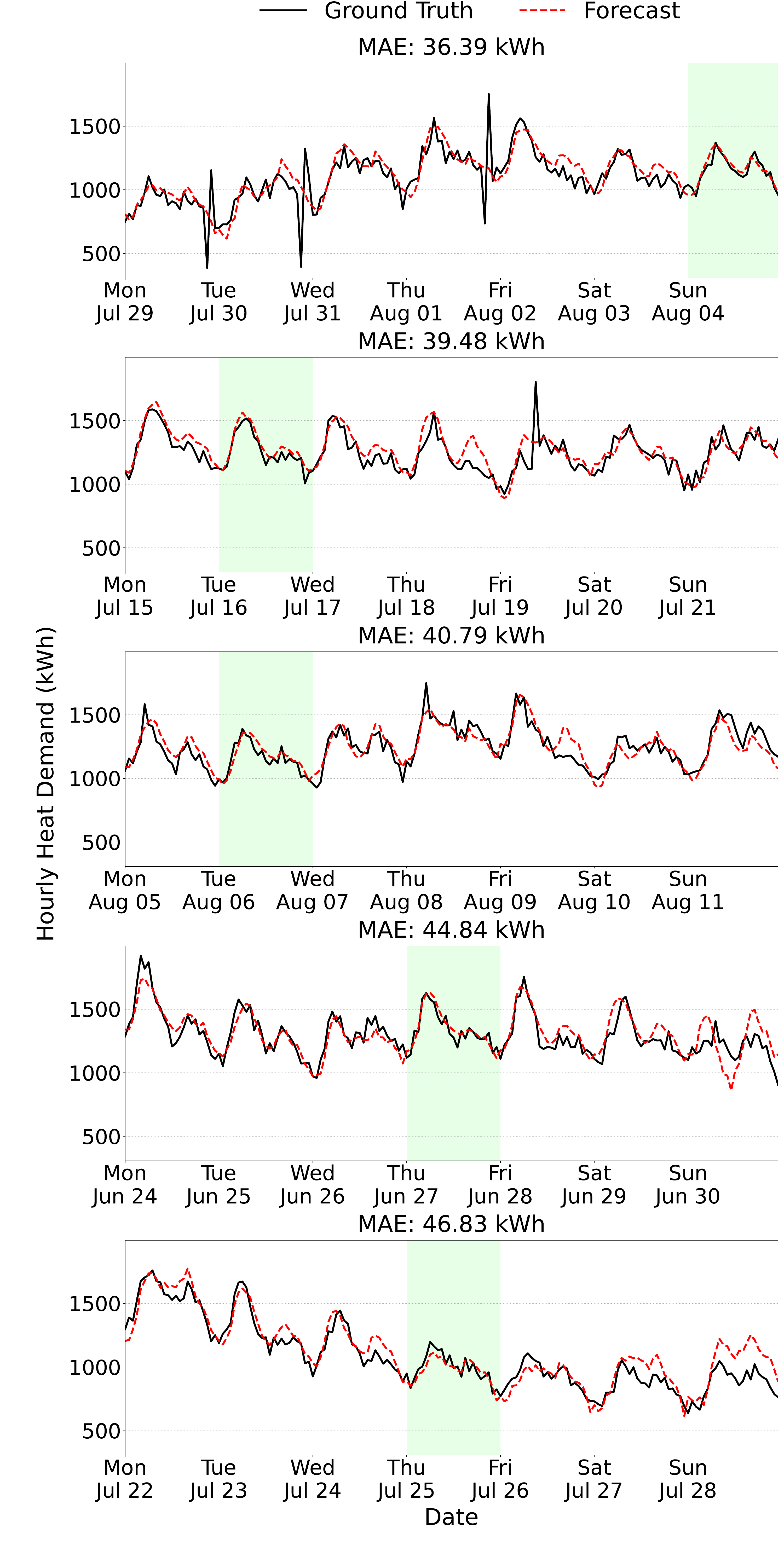}
        \caption{\acrshort{dma} B}
        \label{fig:best_dma_b}
    \end{subfigure}
    \vspace{0.5cm} 
    \begin{subfigure}{0.32\linewidth}
        \includegraphics[width=\linewidth]{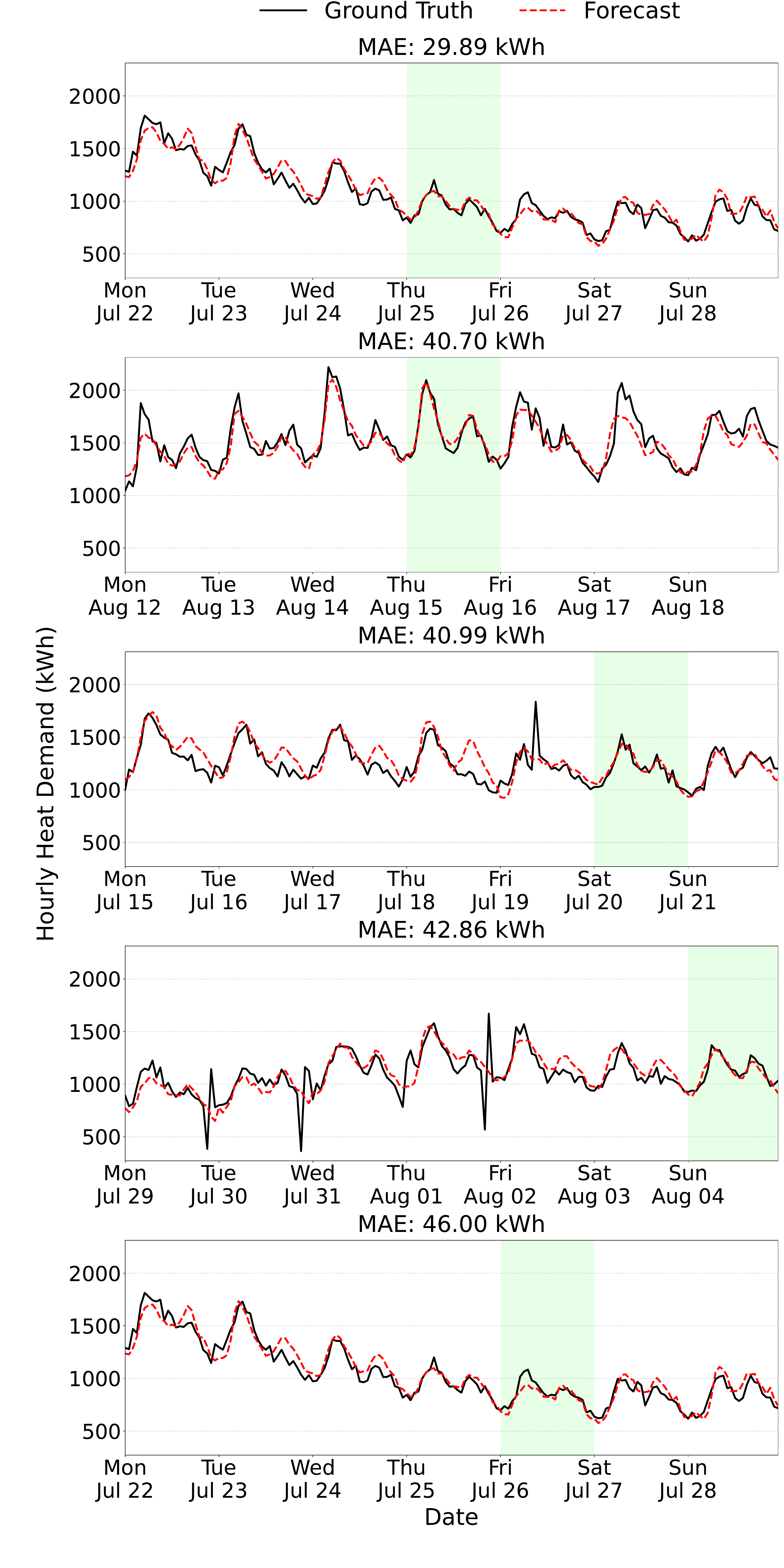}
        \caption{\acrshort{dma} C}
        \label{fig:best_dma_c}
    \end{subfigure}
    
    \caption{Qualitative forecast comparison for the top five days of best forecasts across all three districts: (a) \acrshort{dma} A, (b) \acrshort{dma} B, and (c) \acrshort{dma} C.}
    \label{fig:forecast_best_all_dmas}
\end{figure}

While these aggregate metrics provide a clear performance ranking, we perform further qualitative analysis by visualising forecast predictions against the ground truth for representative weeks. To identify these weeks, we selected scenarios representing the lower and higher ends of the collective error spectrum. This is achieved by normalizing the weekly \acrshort{mae} for each model and aggregating these scores. Lower sums indicate weeks that are collectively easier to forecast, whereas higher sums correspond to volatile weeks that challenge all architectures.

The forecast plots for the high error week, as seen in Figure \ref{fig:forecast_worst_all_dmas_all_models}, reveal that the models of Preformer and Autoformer produce overly smoothed forecasts, failing to capture the sharp daily peaks and troughs that are critical for operational planning. Other methods, such as \acrshort{sarimax} and XGBoost, while capturing some volatility, often suffer from significant phase lag or large deviation between predictions and actual demand, over-predicting off-peak demand and under-predicting peaks.  Modern MLP-based architectures (DLinear, TimeMixer) along with the proposed method capture the phase alignment, and the volatility. The proposed approach, juxtaposed with the presented models, exhibits the ability to capture change in behaviour as seen in \ref{fig:all_models_worst_dma_a} and  \ref{fig:all_models_worst_dma_b} which deviates from the trend of lagged features. This resilience is attributed to the time-frequency and disentangled representation of the trend components that allows the model to anticipate ramp-events based on exogenous drivers, while the residual component enables the reconstruction of high-frequency volatility.

In Figure \ref{fig:forecast_best_all_dmas_all_models}, representing the week with the lower errors, characterized by more regular and predictable demand patterns, the performance gap between models naturally narrows. However, even under these favorable conditions, subtle differences persist. The baseline models still tend to produce smoothing artifacts or minor lags. Our approach, however, maintains the tightest fit to the ground truth, successfully distinguishing between stochastic noise (common in summer domestic hot water profiles) and structural signal.

To assess the operational limits of the framework, we examine the specific days with the highest and lowest \acrshort{mae} for the proposed model in Figure \ref{fig:forecast_worst_all_dmas} and Figure \ref{fig:forecast_best_all_dmas}. The worst-case scenarios (Figure \ref{fig:forecast_worst_all_dmas}) are particularly revealing. The largest errors often correspond to abrupt operational anomalies, such as the near-vertical demand drop observed on Wednesday, Nov 06, in \acrshort{dma} A (likely a sensor fault or maintenance shutdown). In this instance, the model predicts the \textit{expected} normal demand profile rather than the anomaly. Crucially, this behaviour is a feature of robustness, not a failure in a real-world control loop, predicting the expected load ensures security of supply. A tendency to slightly over-predict during erratic system faults is operationally safer than under-predicting, which could risk insufficient heat delivery. Conversely, Figure \ref{fig:forecast_best_all_dmas} confirms that during standard operations, the model captures characteristic bimodal peaks with high precision, offering a dependable basis for automated grid optimisation.

\subsubsection{Time and Calendrical Features -  Experiment VI}

\begin{figure}[htbp]
    \centering
    \begin{subfigure}[b]{0.46\textwidth}
        \centering
        \includegraphics[width=\textwidth]{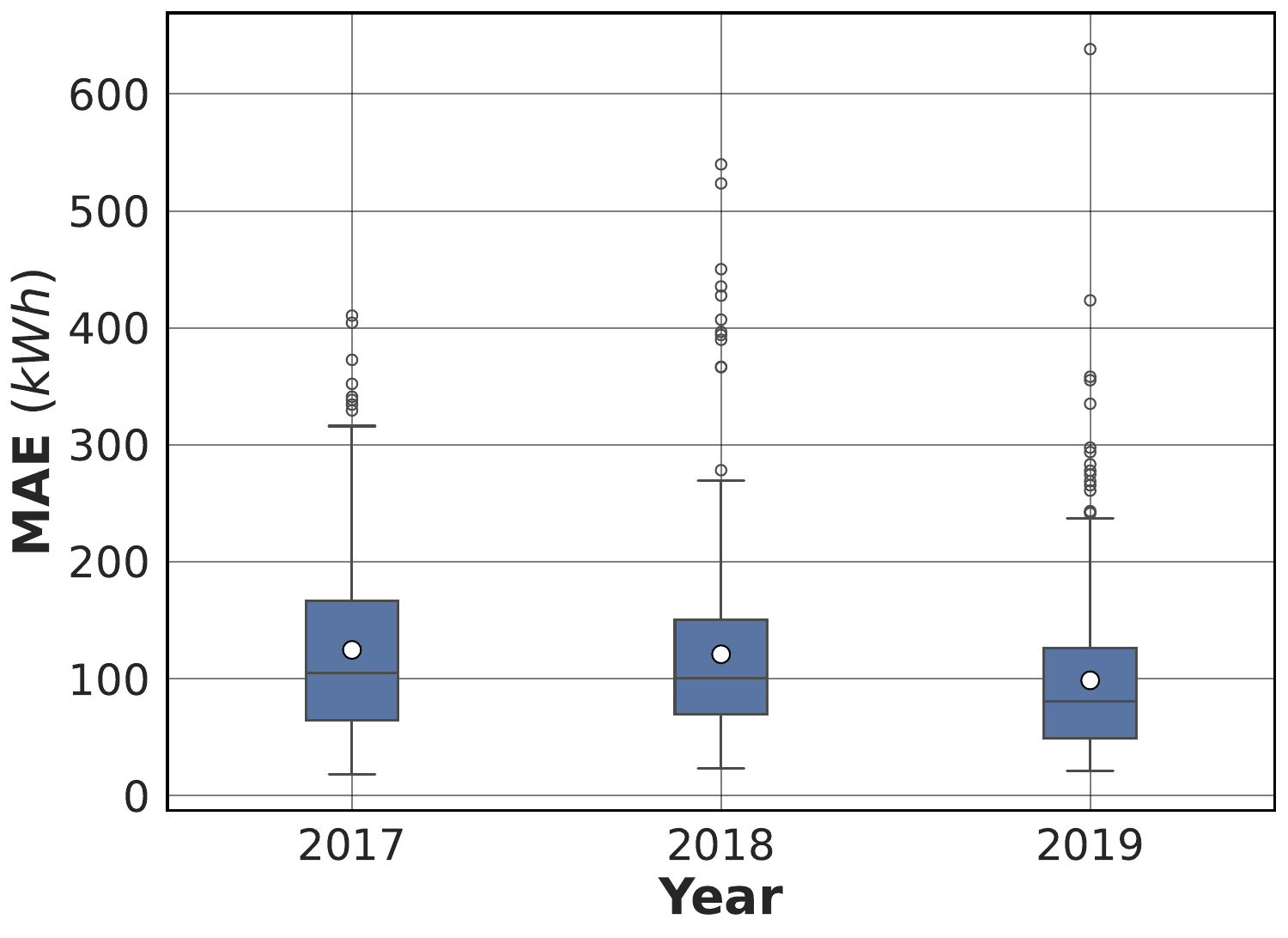} 
        \caption{DMA A}
        \label{fig:dma_a}
    \end{subfigure}
    \hfill
    \begin{subfigure}[b]{0.46\textwidth}
        \centering
        \includegraphics[width=\textwidth]{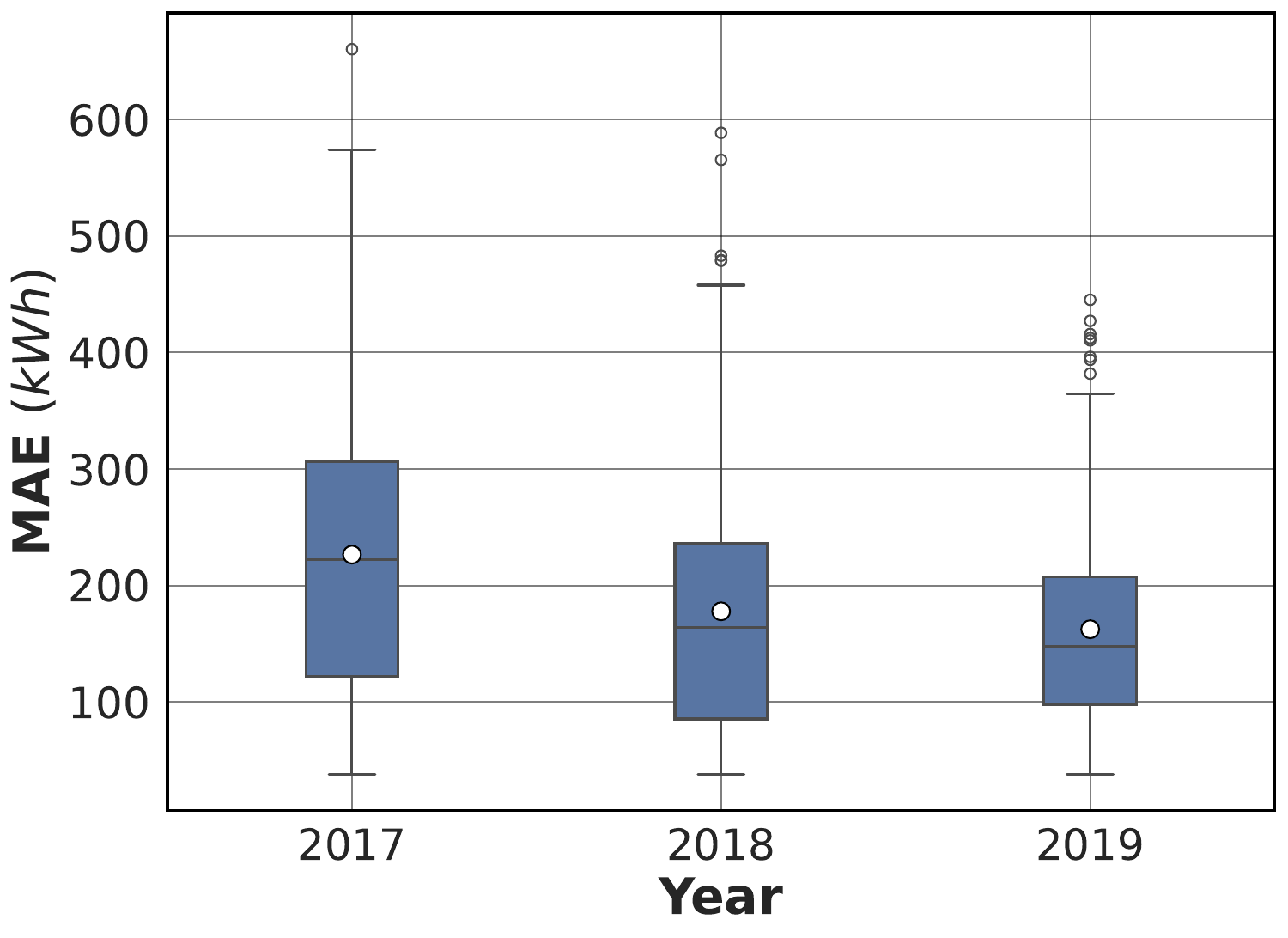} 
        \caption{DMA B}
        \label{fig:dma_b}
    \end{subfigure}

    \par\bigskip 

    \begin{subfigure}[b]{0.46\textwidth}
        \centering
        \includegraphics[width=\textwidth]{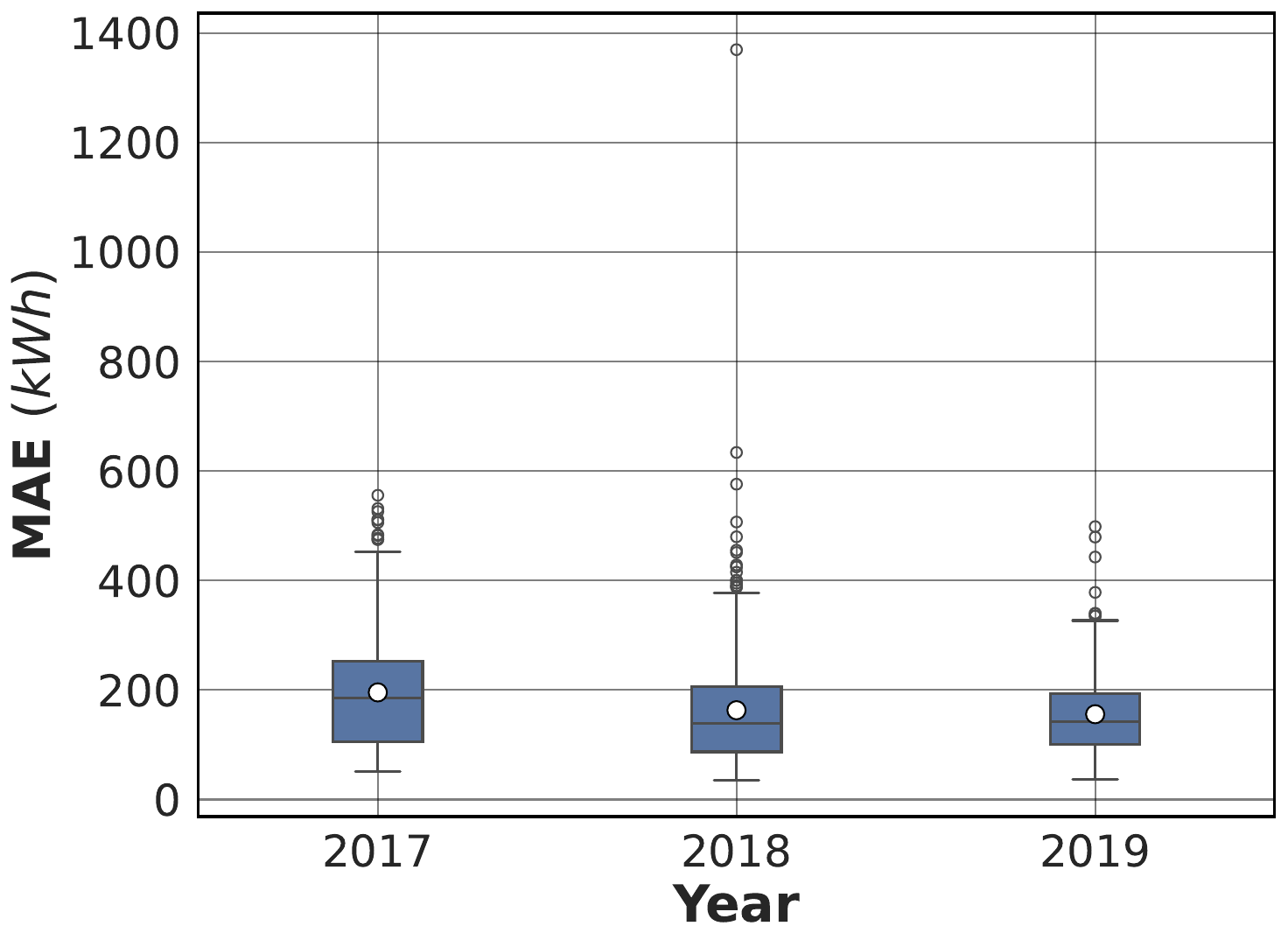} 
        \caption{DMA C}
        \label{fig:dma_c}
    \end{subfigure}
    
    \caption{Distribution of Mean Absolute Error (MAE) across different years for the three DMAs (top row) and the two city-level datasets (bottom row).}
    \label{fig:bronderslev_mae_boxplots}
\end{figure}

\begin{figure}[htbp]
    \centering
    \begin{subfigure}[b]{0.48\textwidth}
        \centering
        \includegraphics[width=\textwidth]{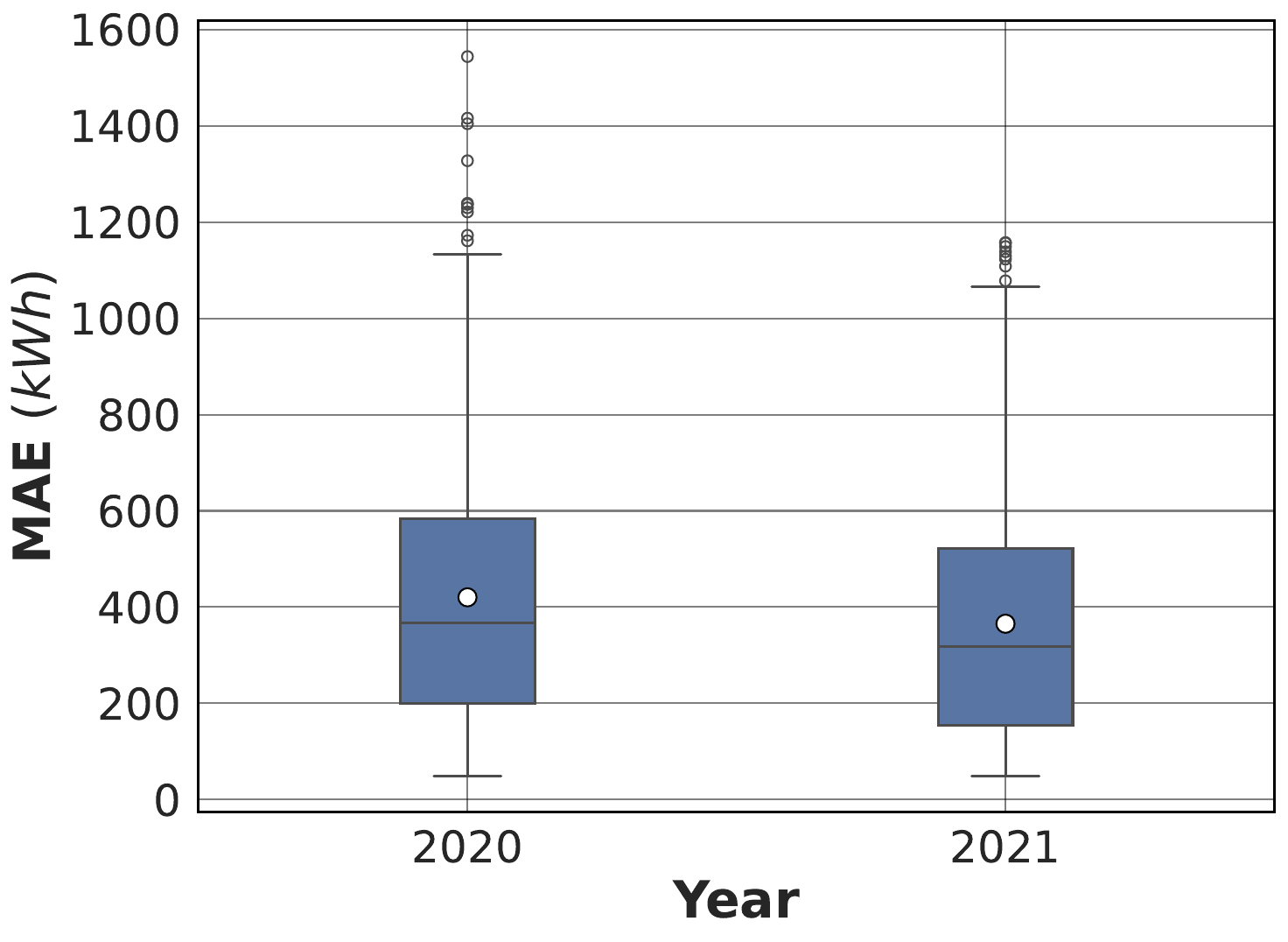} 
        \caption{Aalborg}
        \label{fig:rolling_mae_aalborg}
    \end{subfigure}
    \hfill
    \begin{subfigure}[b]{0.48\textwidth}
        \centering
        \includegraphics[width=\textwidth]{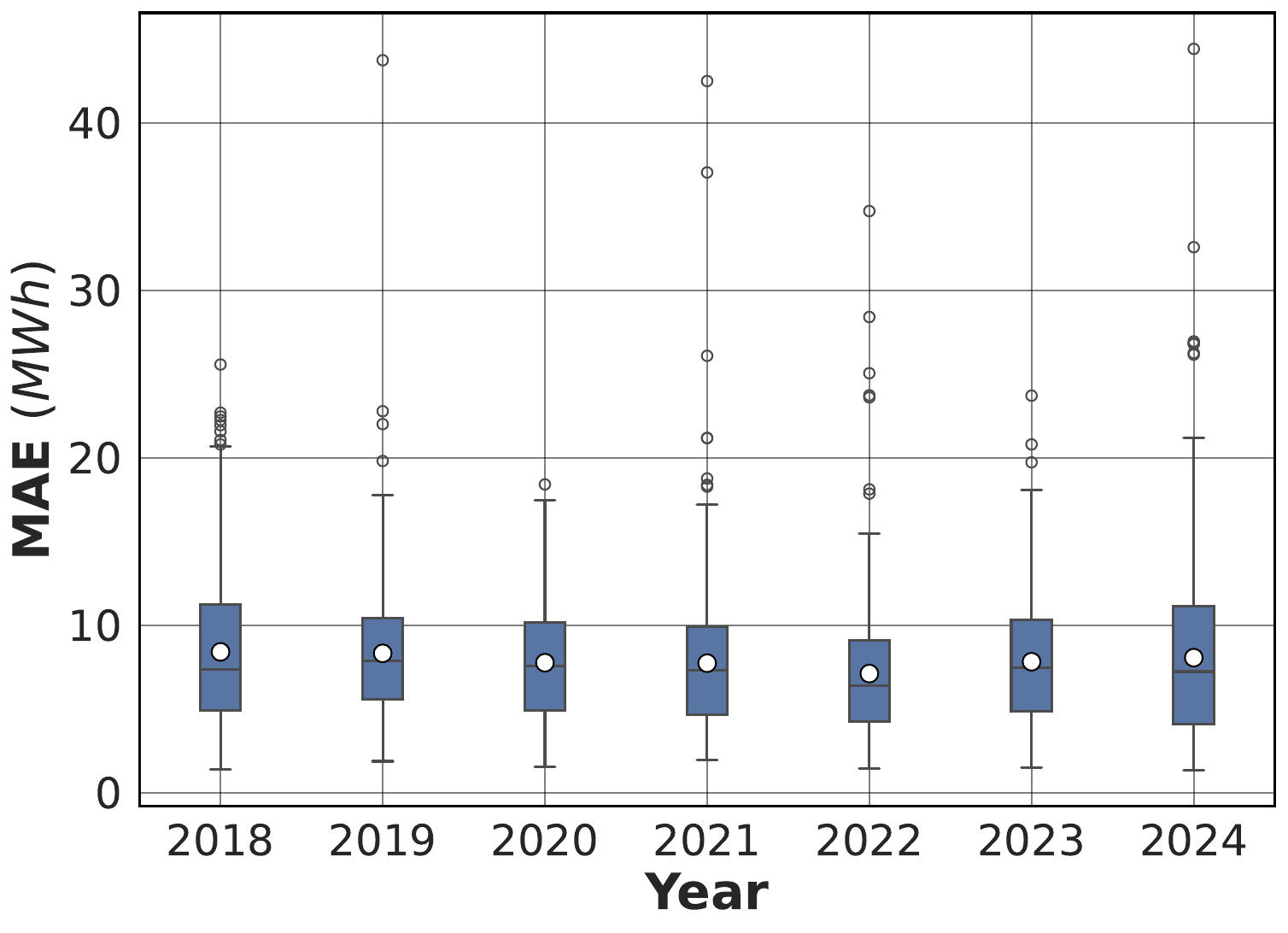} 
        \caption{Flensburg}
        \label{fig:rolling_mae_flensburg}
    \end{subfigure}
    
    \caption{Distribution of Mean Absolute Error (MAE) across different years for the three DMAs (top row) and the two city-level datasets (bottom row).}
    \label{fig:public_mae_boxplots}
\end{figure}

\subsubsection{Longitudinal Stability and Multi-Dataset Evaluation - Experiment VI}

To assess the temporal robustness of the proposed framework and its ability to generalise across diverse district heating systems across Denmark and Germany, we conducted a longitudinal evaluation using a rolling block windowing strategy. This experiment, performed on the multi-district dataset from Brønderslev, and two external public datasets for the cities of Flensburg and Aalborg, to simulate a real-world deployment scenario where the model is periodically retrained as new historical data becomes available.

Figure \ref{fig:bronderslev_mae_boxplots} illustrates the year-over-year performance for the three private districts (\acrshort{dma} A, B, C). A consistent trend of stability is observed with increasing data. For instance, in \acrshort{dma} A, the median \acrshort{mae} remains effectively constant between 2017 and 2019, with a slight reduction in interquartile range in the final year. This indicates that the model effectively leverages accumulating historical data to refine its internal representations. Similar stability is evident in \acrshort{dma} B and C, where error distributions remain compact across the evaluation years. 

The evaluation on the Aalborg residential dataset as seen in Figure \ref{fig:rolling_mae_aalborg} confirms that the Wavelet-CNN architecture generalises effectively to different aggregation levels, maintaining high precision even for residential-specific load profiles. 
The framework's robustness is further validated on the long-duration Flensburg dataset, which spans seven years (2017-2024). As shown in Figure \ref{fig:rolling_mae_flensburg}, the forecasting accuracy remains remarkably consistent over this extended horizon, with the median \acrshort{mae} hovering around $8$ MWh despite significant variations in annual weather conditions and potential infrastructure changes over the decade, although a slight decrease in forecasting performance is noted after 5 years. 

Collectively, these results demonstrate the applicability of the proposed framework in a broader context of a robust, engineering-grade solution capable of maintaining high performance over multi-year operational life cycles across geographically and structurally diverse heating networks.
\section{Conclusion}

In this work, we introduce a novel deep learning framework for multi-step day-ahead hourly heat demand forecasting in District Heating Systems. Our approach deviates from conventional time-domain models by transforming multivariate time series inputs, including historical consumption, weather data, and calendrical features, into two-dimensional time-frequency representations using the Continuous Wavelet Transform. These image-like scalograms are then processed by a convolution-based model, which is adept at extracting meaningful patterns while handling non-stationarity present in the time series data.
Our comprehensive evaluation across three distinct real-world District Metered Areas and two public benchmark datasets (Flensburg and Aalborg), demonstrates the efficacy and robustness of this framework. The quantitative results show that our model consistently outperforms a wide range of benchmarks, including classical statistical models, established machine learning methods, state-of-the-art deep learning architectures, and foundation models. Notably, the proposed framework achieved a Mean Absolute Error reduction of approximately 36\% to 43\% compared to the strongest baseline, TimeMixer. Longitudinal analysis using a rolling-window strategy over seven years further confirmed the model's temporal stability, and the ability to retain historical information.

From an application perspective, the superior peak-tracking fidelity and low error variance offered by this framework have direct implications for the intelligent control of energy systems. By minimizing forecast uncertainty, operators can safely reduce supply temperature safety margins without risking under-supply. This capability directly supports the decarbonization of the heating sector by minimizing distribution losses and facilitating the tighter integration of variable renewable heat sources. Furthermore, the competitive performance of lightweight architectures observed in our benchmarking underscores that massive computational scale is not strictly necessary for high precision, and smaller, resource-efficient models offer a promising and practical route for scalable deployment in resource-constrained industrial environments.

A key contribution of this paper, beyond the novel architecture, is the systematic elucidation of the role of different input features. Through a series of extensive ablation studies, we quantified the impact of autoregressive, meteorological, and calendrical data. We demonstrated that while daily consumption lags provide a strong baseline, the inclusion of a weekly lag ($\mathbf{c}_{168}$) anchored forecasts during transitional periods like weekends, and also provided the ability to handle rare festive events dynamically. Our analysis confirmed that ambient temperature is the single most impactful exogenous variable, and we establish that augmenting ambient temperature with additional temperature variables, along with decomposition provides a route for forecasting optimality.

Future work will focus on enhancing the model's capability to predict demand during rare and non-recurring events, a key challenge identified in this study. The current representation of holiday features, while beneficial, provides varied results for varied events.

We identify several promising avenues for future research from this study. First, the prediction of rare, non-recurring events remains a challenge due to their sparse representation in historical training samples. Future work should investigate specialized architectures, such as memory-augmented networks or few-shot learning frameworks, designed explicitly to model high-impact anomalies that standard regression models tend to smooth over. 
Additionally, at the feature transformation stage, exploration of adaptive or multi-wavelet representations, investigating whether dynamically selecting wavelet bases for different features can yield marginal gains over the fixed approach used here.
Second, from a data perspective, enhancing the model's situational awareness offers significant potential. Integrating granular external features, such as industrial operation schedules, school calendars, and municipal event logs such as concerts could bridge the gap between statistical patterns and real-world causality, allowing the model to anticipate demand shifts driven by specific cultural or operational activities. 
Finally, while our longitudinal analysis demonstrated stability, the acquisition of even larger datasets is essential to fully characterize long-term concept drift. In tandem, a deeper and more extensive evaluation of emerging foundation models is warranted to determine how their generalisation capabilities can best be harmonized with the precision of domain-specific architectures like the one presented here. 


\section*{Glossaries}
\begin{table}[h!]
\centering
\begin{tabular}{|lp{6.5cm}lp{5.5cm}|}
\toprule
\multicolumn{4}{|l|}{\textbf{Abbreviations}} \\
        &                                   &         &                                  \\
AIC     & Akaike Information Criterion      & MAE     & Mean Absolute Error              \\
CNN     & Convolutional Neural Network      & MAPE    & Mean Absolute Percentage Error   \\
CWT     & Continuous Wavelet Transform      & ML      & Machine Learning                 \\
DHS     & District Heating Systems          & MLP     & Multi-Layer Perceptron           \\
DL      & Deep Learning                     & MLR     & Multiple Linear Regression       \\
DMA     & District Metered Area             & MSE     & Mean Squared Error               \\
DNN     & Deep Neural Network               & ReLU    & Rectified Linear Unit            \\
DWT     & Discrete Wavelet Transform        & SARIMA  & Seasonal ARIMA                   \\
FFNN    & Feed Forward Neural Networks      & SARIMAX & SARIMA with eXogenous regressors \\
GPR     & Gaussian Process Regression       & SVR     & Support Vector Regression        \\
KAN     & Kolmogorov-Arnold network         & TFT     & Temporal Fusion Transformer      \\
LSTM    & Long Short-Term Memory            & TTM     & Tiny Time Mixer                  \\
\bottomrule
\end{tabular}
\end{table}

\section*{CRediT authorship contribution statement}

\textbf{Adithya Ramachandran:} Conceptualization, Data curation, Formal analysis, Investigation, Methodology, Validation, Visualization, Writing – original draft, Writing – review and editing.
\textbf{Satyaki Chatterjee:} Conceptualization, Writing – review and editing. \textbf{Thorkil Flensmark B. Neergaard:} Data curation. \textbf{Maximilian Oberndoerfer:} Resources, Data Curation. \textbf{Andreas Maier:} Resources, Supervision, Writing – review and editing. \textbf{Siming Bayer:} Conceptualization, Funding acquisition, Project administration, Resources, Supervision, Writing – review and editing.

\section*{Acknowledgements}
The authors are grateful to our research partners at Diehl Metering GmbH (Maximillian Oberndorfer, Lena Yoshihara) and Brønderslev Forsyning A/S for their domain expertise and their continuous support.

\section*{Funding sources}
This work was funded by the Bavarian State Ministry of Economic Affairs and Media, Energy and Technology, grant DIK0325/01.

\section*{Declaration of competing interests}
I have nothing to declare.


\section*{Data statement}
The data and the code base is made available in the github repository - https://github.com/lme-dpui/heat-demand-forecasting.

\newpage

\bibliographystyle{elsarticle-num} 
\bibliography{references}

\appendix

\section*{Appendices}
\renewcommand{\thesection}{\Alph{section}}

\section{Architectural Design Choices}
\label{architecture_design_choice}

\subsection{Pooling Layers}
\label{pooling_layers}

\begin{figure}[htbp]
\centering
\includegraphics[width=.75\linewidth]{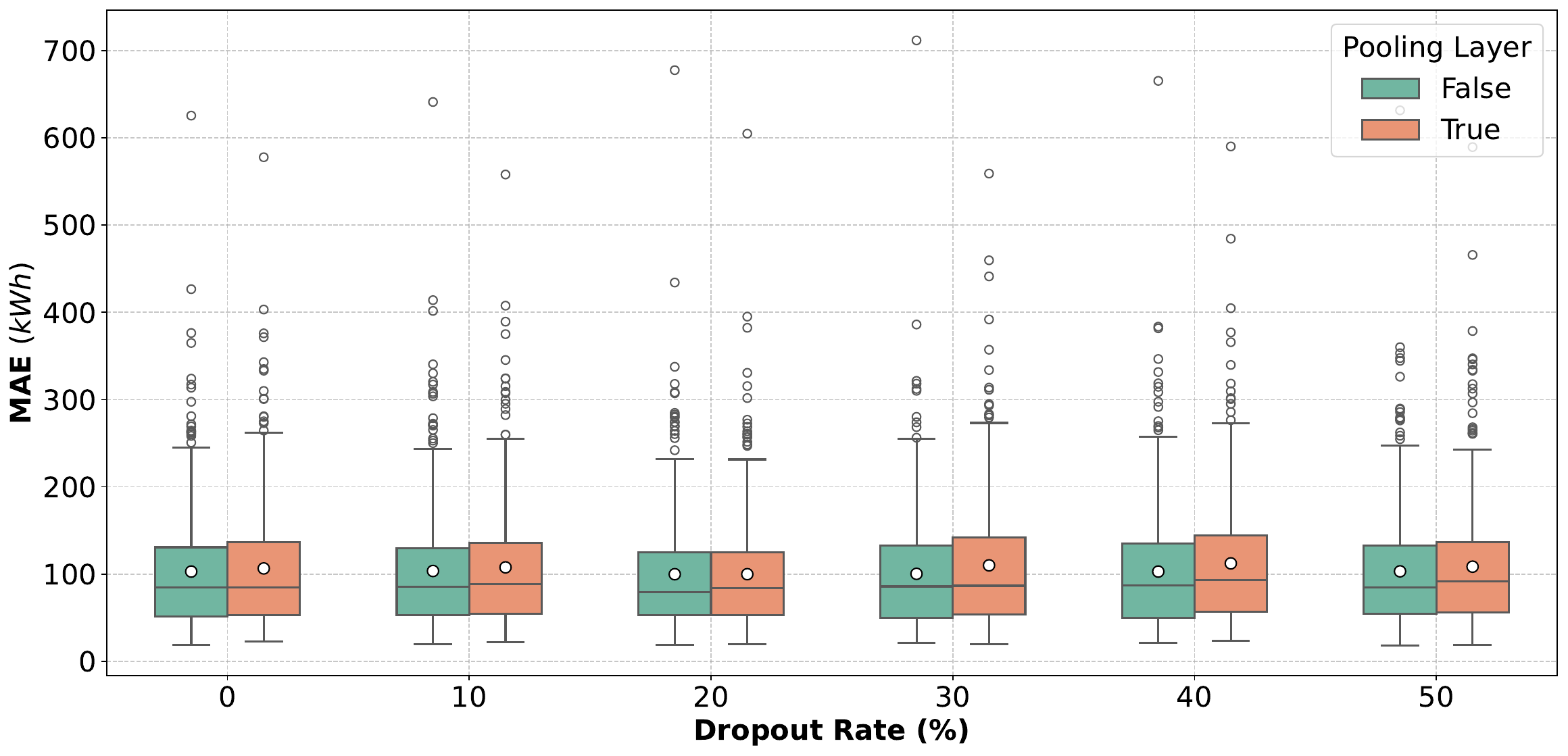}
\caption{Quantitative comparison on the effect of employing dropout with varying probabilities and the effect of pooling layer on the forecasting performance.}
\label{fig:pooling_ablation}
\end{figure}

To empirically validate the architectural decision to not employ pooling layers, we conducted an ablation study comparing the model performance with and without pooling layers across various dropout rates. The results, visualised in Figure \ref{fig:pooling_ablation}, reveal that the inclusion of pooling layers degrades forecasting accuracy compared to the non-pooled configuration. Across all tested dropout probabilities, the models without pooling achieve lower mean and median \acrshort{mae}, as well as a lower quartile range.

In standard computer vision tasks, pooling is employed to achieve translational invariance, ensuring that an object is recognized regardless of its position in the image. However, in time-series forecasting via time-frequency scalograms, in which the horizontal axis represents the exact timing, or phase, of demand events, the spatial location of a feature carries critical semantic meaning . Introducing pooling layers forces spatial downsampling, which reduces this temporal resolution. This loss of phase coherence impairs the model's ability to distinguish between a demand occurring at time $t$ versus $t+1$. Consequently, not employing pooling, the proposed architecture preserves translational equivariance, maintaining the high-fidelity localization of features required for precise multi-step regression.

\subsection{Choice of Mother Wavelet}
\label{mother_wavelet_selection}
\begin{figure}[htbp]
\centering
\includegraphics[width=0.95\linewidth]{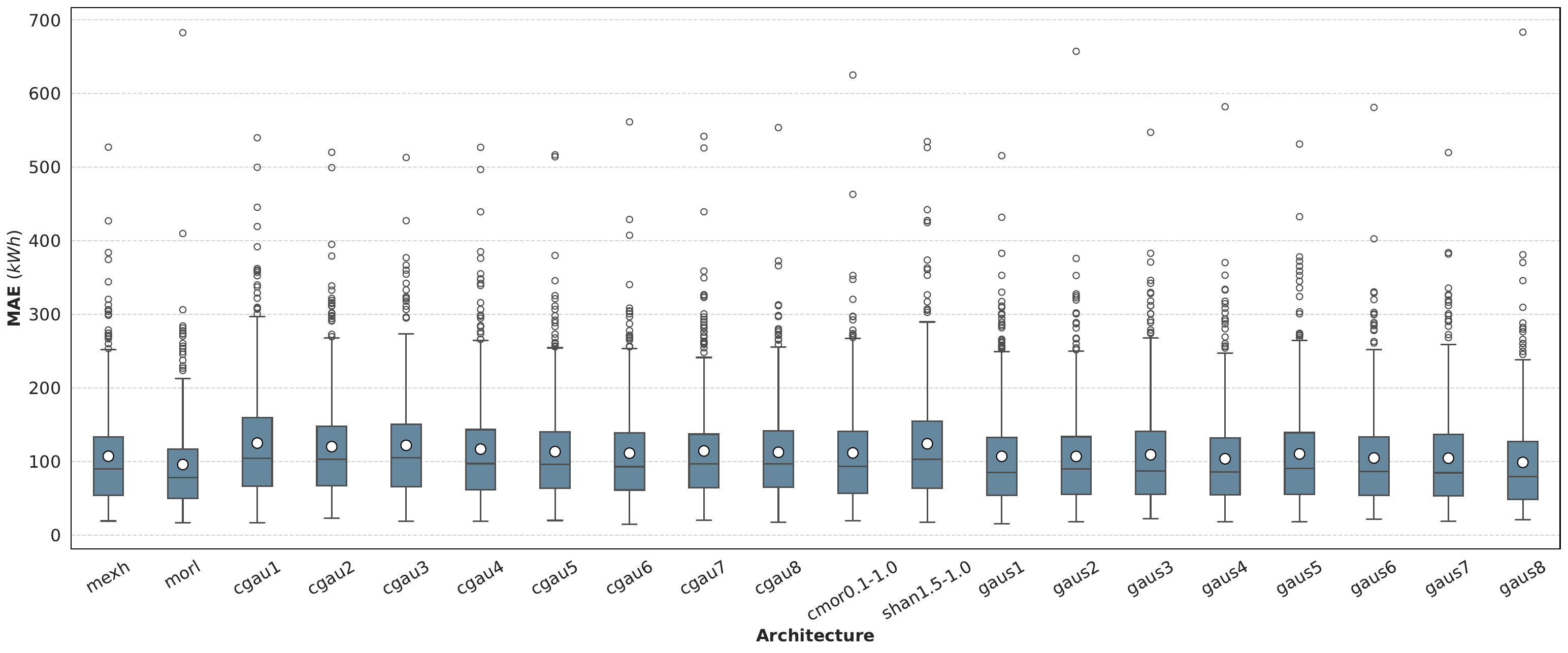}
\caption{Quantitative evaluation comparing the effect of candidate mother wavelet for \acrshort{cwt} on the forecasting performance in terms of \acrshort{mae} for \acrshort{dma} A}
\label{fig:wavelet_ablation}
\end{figure}

To assess the framework's sensitivity to the choice of mother wavelet $\psi(t)$, we evaluated performance across diverse families, including Mexican Hat, Morlet, and Gaussian derivatives. As shown in Figure \ref{fig:wavelet_ablation}, the error distributions remain remarkably stable across all candidates, with mean \acrshort{mae} variations typically below $2\%$. Further, feature scaling, which standardizes signal amplitudes and preserves relative structural patterns for mother wavelets from the same family (for example, gaussian), eliminates minor differences in captured information. 

The candidate mother wavelets mentioned in Table \ref{tab:hyperparameter_search_space} is selected as part of the hyperparameter optimisation process, and a fixed wavelet is employed across all features, as opposed to feature specific mother wavelets. Consequently, we select the mother wavelet with the best performance which includes 'morl' and 'gaus8'.

\subsection{Hyperparameter Search Space}
\label{wavelet_hyperparameter}
The proposed Wavelet-CNN architecture was optimized for the specific characteristics of heat demand data, we conducted a grid search over key architectural and training hyperparameters. The search focused on balancing model capacity (depth and width) with regularization (dropout) to prevent overfitting, given the high-dimensional nature of the input scalograms.

The search space included variations in the batch size, the depth of the convolutional feature extractors, the inclusion of pooling operations, the width and depth of the fully connected layers, and the dropout rate. Table \ref{tab:hyperparameter_search_space} details the complete search grid explored during the development phase. The final configuration was selected based on the minimization of the \acrshort{mse} loss on the validation set.

\begin{table}[htbp]
\centering
\caption{Hyperparameter search space and selected values for the proposed Wavelet-CNN framework.}
\label{tab:hyperparameter_search_space}
\renewcommand{\arraystretch}{1.2}
\begin{tabular}{l|l|c}
\hline
\textbf{Hyperparameter} & \textbf{Search Space / Grid} & \textbf{Selected Value} \\
\hline
\textbf{Training Dynamics} & & \\
Learning Rate ($\eta$) & $\{0.01, 0.001, 0.0001\}$ & $0.001$ \\
Batch Size & $\{16, 32, 64, 256\}$ & $32$ \\
\hline
\textbf{Feature Transformation} & & \\
 & mexh, morl & \multirow{3}{*}{\begin{tabular}[c]{@{}c@{}}morl (DMA A, C)\\ gaus8 (DMA B)\end{tabular}} \\
Mother Wavelet Candidate & gaus1-gaus8, cgau1-cgau8 & \\
 & cmor0.1-1.0, shan1.5-1.0 & \\
\hline
\textbf{Convolutional Block} & & \\
Filter Configuration & $\{[16, 32, 64], [32, 64, 128]\}$ & $[32, 64, 128]$ \\
Pooling Operation & $\{\text{True}, \text{False}\}$ & $\text{False}$ \\
\hline
\textbf{Linear Head} & & \\
 & $[1024, 1024]$ & \\
 & $[2048, 1024]$ & \\
 & $[2048, 2048]$ & \\
Hidden Layer Dimensions & $[4096, 1024]$ & $[1024, 1024]$ \\
(Neurons) & $[4096, 2048]$ & \\
 & $[4096, 4096]$ & \\
 & $[4096, 2048, 1024]$ & \\
\hline
\textbf{Regularization} & & \\
Dropout Rate ($p$) & $\{0.0, 0.1, 0.2, 0.3, 0.4, 0.5\}$ & $0.1$ \\
\hline
\end{tabular}
\end{table}

\subsection{Model Convergence}
\begin{figure}[h!]
    \centering 

    \begin{subfigure}[b]{0.45\linewidth}    
        \centering
\includegraphics[width=0.9\linewidth]{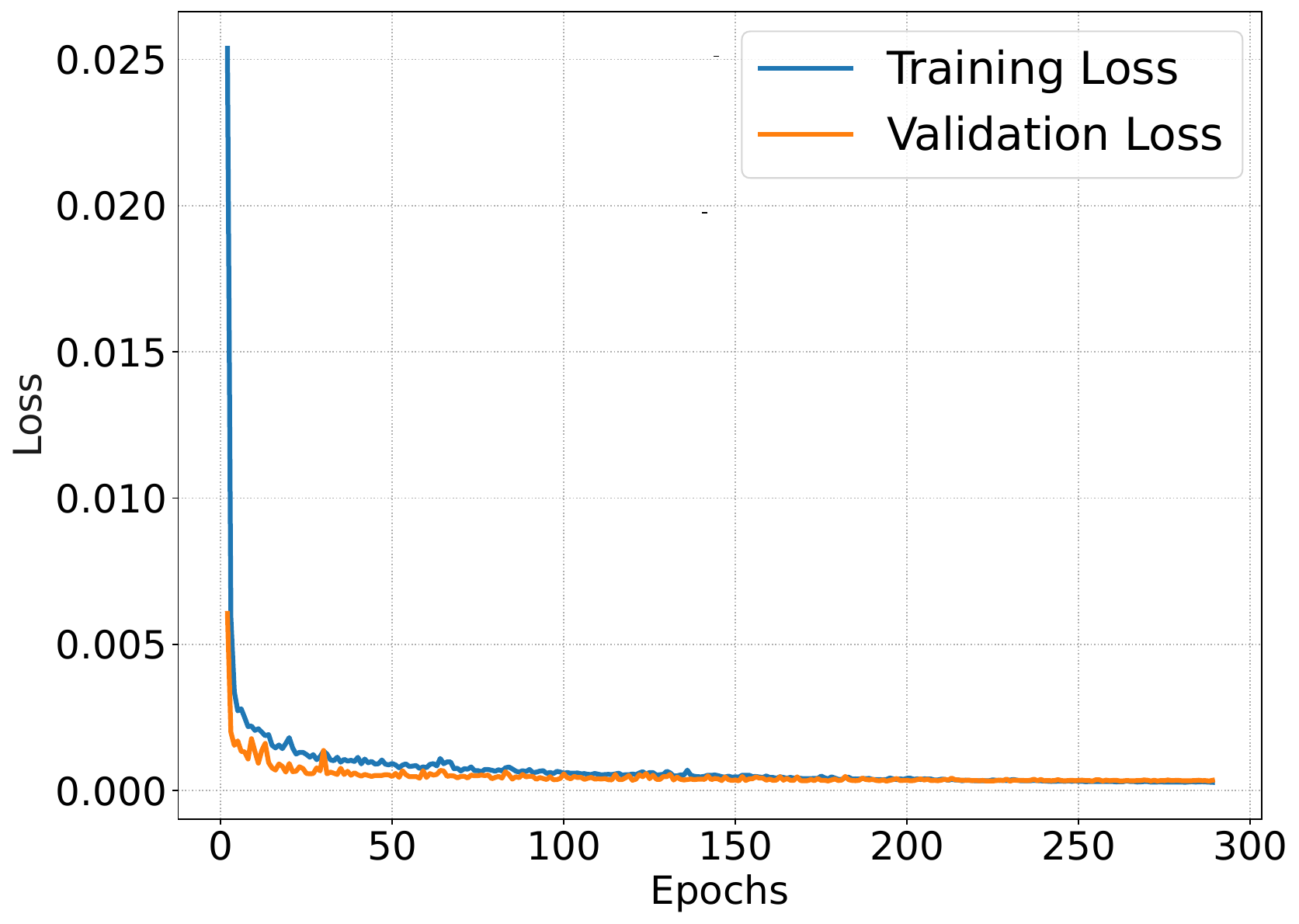}
        \caption{\acrshort{dma} A}
        \label{fig:dma_a_loss}
    \end{subfigure}    
    \begin{subfigure}[b]{0.45\linewidth}
        \centering
\includegraphics[width=0.9\linewidth]{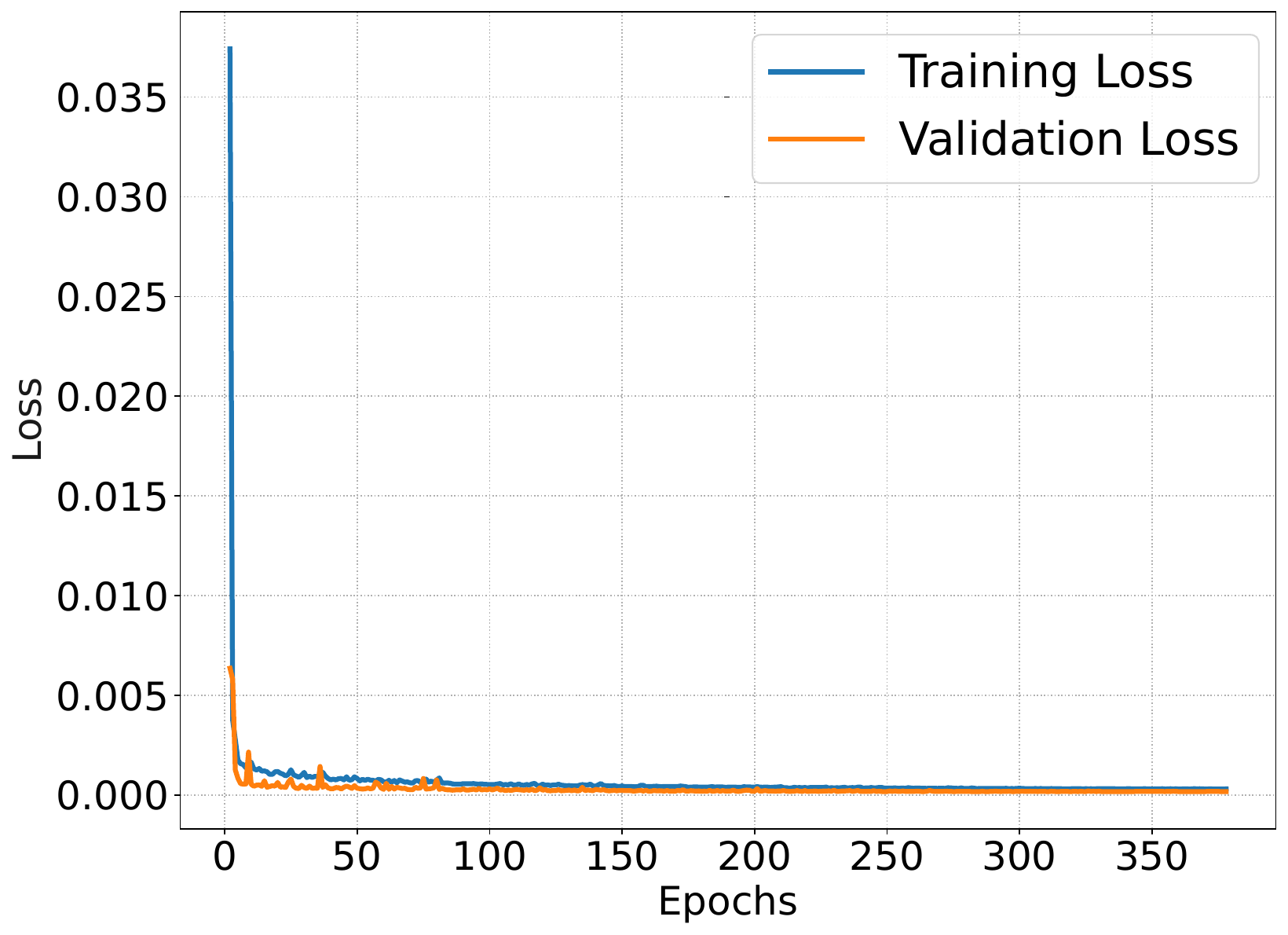}
        \caption{\acrshort{dma} B}
        \label{fig:dma_b_loss}
    \end{subfigure} 

\vspace{0.25cm}
    
    \begin{subfigure}[b]{0.5\linewidth}
        \centering
\includegraphics[width=0.9\linewidth]{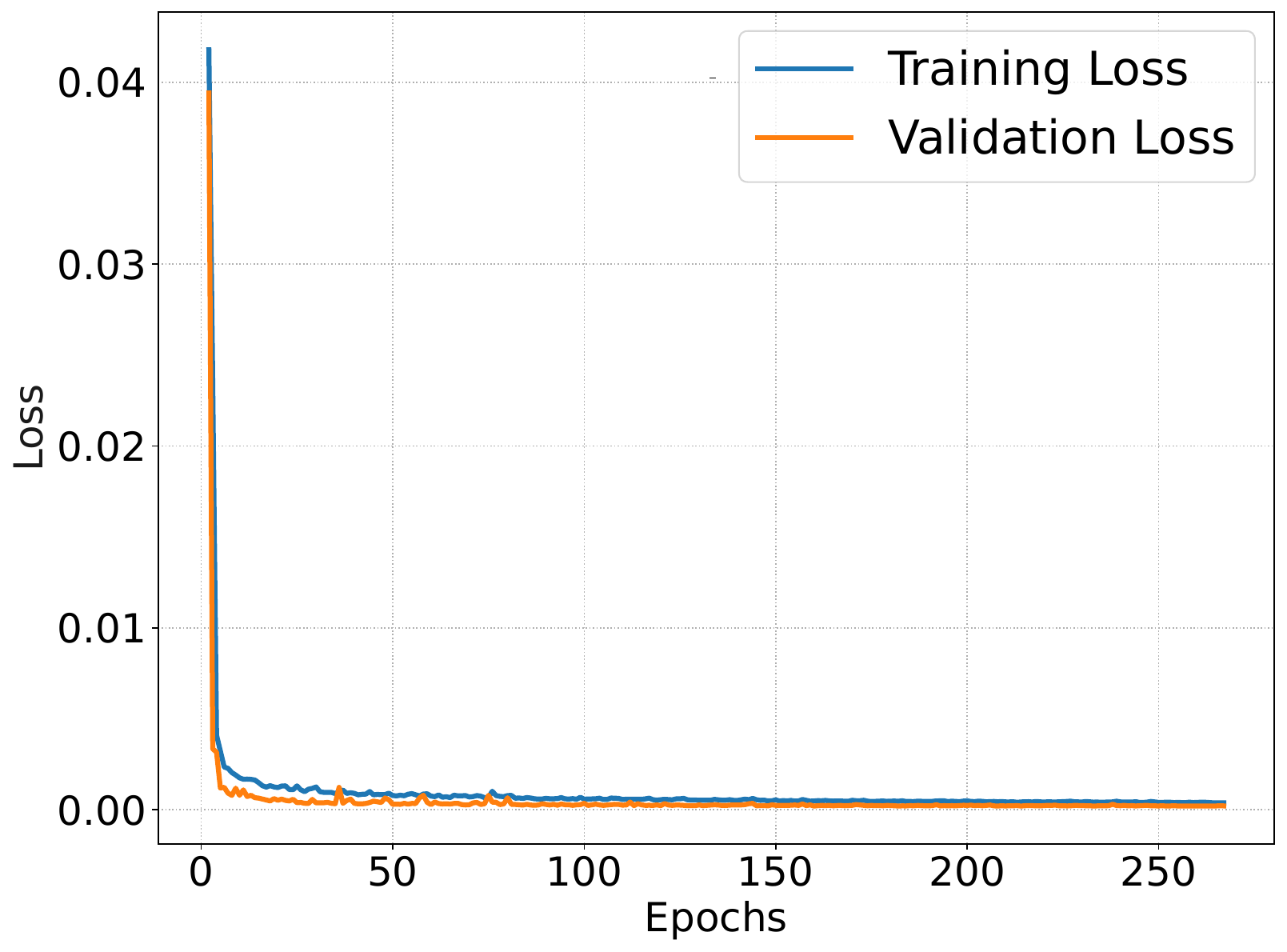}
        \caption{\acrshort{dma} C}
        \label{fig:dma_c_loss}
    \end{subfigure}

    \caption{Illustration of training and validation loss curves for the proposed Wavelet-CNN models trained for a maximum of 1000 epochs with a patience 50 epochs. The box plots in (a), (b), and (c) show the loss curves for \acrshort{dma} A, \acrshort{dma} B, and \acrshort{dma} C, respectively.}

    \label{fig:dma_loss_curves}
    
\end{figure}

The learning curves for the proposed model across all three districts demonstrate convergence and training stability. As illustrated in Figure \ref{fig:dma_loss_curves}, both training and validation losses exhibit a sharp initial descent within the first 20 epochs before settling into a stable plateau. Crucially, the validation loss tracks the training loss closely throughout the entire optimisation process, with no evidence of divergence indicating overfitting to the training set.

\subsection{Training and Inference Efficiency}

To assess the operational feasibility of the Wavelet-CNN framework, we evaluated the computational costs across three the \acrshort{dma}s over 5 iterations and the findings are documented in Table \ref{tab:computational_efficiency}.

\begin{table}[htbp]
\centering
\caption{Computational efficiency metrics per \acrshort{dma}. Values represent the Mean $\pm$ Standard Deviation across 5 iterations. Inference latency is reported in milliseconds per sample.}
\label{tab:computational_efficiency}
\renewcommand{\arraystretch}{1.2}
\setlength{\tabcolsep}{12pt}
\begin{tabular}{lcc}
\toprule
\textbf{Dataset} & \textbf{Training Time (min)} & \textbf{Inference Latency (ms/sample)} \\
\midrule
DMA A & $2.76 \pm 0.26$ & $0.23 \pm 0.14$ \\
DMA B & $3.10 \pm 0.46$ & $0.24 \pm 0.17$ \\
DMA C & $2.48 \pm 0.11$ & $0.22 \pm 0.14$ \\
\midrule
\textbf{Average} & $\mathbf{2.78 \pm 0.24}$ & $\mathbf{0.23 \pm 0.15}$ \\
\bottomrule
\end{tabular}
\end{table}

The results indicate consistent efficiency across differing datasets. The training time varies slightly by \acrshort{dma}, due to convergence rates. However, all models converge in under $3.5$ minutes, facilitating rapid retraining cycles.

Most critically for deployment, the inference latency remains uniformly low across all areas, averaging approximately $\mathbf{0.23}$ \textbf{ms/sample}. This sub-millisecond response time confirms that the proposed framework is well-suited for real-time control applications, capable of handling high-throughput forecasting tasks across multiple district heating sectors simultaneously. 

All experiments were conducted on a workstation equipped with an NVIDIA Quadro RTX 8000 GPU (48 GB VRAM).

\section{Hyper-Parameter Tuning} \label{benchmarks}
To ensure a fair and rigorous comparative evaluation, all benchmark models discussed in the main paper were subjected to hyperparameter optimisation. The models were tuned to achieve their best performance on our specific heat demand forecasting task. This section details the search space explored, the optimisation procedures, and the final configurations for each model.

\subsection{Statistical and Machine Learning Baselines}

\acrshort{sarimax} model is implemented using the pmdarima library's \texttt{auto\_arima} function to find the optimal parameters using a stepwise search strategy. This function automatically searches for the optimal model order by minimizing the Akaike Information Criterion (AIC). The search was configured with the following constraints:
\begin{itemize}
    \item \textbf{Seasonal Period ($m$):} Set to 24 to capture the daily periodicity in the hourly data.
    \item \textbf{Non-seasonal Orders:} The maximum orders for the autoregressive ($p$), moving average ($q$), and differencing ($d$) components were set to 6, 6, and 1, respectively.
    \item \textbf{Seasonal Orders:} The maximum orders for the seasonal autoregressive ($P$), seasonal moving average ($Q$), and seasonal differencing ($D$) components were set to 1, 1, and 1, respectively.

\end{itemize}

The XGBoost model was optimized through a grid search over its most critical hyperparameters. The model was configured with a recursive refitting strategy, where it was retrained for each 24-hour forecast window using the most recent actual values. The search space and the final selected hyperparameters are detailed in Table~\ref{tab:xgboost_params}.

\begin{table}[h!]
\centering
\caption{Hyperparameter tuning for the XGBoost model.}
\label{tab:xgboost_params}
\begin{tabular}{@{}l|l|c@{}}
\toprule
\textbf{Hyperparameter} & \textbf{Search Space} & \textbf{Selected Value} \\ \midrule
N Estimators & \{200, 400, 600, 800, 1000, 1500, 2000\} & 200 \\
Learning Rate & \{0.1, 0.01, 0.001, 0.0001\} & 0.1 \\
Maximum Depth & \{2, 4, 6, 8, 10, 12, 24\} & 4 \\
Subsample & [0.6, 0.7, 0.8, 0.9, 1.0] & 0.8 \\ \bottomrule
\end{tabular}
\end{table}

\subsubsection{Deep Learning Models}

A canonical \acrshort{lstm} network was included as a neural network baseline. The architecture consists of a stack of LSTM layers followed by a fully connected (linear) layer for the final prediction with 24 neurons. The hyperparameter search space and the final selected configuration are detailed in Table~\ref{tab:lstm_params}.

\begin{table}[h!]
\centering
\caption{Hyperparameter tuning for the LSTM model.}
\label{tab:lstm_params}
\begin{tabular}{@{}lll@{}}
\toprule
\textbf{Hyperparameter}       & \textbf{Search Space}      & \textbf{Selected Value} \\ \midrule
{LSTM Layers}          & \{1, 2, 4\}                & 4                       \\
{LSTM Hidden Units}    & \{16, 32, 64, 128\}        & 32                      \\
{Learning Rate}        & \{1e-4, 5e-4, 1e-3\}       & 1e-3                    \\ \bottomrule
\end{tabular}
\end{table}

To ensure a fair and rigorous comparison, equal computing power, all Transformer and MLP-based baselines were benchmarked under a unified experimental protocol. We utilized the Optuna framework to conduct a systematic hyperparameter search for each architecture. The search budget was standardized to 50 trials per model to prevent implicit tuning advantages. The objective was to minimise the validation \acrshort{mse} loss.

All baseline models were trained using the same computational constraints as the proposed framework with a fixed batch size of 32, a maximum of $1000$ epochs, and an early stopping patience of $50$ epochs to prevent overfitting. Table~\ref{tab:transformer_params} provides a comprehensive summary of this process, detailing the specific search space explored for each major hyperparameter and the final, best-performing configuration selected for each model. For the input sequence length and label length pairs, values of (24, 12), (48, 24), (96, 48), (168, 96), and (336, 168), are analysed and (96, 48) is choosen for the transformer based models.

\begin{table}[h!]
\centering
\caption{Hyperparameter search space and final configurations for Deep Learning baselines. All models were tuned using Optuna. The search space reflects the grid swept during the 50-trial optimisation process.}
\label{tab:transformer_params}
\resizebox{\textwidth}{!}{%
\begin{tabular}{@{}l l cccccc@{}}
\toprule
\textbf{Hyperparameter} & \textbf{Search Space} & \textbf{Informer} & \textbf{Autoformer} & \textbf{PatchTST} & \textbf{DLinear} & \textbf{TimesNet} & \textbf{TimeMixer} \\ \midrule
\multicolumn{8}{c}{\textit{General Training Parameters}} \\ \midrule
Learning Rate & \{0.01, 0.001, 0.0001\} & 0.0001 & 0.0001 & 0.0001 & 0.001 & 0.0001 & 0.001 \\
Dropout & $[0.05, 0.3]$ & 0.1 & 0.1 & 0.1 & 0.1 & 0.1 & 0.1 \\ \midrule

\multicolumn{8}{c}{\textit{Common Architectural Parameters}} \\ \midrule
$d_{model}$ & \{32, 64, 256, 512, 1024\} & 512 & 1024 & 512 & -- & 1024 & 32 \\
$n_{heads}$ & \{4, 6, 8, 10\} & 6 & 8 & 8 & -- & 8 & -- \\
$d_{ff}$ & \{128, 256, 512, 1024, 2048, 4096\} & 2048 & 4096 & 2048 & -- & 4096 & 256 \\
$e_{layers}$ & \{1, 2, 4, 6, 8\} & 4 & 2 & 2 & 1 & 2 & 2 \\
$d_{layers}$ & \{1, 2, 4\} & 2 & 1 & -- & -- & 1 & -- \\ \midrule

\multicolumn{8}{c}{\textit{Model-Specific Parameters}} \\ \midrule
Moving Avg. & \{13, 25, 49\} & -- & 25 & -- & 25 & -- & 25 \\
Patch/Seg Len & \{8, 16, 24\} & -- & -- & 16 & -- & -- & 16 \\
Factor & \{7, 9\} & 9 & -- & -- & -- & -- & -- \\
Top-$k$ & $[3, 7, 10]$ & -- & 5 & -- & -- & 5 & 5 \\ 
Channel Indep. & \{0, 1\} & 0 & 0 & 1 & 1 & 0 & 0 \\ \midrule

\multicolumn{8}{c}{\textit{TimeMixer}} \\ \midrule
Downsample Layers & $[1, 3]$ & -- & -- & -- & -- & -- & 2 \\
Downsample Window & \{2, 3, 5\} & -- & -- & -- & -- & -- & 2 \\
\bottomrule
\end{tabular}
}
\end{table}

\subsection{Foundation Models}
To assess the emerging paradigm of pre-trained models, we evaluate two foundation models fine-tuned for this task.
First, we utilize the \acrshort{ttm}-r2 variant \cite{ekambaram2024tinytimemixersttms}, pre-trained on a diverse corpus of energy time series. The model is fine-tuned on our dataset for 1000 epochs with a learning rate of $1 \times 10^{-5}$ to adapt its representations without catastrophic forgetting, as suggested by the authors. A context length of 1024 was used.
Similarly, we evaluate Chronos-2 \cite{ansari2025chronos2}, with the 1000 epochs and a learning rate of $1 \times 10^{-5}$. We employ the base variant and fine-tune it using the same protocol to ensure fair comparison against the domain-specific models.

\section{Holidays}
Table~\ref{tab:holiday_details} provides a comprehensive list of the public holidays observed in Denmark for the year 2019, which is a representative year from our dataset. The table includes the specific date, the day of the week on which the holiday fell, its official Danish name, and an English translation for clarity. This information serves as a reference for the specific holidays that are analyzed in our experimental evaluation of different holiday handling strategies.

\begin{table}[h!]
\centering
\caption{Translation and Day of the Week for Danish Holidays (2019).}
\label{tab:holiday_details}
\begin{tabular}{llll}
\toprule
\textbf{Date (2019)} & \textbf{Day of the Week} & \textbf{Danish Name} & \textbf{English Translation} \\
\midrule
2019-01-01 & Tuesday   & Nytårsdag             & New Year's Day \\
2019-04-18 & Thursday  & Skærtorsdag           & Maundy Thursday \\
2019-04-19 & Friday    & Langfredag            & Good Friday \\
2019-04-21 & Sunday    & Påskedag              & Easter Sunday \\
2019-04-22 & Monday    & Anden påskedag        & Easter Monday \\
2019-05-17 & Friday    & Store bededag         & Great Prayer Day \\
2019-05-30 & Thursday  & Kristi himmelfartsdag & Ascension Day \\
2019-06-09 & Sunday    & Pinsedag              & Pentecost Sunday (Whit Sunday) \\
2019-06-10 & Monday    & Anden pinsedag        & Whit Monday \\
2019-12-25 & Wednesday & Juledag               & Christmas Day \\
2019-12-26 & Thursday  & Anden juledag         & Second Day of Christmas (Boxing Day) \\
\bottomrule
\end{tabular}
\end{table}





\end{document}